\documentclass[journal]{IEEEtran}
\IEEEoverridecommandlockouts
\usepackage{cite}
\usepackage{amsmath,amssymb,amsfonts}
\usepackage{algorithmic}
\usepackage{graphicx}
\usepackage{textcomp}
\usepackage{xcolor}
\def\BibTeX{{\rm B\kern-.05em{\sc i\kern-.025em b}\kern-.08em
    T\kern-.1667em\lower.7ex\hbox{E}\kern-.125emX}}

\usepackage{array,booktabs,ragged2e}
\newcolumntype{R}[1]{>{\RaggedLeft\arraybackslash}p{#1}}
\newcommand*\rot{\rotatebox{90}}
\usepackage{siunitx}
\usepackage{subfig}
\usepackage{graphicx}
\usepackage{multirow}
\usepackage{makecell}
\usepackage{pgfplots}
\usepackage{hyperref}
\pgfplotsset{compat=1.8}
\usepgfplotslibrary{statistics}

\usepackage{academicons}
\definecolor{orcidlogocol}{HTML}{A6CE39}
\DeclareRobustCommand{\orcidicon}{%
	\begin{tikzpicture}
		\draw[lime, fill=lime] (0,0) 
		circle [radius=0.16] 
		node[white] {{\fontfamily{qag}\selectfont \tiny ID}};
		\draw[white, fill=white] (-0.0625,0.095) 
		circle [radius=0.007];
	\end{tikzpicture}
	\hspace{-2mm}
}
\newcommand{\orcidChristian}{\href{https://orcid.org/0000-0003-4822-2844}{\orcidicon}}
\newcommand{\orcidWalter}{\href{https://orcid.org/0000-0003-4565-1272}{\orcidicon}}
\newcommand{\orcidKnoll
}{\href{https://orcid.org/0000-0003-4840-076X}{\orcidicon}}

\begin{document}

\title{TUMTraf Event: Calibration and Fusion\\ Resulting in a Dataset for\\ Roadside Event-Based and RGB Cameras}

\author{\IEEEauthorblockN{Christian Cre\ss$^{\star}$\orcidChristian \qquad Walter Zimmer\orcidWalter \qquad Nils Purschke \qquad Bach Ngoc Doan \qquad Sven Kirchner \newline Venkatnarayanan Lakshminarasimhan \qquad Leah Strand \qquad  Alois C. Knoll\orcidKnoll}
	\thanks{All authors are with the Chair of Robotics, Artificial Intelligence and Real-time Systems, TUM School of Computation, Information and Technology, Technical University of Munich, Munich, Germany. \newline
		E-mail: christian.cress@tum.de, \newline knoll@in.tum.de \newline
		$^{\star}$ Corresponding author.			
}}


\maketitle

\begin{abstract}
Event-based cameras are predestined for Intelligent Transportation Systems (ITS). They provide very high temporal resolution and dynamic range, which can eliminate motion blur and improve detection performance at night. However, event-based images lack color and texture compared to images from a conventional RGB camera. Considering that, data fusion between event-based and conventional cameras can combine the strengths of both modalities. For this purpose, extrinsic calibration is necessary. To the best of our knowledge, no targetless calibration between event-based and RGB cameras can handle multiple moving objects, nor does data fusion optimized for the domain of roadside ITS exist. Furthermore, synchronized event-based and RGB camera datasets considering roadside perspective are not yet published. To fill these research gaps, based on our previous work, we extended our targetless calibration approach with clustering methods to handle multiple moving objects. Furthermore, we developed an early fusion, simple late fusion, and a novel spatiotemporal late fusion method. Lastly, we published the TUMTraf Event Dataset, which contains more than 4,111 synchronized event-based and RGB images with 50,496 labeled 2D boxes. During our extensive experiments, we verified the effectiveness of our calibration method with multiple moving objects. Furthermore, compared to a single RGB camera, we increased the detection performance of up to +9\% mAP in the day and up to +13\% mAP during the challenging night with our presented event-based sensor fusion methods. The TUMTraf Event Dataset is available at \href{https://innovation-mobility.com/tumtraf-dataset}{https://innovation-mobility.com/tumtraf-dataset}. 
\end{abstract}

\begin{IEEEkeywords}
Event-Based Cameras, RGB Cameras, Sensor Fusion, Targetless Calibration, Multi-modal Dataset, Intelligent Transportation Systems
\end{IEEEkeywords}

\begin{figure}[htb]
	\centerline{\includegraphics[width=\linewidth]{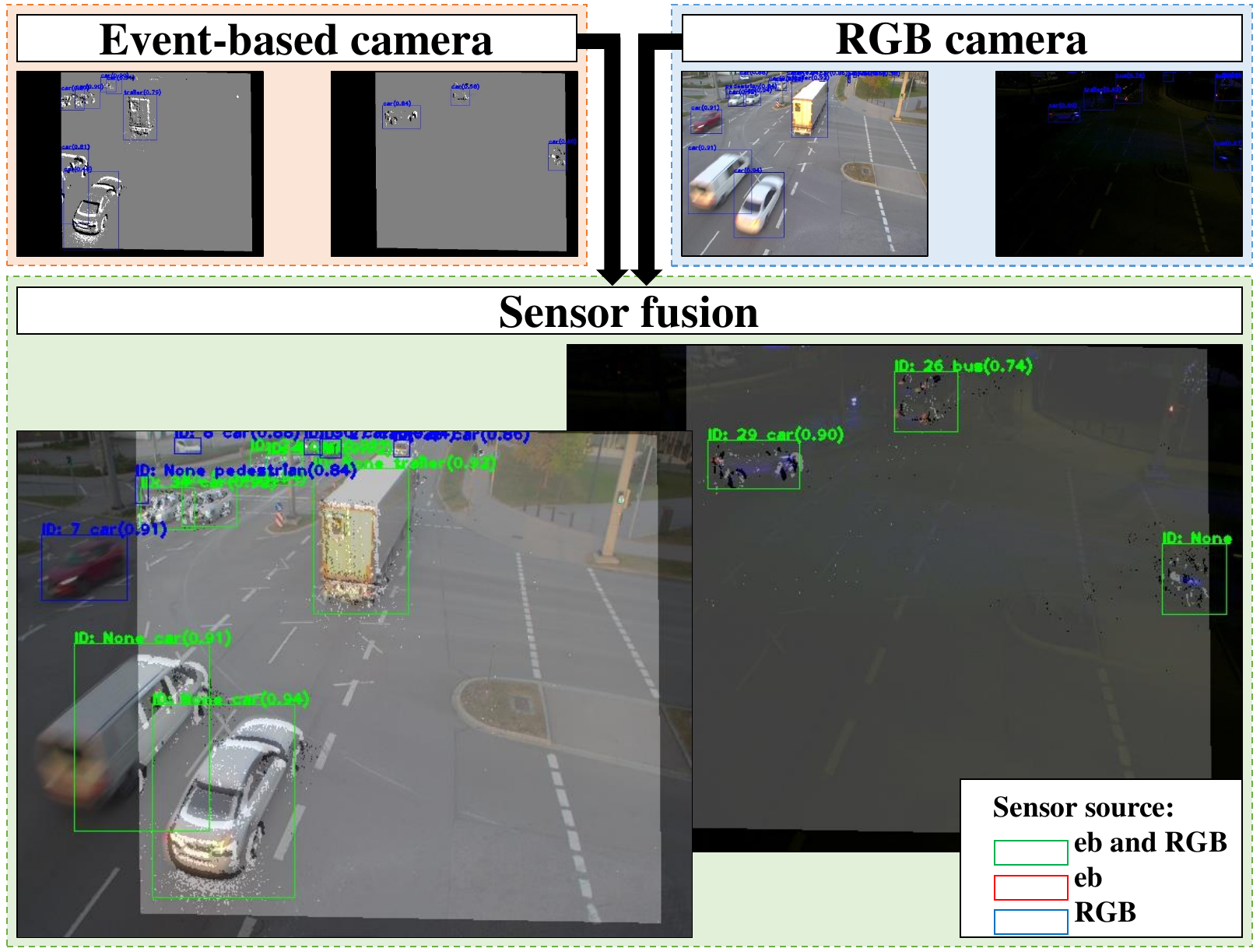}}
	\caption{This figure shows the sensor fusion between event-based and RGB cameras and its impact during a sunny day and a night in sleet. The blue bounding boxes in the event-based respectively RGB camera section represent detections without fusion. However, in the sensor fusion section, a green bounding box indicates that an object was detected by the event-based and the RGB camera. A blue bounding box shows detection exclusively by the RGB camera, and a red bounding box shows detection exclusively by the event-based camera (not available here). A unique track ID is assigned when objects are detected in several frames.}
	\label{fig:intro}
\end{figure}

\section{Introduction}
The principle of event-based cameras is the asynchronous recognition of changes in the brightness of each pixel. This technique results in a very high temporal resolution and a very high dynamic range \cite{Chen.2020b, Gallego.2022}. Therefore, an event-based camera is predestined for Intelligent Transportation Systems (ITS). Even in difficult visibility conditions, these systems require robust and accurate perception of traffic participants. Here, event-based cameras can achieve significant improvements, e.g., at night, with poor visibility or even fast-moving objects, leading to motion blur when using conventional cameras. However, the disadvantages of these novel sensors are the lack of color and texture information compared to conventional cameras and the fact that event-based cameras only detect moving objects. An optimized combination of roadside event-based and conventional RGB cameras offers advantages from both modalities. So far, ITS perception mainly relies on conventional cameras, Radars, and Lidars. Event-based cameras are yet to be widespread but are slowly being established in this area \cite{Cre.2023}. For this reason, an investigation into calibration, detection, and data fusion between roadside event-based and conventional cameras is necessary.  

Detection and tracking with a stationary event-based camera mounted on roadside ITS, as far as we know, was first performed by \cite{Chen.2018}. Here, clustering and tracking methods achieved sufficient performance. In addition, \cite{Chen.2019} used a stationary event-based camera for pedestrian detection with a convolutional neural network (CNN). For data fusion, the authors \cite{Cao.2021, Zhang.61720236242023, Zhou.2023, Tomy.2022} carried out fusion on the feature level from a perspective with ego-motion. Furthermore, \cite{Tomy.2022} performed early fusion, and \cite{Sun.2023b} performed late fusion. Nevertheless, a fusion between event-based and RGB cameras is a developing field \cite{Tomy.2022}; more knowledge about this topic in the area of stationary sensors in ITS is desirable. The authors \cite{Binas.04112017, Mueggler.2017b, Sironi.2018, Zhu.2018b, Cheng.2019, Hu.18052020, Perot.2020, Tournemire.23012020, Gehrig.2021, Gehrig.22112022, DSEC-Detection.27022024} provided datasets from an ego-motion perspective. Unfortunately, data with ego-motion significantly differs from data from stationary cameras. Simulators \cite{HenriRebecq.2018b, Mueggler.2017b} could tackle this problem but suffer from the sim2real gap \cite{Tomy.2022}. Another possibility is using pseudo-labels based on RGB camera detections \cite{Chen.2018b}. This approach promises sufficient results, but an extrinsic calibration between event-based and RGB cameras is required. To calculate the intrinsic camera matrix, the authors \cite{Muglikar.2021, Huang.2021, Q.Zhang.2020, DominguezMorales.2019, E.Mueggler.2015, Plasberg.742022772022, E.Mueggler.2014} used classical checkerboards, which were moved in front of the camera, or used checkerboards, which emitted changes in brightness. However, calibration patterns are impractical on an ITS. Therefore, our previous work \cite{Cre.642023672023} presented a novel targetless extrinsic calibration method between event-based and RGB cameras. The approach produced adequate results but needed to be more robust in situations with multiple moving objects (e.g., several cars or shadows), which were not equally imaged by both cameras. These gaps are intended to be closed with this work.

To the best of our knowledge, there exists neither a targetless calibration approach that can handle multiple moving objects nor a data fusion between event-based and RGB cameras that takes the unique characteristics of a stationary camera setup during day and night in the domain of roadside ITS into account. Furthermore, there is still a lack of datasets in the mentioned domain.
 
For this reason, we improve in this work our previously presented targetless calibration approach between event-based and RGB cameras \cite{Cre.642023672023} to increase the practicability and the handling of multiple moving objects. Furthermore, to combine the advantages of both sensor modalities, we provide three fusion approaches between event-based and conventional cameras: Early Fusion (EF), Simple Late Fusion (SLF), and Spatiotemporal Late Fusion (STLF) based on SORT \cite{Bewley.2016} tracking. Here, we demonstrate the effectiveness of our calibration and fusion methods with comprehensive experiments based on real data and comparisons with state-of-the-art datasets and fusion methods. For the experiments with sensor fusion, we analyzed the combination of event-based and RGB cameras during a sunny day and a night in the sleet on our ITS \cite{A9_Testbed}, see Figure \ref{fig:intro}. Lastly, we would like to share our novel $\text{TUMTraf Event Dataset}$. It contains synchronized event-based and RGB images, which show a complex road intersection with several traffic scenarios during the day and night. The dataset labels for training and validation are based on partially optimized pseudo-labels extracted from the YoloV7 \cite{arXiv.org.07072022b, Wong.14012023, Chen.02012020} detector using an extrinsic calibration matrix obtained by our targetless calibration tool. The test dataset is carefully labeled and allows accurate ground truth data analysis.

In summary, the main contributions of this work are:
\begin{itemize}
	\item Based on our previous work \cite{Cre.642023672023}, an improved targetless calibration between event-based and RGB cameras, which can handle multiple moving objects.
	\item The fusion methods Early Fusion (EF), Simple Late Fusion (SLF), and Spatiotemporal Late Fusion (STLF) between event-based and RGB cameras to profit from both sensor modalities' advantages and reduce their limitations.
	\item Comprehensive experiments with our fusion algorithms based on real data and comparisons with other state-of-the-art datasets and methods.
	\item The novel TUMTraf Event Dataset, which contains spatiotemporal calibrated event-based and RGB images. The dataset can be used to show robust detection of traffic participants with an event-based camera during day and night in the domain of Intelligent Transportation Systems.   
\end{itemize}

\begin{table*}[h]
	\caption{This table lists popular event-based camera datasets. Here, we compare the general purpose, the perspective, the resolution of the event-based camera used, and the illumination scenarios of the datasets. Furthermore, in the case of 2D detection and lane extraction, we analyze the number of labeled frames, object classes, and labels.}
	\begin{center}
		\begin{tabular}{|r|r|r|r|r|rl|r|r|r|}
			\hline
			\multicolumn{7}{|c|}{\textbf{General Properties}} & \multicolumn{3}{c|}{\textbf{2D Detection Properties}} \\
			\hline
			\textbf{Name} & \textbf{Year} & \textbf{Purpose} & \textbf{Perspective} & \textbf{EB-Resolution} & \multicolumn{2}{c|}{\textbf{Day} \textbf{Night}} & \textbf{\# Lab. Frames} & \textbf{\# Classes} & \textbf{\# Labels} \\
			\hline
			DDD17 \cite{Binas.04112017} & 2017 & End-to-End Driving & Vehicle & 346 $\times$ 260 & \hspace{3pt} \textbf{X} & \textbf{X} &  &  &  \\
			DAVIS 240C \cite{Mueggler.2017b} & 2017 & Depth \& Pose & Mov. objects & 240 $\times$ 180 & X & - &  &  &  \\
			N-CARS \cite{Sironi.2018} & 2018 & 2D Detection & Vehicle & 304 $\times$ 240 & X & -$^\mathrm{*}$ & 48,000 & 2 & 24,029 \\
			MVSEC \cite{Zhu.2018b} & 2018 & Depth \& Pose & Mov. objects & 346 $\times$ 260 & X & - &  &  &  \\
			DET \cite{Cheng.2019} & 2019 & Lane Extraction & Vehicle & \textbf{1280} $\times$ \textbf{800} & X & -$^\mathrm{*}$ & 5,424 & 1 & 17,103 \\
			DDD20 \cite{Hu.18052020} & 2020 & End-to-End Driving & Vehicle & 346 $\times$ 260 & \textbf{X} & \textbf{X} &  &  &  \\
			1 MADD \cite{Perot.2020} & 2020 & 2D Detection & Vehicle & \underline{1280 $\times$ 720} & X & - & \textbf{3,164,400} & 3 & \textbf{25,000,000} \\
			ATIS ADD \cite{Tournemire.23012020} & 2020 & 2D Detection & Vehicle & 304 $\times$ 240 & X & -$^\mathrm{*}$ & \underline{1,404,000} & 2 & 255,781 \\
			DSEC-Det. \cite{Gehrig.2021, Gehrig.22112022, DSEC-Detection.27022024} & 2023 & 2D Detection & Vehicle & 640 $\times$ 480 & \textbf{X} & \textbf{X} & 70,379 & \textbf{8} & \underline{390,118} \\
			\hline
			TUMTraf Event (Ours) & 2024 & 2D Detection & \textbf{Roadside} & 640 $\times$ 480 & \textbf{X} & \textbf{X} & 4,111 & \underline{7} & 50,496 \\
			\hline	
			\multicolumn{10}{l}{$^{\mathrm{*}}$Information was not available.}	
		\end{tabular}
		\label{table:event-based-datasets}
	\end{center}
\end{table*}

\section{Related Work}
\label{section:related_work}
First, we give an overview of event-based cameras and their usage in roadside ITS. Furthermore, we briefly summarize existing calibration algorithms for multi-sensor setups and state-of-the-art detection and fusion between event-based and conventional cameras.

\subsection{Event-based cameras in roadside ITS}
Event-based cameras recognize changes in the brightness of each pixel asynchronously. As explained in our previous work \cite{Cre.642023672023}, which refers to \cite{Gallego.2022}, each pixel responds independently to brightness changes in the continuous log brightness signal $L(\boldsymbol{u_{k}}, t)$. Here, an event $e_{k} = (\boldsymbol{u_{k}}, t_{k}, p_{k})$ is triggered at pixel $\boldsymbol{u_{k}} = (x_{k}, y_{k})^{T}$ at time $t_{k}$ when the brightness change $\Delta L(\boldsymbol{u_{k}}, t_{k})$ since the last event at the same pixel $\Delta t_{k}$ reaches a threshold $C$:

\begin{equation}
	\Delta L(\boldsymbol{u_{k}}, t_{k}) = L(\boldsymbol{u_{k}}, t_{k}) - L(\boldsymbol{u_{k}}, t_{k} - \Delta t_{k}) = p_{k}C,
\end{equation}
where $C > 0$. The event polarity is the sign of the brightness change $p{k} \in \{+1, -1\}$. Visually, we can interpret an event as motion. As a result of this, event-based cameras have a very high temporal resolution and a dynamic range of 140 dB, far more than conventional cameras (60 dB) \cite{Gallego.2022}. So, the perception system of an ITS could benefit from low energy consumption, low latency, and higher detection performance in challenging conditions (night vision, no motion blur from high-speed vehicles) \cite{Chen.2020b}. Because of these advantages, event-based cameras are slowly being established in roadside ITS \cite{Cre.2023}. The authors of \cite{Chen.2018} presented the first and only detection and tracking approach we know, using a stationary event-based camera mounted on roadside infrastructure. The approach contains several clustering (e.g., DBSCAN \cite{EsterMartinandKriegelHansPeterandSanderJorgandXuXiaowei.1996}) and tracking methods (e.g., SORT \cite{Bewley.2016}) for object detection. They achieved sufficient detection performance with more than $110\text{ Hz}$ frame rate. However, the authors noted a lack of datasets in the ITS domain. In this scope, we recognize research gaps in using state-of-the-art CNNs, analysis of performance in different lighting conditions (e.g., day or night), and the fusion with other sensor systems in the domain of roadside ITS.

\subsection{Calibration}
Data fusion between event-based and conventional cameras requires an accurate calibration. We can generally distinguish between target-based (e.g., with a checkerboard pattern) and targetless methods. \cite{Muglikar.2021, Huang.2021} used a classical checkerboard, which was moved in front of the camera. \cite{Q.Zhang.2020} also calibrated an event-based camera with a checkerboard, whose lighting was changed using a flashlight to trigger events. Furthermore, \cite{DominguezMorales.2019, E.Mueggler.2015} apply a flashing LED grid pattern and \cite{Plasberg.742022772022, E.Mueggler.2014} apply a flashing screen with a shown checkerboard. The usage of these targets in the domain of roadside ITS is impracticable. Consequently, a targetless method is required.     

In general, extrinsic calibration algorithms for multi-sensor systems are based on the principle of registering similarities. Due to thermal factors, wind gusts, or other uncontrolled movements, \cite{Simarro.2021} performed a multi-camera autocalibration to improve the calibration accuracy during runtime. \cite{Fu.28092021} calibrated targetless cameras, thermal cameras, and laser sensors, based on key points extraction in the images using SIFT \cite{Lowe.2004} and point cloud registration between the modalities using ICP \cite{Besl.1992}. The extrinsic calibration approach between camera and Lidar of \cite{Tsaregorodtsev.08082022} used the assignment of the point cloud segmentation map and the semantic segmentation map of the camera image. Another calibration approach for camera and Lidar in an unstructured environment using Kalman Filter was developed by \cite{MunozBanon.2020}. The authors \cite{Wang.2021} performed extrinsic targetless calibration between a stationary camera and Radar based on point cloud clustering with DBSCAN \cite{EsterMartinandKriegelHansPeterandSanderJorgandXuXiaowei.1996}. Each Radar object cluster is tracked first, then assigned with the camera detections. Similar to this, \cite{Sengupta.2022b} also have developed calibration for camera and Radar using the Hungarian Method to find an association between the Radar clusters, generated via DBSCAN \cite{EsterMartinandKriegelHansPeterandSanderJorgandXuXiaowei.1996}, and the camera bounding boxes, generated via YoloV3 \cite{Redmon.2016}. Interestingly, \cite{Sengupta.2022b} provided an automatically labeled Radar dataset based on the RGB detections.

In our previous work \cite{Cre.642023672023}, we developed a targetless calibration method. In this work, we noticed that established image registration methods (e.g., SIFT \cite{Lowe.2004}) cannot be applied to event-based images. Therefore, we extracted the edges of moving objects and registered them to each other. Our approach provided sufficient calibration accuracy if the cameras captured the same object. The main limitation was the robustness against disturbances (e.g., shadow) or multiple moving objects not equally imaged by both cameras. In these cases, the algorithm was unable to produce accurate results. To the best of our knowledge, we are unaware of other targetless calibration methods for stationary event-based cameras.

\subsection{Detection}
Object detection with event-based cameras can be realized using an unsupervised learning method (e.g., clustering), spiking neural networks (SNNs), or convolutional neural networks (CNNs). Here, the weights of the neural networks can be initialized with pre-trained knowledge from a different domain \cite{Chen.2020b}. Based on these methods, robust object detection or several fusion approaches can be implemented. As already mentioned, \cite{Chen.2018} used several clustering methods (e.g., DBSCAN \cite{EsterMartinandKriegelHansPeterandSanderJorgandXuXiaowei.1996}) for object detection with event-based cameras. In Contrast, \cite{Chen.2019} applied a YoloV3 \cite{Redmon.2016} object detector for pedestrian detection with a stationary event-based camera. 

A primary factor for high-quality detections is a broad dataset. Therefore, \cite{Binas.04112017, Mueggler.2017b, Sironi.2018, Zhu.2018b, Cheng.2019, Hu.18052020, Perot.2020, Tournemire.23012020, Gehrig.2021, Gehrig.22112022, DSEC-Detection.27022024} provided data from an event-based camera from an ego-motion perspective (e.g., recorded from a vehicle). Table \ref{table:event-based-datasets} gives an overview of the datasets mentioned and shows their purposes and properties. Unfortunately, there are significant differences in the data between an event-based camera with ego-motion and a stationary-mounted event-based camera on a roadside ITS: In addition to the difference in the perspective, the representation of non-moving objects is logically not displayed in stationary event-based cameras. Although many datasets have been published, there is still a lack of labeled event-based camera datasets from a stationary roadside perspective \cite{Chen.2019, Chen.2020b}. To tackle this problem, the simulators \cite{HenriRebecq.2018b, Mueggler.2017b} could create synthetic datasets. However, particularly for event-based cameras, \cite{Tomy.2022} mentioned the existing sim2real gap. Another way to get around the lack of labeled datasets is the approach of \cite{Chen.2018b}, which is generated in a multi-sensor setup using pseudo-labels. With this, the authors transformed the detections of an RGB camera into the domain of an event-based camera and used them as labels. This approach promises sufficient results for training CNNs with event-based images and is also used in our work. 

\begin{figure*}[!t]
	\centerline{\includegraphics[width=\linewidth]{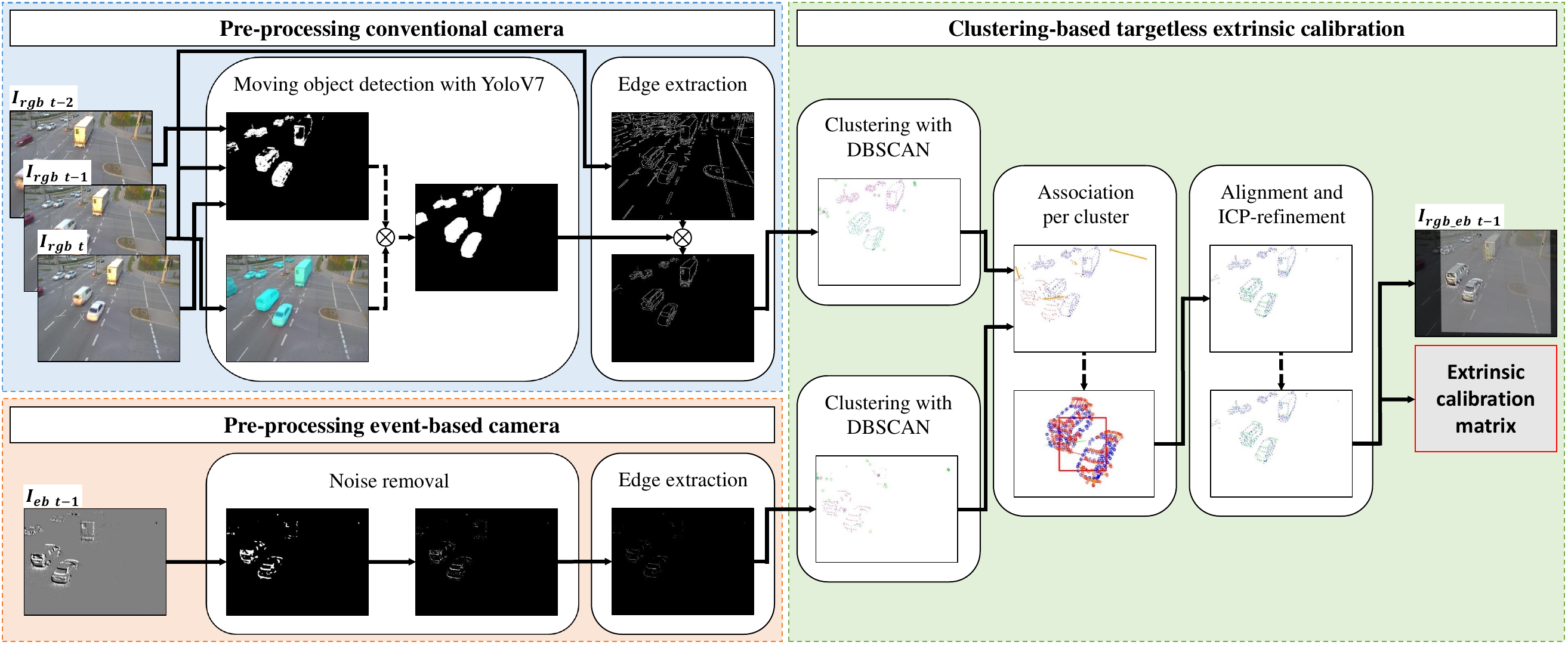}}
	\caption{The main components of our targetless extrinsic calibration algorithm are ``pre-processing conventional camera,'' ``pre-processing event-based camera,'' and ``clustering-based targetless extrinsic calibration.'' The event-based camera indicates moving image regions. However, we identify such areas in the RGB camera by analyzing the last three images. We extended our previous work \cite{Cre.642023672023} with DBSCAN \cite{EsterMartinandKriegelHansPeterandSanderJorgandXuXiaowei.1996} and can now handle multiple moving objects. The fundamental goal is to find associations per cluster pair to calculate a global transformation matrix. This approach allows us to calibrate in more complex traffic scenarios.}
	\label{fig:methodology_calibration}
\end{figure*}

\subsection{Fusion}
In addition to the previously mentioned strengths of event-based cameras, there are limitations, such as the lack of color and texture information \cite{Cao.2021, Zhang.61720236242023}. Nevertheless, these systems can complement frame-based RGB cameras with intelligent data fusion \cite{Tomy.2022}. However, compared to the fusion of Radar, camera, and Lidar, the fusion with event-based cameras is a relatively nascent field \cite{Tomy.2022}. According to \cite{Brenner.2023}, we can distinct between the levels ``Data level (early) fusion,'' ``Feature level (middle) fusion,'' and ``Decision level (late) fusion.'' On an early level, raw data is fused. This approach is useful when the data of the different modalities are comparable and compatible. Fusion on the feature level extracts features from each modality and combines them before being fed into a classifier. The output can be used in learning algorithms. In contrast, late fusion considers the detections of each sensor modality and combines them to produce a final decision. This optimal assignment between the object detections can, e.g., be implemented using a modified Jonker-Volgenant algorithm \cite{Crouse.2016}, as in the late fusion for camera and Lidar of \cite{Zimmer.642023672023}. 

Several data fusion approaches for cameras have already been developed. First, we want to highlight the fusion approach IFCNN \cite{Zhang.2020}. This approach uses a convolutional neural network to extract distinctive image features of several image modalities and fuse them by an appropriate fusion rule. Finally, the fused features are reconstructed by two convolutional layers to produce the informative fusion image. This fusion approach is particularly noteworthy because of its generalizability: In their experiments, the authors have achieved impressive results in fusing various types of images, such as multi-focus, infrared-visual, multi-modal medical, and multi-exposure images. The authors \cite{Cao.2021, Zhang.61720236242023, Zhou.2023, Tomy.2022} also fused the separately extracted features of event-based and RGB images generated from a perspective with ego-motion. Here, \cite{Tomy.2022} chose a voxel grid to represent event-based data and extracted their features. After applying the homography transformation from event-based to the RGB camera, the features were fused and fed as input to a RetinaNet \cite{Lin.2017} for object detection. The dataset used was also recorded from a moving vehicle. In addition, \cite{Tomy.2022} experimented with image reconstruction using the method proposed in \cite{Rebecq.2021} with the result of high computation costs. They also performed an early fusion and detection by combining the RGB and event voxels. Last but not least, a late fusion between event-based and RGB cameras was developed by \cite{Sun.2023b}. The authors used DBSCAN \cite{EsterMartinandKriegelHansPeterandSanderJorgandXuXiaowei.1996} clustering as a detector for the event-based camera and RetinaNet \cite{Lin.2017} for the RGB camera. Nevertheless, there is still a lack of knowledge about the detection performance of early and late fusion between a stationary event-based and RGB camera in the domain of ITS with real data under adverse weather conditions.

\section{Methodology}
\label{section:methodology}
In this section, we describe our targetless extrinsic calibration approach, which is an essential improvement of our previous work \cite{Cre.642023672023}. Furthermore, we present our method to generate a synchronized event-based and conventional RGB camera dataset, as well as the object detector that we used to train on our dataset. Lastly, we close this section with our developed Early Fusion, Simple Late Fusion (SLF), and novel Spatiotemporal Late Fusion (STLF) algorithms between event-based and RGB cameras optimized for stationary usage at a roadside ITS. 

\subsection{Targetless calibration}
\label{subsection:targetless-calibration}
Figure \ref{fig:methodology_calibration} gives an overview of our extrinsic calibration approach based on multiple objects. As mentioned in our previous work, the image content of event-based cameras is indicated by motion in the scene. The optical flow represents the motion in images from conventional cameras. For calculating the extrinsic calibration matrix, this approach aims to accurately match the detected motion between event-based and RGB cameras to find image correspondences between both sensor modalities. The calibration algorithm can be divided into ``Pre-processing conventional camera,'' ``Pre-processing event-based camera,'' and ``Clustering-based targetless extrinsic calibration.'' With this cluster analysis, we can tackle the main weakness of our previous work \cite{Cre.642023672023} that only one dynamic object can be in the field of view of all cameras. The other assumptions are still valid: We need a time-synchronized event-based and conventional camera setup for correct targetless calibration that recognizes the same objects from almost the same perspective. In addition, the objects must have a sufficient distance from the camera to be considered as a planar plane.

In the first step, as in our previous work \cite{Cre.642023672023}, our algorithm accurately extracts the edges of moving objects in the event-based camera. This grayscale image contains the brightness changes accumulated over the last $\SI{5000}{\micro\second}$. White areas indicate event polarity of $+1$, and black areas of $-1$. Gray areas show no motion. For simplification, we ignore the polarity of the events and, therefore, convert the image into a black-and-white image: Black defines static image areas and white dynamic image areas. In contrast to \cite{Cre.642023672023}, we directly apply a dilation operation on the image with a kernel size of $ksize = 3 \times 3$. Then, a median filter with $ksize = 3$ removes noise from the binary image. The faster an object moves, the larger the white area in the processed event-based image. As previously shown in \cite{Cre.642023672023}, to enhance an edge image $\boldsymbol{E} \in\mathbb{R}^{2}$, we apply efficient morphological hit-miss operations using a combination of structuring elements (kernels $\boldsymbol{K} \in\mathbb{R}^{3}$) in vertical, horizontal and diagonal directions, as follows:
\begin{equation}
	\begin{split}
		\boldsymbol{K_{verti}} = \left[ \begin{array}{rrr}
			0 & 1 & 0 \\ 
			0 & 1 & -1 \\
			0 & 1 & 0 \\ 
		\end{array}\right],
		\boldsymbol{K_{horiz}} = \left[ \begin{array}{rrr}
			0 & -1 & 0 \\ 
			1 & 1 & 1 \\
			0 & 0 & 0 \\ 
		\end{array}\right],
		\\
		\boldsymbol{K_{diag1}} =\left[ \begin{array}{rrr}
			0 & -1 & 1 \\ 
			0 & 1 & 0 \\
			1 & 0 & 0 \\ 
		\end{array}\right],
		\boldsymbol{K_{diag2}} = \left[ \begin{array}{rrr}
			1 & -1 & 0 \\ 
			0 & 1 & 0 \\
			0 & 0 & 1 \\ 
		\end{array}\right]. 
	\end{split}
\end{equation}
Then, we combine the edge images based on the kernels mentioned, with $\boldsymbol{E} = \boldsymbol{E_{verti}} + \boldsymbol{E_{horiz}} + \boldsymbol{E_{diag1}} + \boldsymbol{E_{diag2}}$.

In the second step, we detect the edges of multiple moving objects in the conventional camera. To enable more accurate motion detection, inspired by \cite{Ross.2007}, unlike \cite{Cre.642023672023}, we extract motion $\boldsymbol{M}_{t} \in\mathbb{R}^{2}$ with the last three grayscale images $\boldsymbol{I}_{t} \in\mathbb{R}^{2}$, $\boldsymbol{I}_{t-1}$, and $\boldsymbol{I}_{t-2}$ and a binary threshold with $T \in\mathbb{Z}$, e.g., $T=10$:
\begin{equation}
	\begin{split}
	\boldsymbol{D1_{t}} = | \boldsymbol{I_{t}} - \boldsymbol{I_{t-1}} |, \hspace{41pt}
	\boldsymbol{D2_{t}} = | \boldsymbol{I_{t-1}} - \boldsymbol{I_{t-2}} |, \hspace{27pt} \\
	\boldsymbol{T1_{t}}= \begin{cases}
		255 & \text{if } \boldsymbol{D1_{t}} > T  \\ 0 & \text{otherwise}
	\end{cases} ,\hspace{5pt} 
	\boldsymbol{T2_{t}}= \begin{cases}
		255 & \text{if } \boldsymbol{D2_{t}} > T  \\ 0 & \text{otherwise}
	\end{cases} ,\\
	\boldsymbol{M_{t}} = \boldsymbol{T1_{t}} \land \boldsymbol{T2_{t}}. \hspace{39pt} \\
	\end{split}
\end{equation}

Nevertheless, as in \cite{Cre.642023672023}, we only consider motion caused by moving objects, not environmental influences, e.g., camera vibrations due to wind gusts. Therefore, inspired by \cite{Kim.2013}, we also analyze the optical flow based on the methods Good Features To Track by J. Shi and Tomasi \cite{JianboShi.1994} and Lucas-Kanade optical flow in pyramids \cite{J.Bouguet.1999}: A flow vector $\boldsymbol{v} \in\mathbb{R}^{2}$ with a specific length $l$ is assigned to camera motion if 
\begin{equation}
	\boldsymbol{v_{l}} < (m + C),
\end{equation}
with $m$ as the median of the length of all optical flow vectors and $C \in\mathbb{R}$ as a constant value, e.g., $C=0.5$. The other flow vectors indicate motion by moving objects. Furthermore, we also apply a KNN background subtractor \cite{Zivkovic.2006} to receive the motion from the image sequence. To consider the edge extraction of the complete texture of a moving object, we use deep-learning-based instance segmentation provided by YoloV7 \cite{arXiv.org.07072022b, Wong.14012023, Chen.02012020}, pretrained on the MS Coco dataset \cite{Lin.01052014b}. In contrast to our previous work \cite{Cre.642023672023}, we don't just consider the object containing the most movement. Instead, for calculation $\boldsymbol{M_{yolo}}_{t}$, we consider all detected instance segmentation masks, where the ratio of the motion $r_{motion}$ is greater than $C_{motion} \in\mathbb{R}$, e.g. $C_{motion}=0.2$, and the ratio to the total image $r_{total}$ is greater than $C_{total} \in\mathbb{R}$, e.g. $C_{total}=0.002$, as follows:
\begin{equation}
	\begin{split}
		r_{motion} = \frac{m_{i}}{d_{i}}, \hspace{10pt}
		r_{total} = \frac{d_{i}}{S} \text{ for } i = 0, ..., n
	\end{split}
\end{equation}
with the number of pixels $m$ in the motion mask and the number of pixels $d$ in the instance segmentation mask of each detected object $i$. Here, $n$ is the total number of detected objects, and $S$ is the total number of pixels in the image.

\begin{table*}[h!]
	\caption{The roadside perspective TUMTraf Event Dataset is designed for training an event-based and early fusion detector. ``L-EB'' labels are visible for the event-based camera, and ``L-RGB'' labels for the RGB camera. We generated the training and validation labels via pseudo-labeling. The test set was labeled manually to ensure accurate evaluation and was splitted into the subsets ``Day'', ``N-1'' (``night with street lights on''), and ``N-2'' (``night with street lights off''). In total, the dataset consists of 7 classes and 50,496 labels. Unfortunately, we noticed a lack of pedestrians, bicyclists, and motorcycles in our recordings, particularly at night. For the sake of completeness, we have listed these classes anyway.}
	\begin{center}
		\begin{tabular}{|r|r|R{1.0cm}|R{1.0cm}|R{1.0cm}|R{1.0cm}|R{1.0cm}|R{1.0cm}|R{1.0cm}|R{1.0cm}|R{1.0cm}|R{1.0cm}|}
			\hline
			\textbf{ID} & \textbf{Class} & \textbf{Train} & \textbf{Train} & \textbf{Val} & \textbf{Val} & \multicolumn{3}{r|}{\textbf{Test}} & \multicolumn{3}{r|}{\textbf{Test}} \\	
			& & \textbf{L-EB} & \textbf{L-RGB} & \textbf{L-EB} & \textbf{L-RGB} & \multicolumn{3}{r|}{\textbf{L-EB}} & \multicolumn{3}{r|}{\textbf{L-RGB}} \\		
			\cline{3-12}
			& & \multicolumn{2}{r|}{\textbf{Day}} & \multicolumn{2}{r|}{\textbf{Day}} & \textbf{Day} & \textbf{N-1} & \textbf{N-2} & \textbf{Day} & \textbf{N-1} & \textbf{N-2} \\
			\hline
			0 & Pedestrian & 739 & 1,332 & 61 & 328 & 163 & 0 & 0 & \underline{1,196} & 0 & 0 \\
			1 & Bicycle & 17 & 51 & 0 & 0 & 38 & 0 & 0 & 109 & 0 & 0 \\
			2 & Car & \textbf{8,620} & \textbf{18,710} & \textbf{1,624}& \textbf{3,534} & \textbf{1,211} & \textbf{300} & \textbf{52} & \textbf{3,717} & \textbf{620} & \textbf{90} \\
			3 & Motorcycle & 2 & 0 & 0 & 9 & 17 & 0 & 0 & 21 & 0 & 0 \\
			4 & Bus & 352 & 443 & 16 & 18 & 37 & \underline{22} & 0 & 49 & \underline{42} & 0 \\
			5 & Truck & \underline{754} & \underline{2,197} & \underline{127} & \underline{422} & 155 & 2 & 0 & 400 & 2 & 0 \\
			6 & Trailer & 698 & 1,097 & 112 & 399 & \underline{225} & 0 & 0 & 366 & 0 & 0 \\
			\hline
		\end{tabular}
		\label{table:dataset_number_labels}
	\end{center}
\end{table*}

Then, we combine the motion mask $\boldsymbol{M}_{t}$, described above, and the motion mask $\boldsymbol{M_{yolo}}_{t}$ with a logical OR operation. With this procedure, we obtain motion, which includes moving traffic participants and background, e.g., shadows or blowing trees in the wind. To receive the edges of these moving objects, we first apply Canny edge detection \cite{Canny.1986} on the conventional camera image. Second, we combine it with the extracted motion mask via a bitwise AND operation.

At this point, similar edge images, including moving objects from the event-based and conventional camera, are available. To deal with multiple moving objects, we divide the edge images into several clusters using DBSCAN \cite{EsterMartinandKriegelHansPeterandSanderJorgandXuXiaowei.1996}. We want to emphasize the importance of dividing the event-based image into clusters, which has to be similar to the division of the conventional camera. After clustering, we determine the median centroids of each cluster. An optimal assignment between the clusters from event-based and conventional cameras can be found with these positions. For this purpose, the linear sum assignment problem will be solved with a modified Jonker-Volgenant algorithm \cite{Crouse.2016}.

Next, we search for an association inside each cluster pair. For this, we must optimally align the event-based 2D point cloud with the conventional camera 2D point cloud. To be robust against outliers, we create an imaginary rectangle, see the red rectangle in Figure \ref{fig:methodology_calibration}, between the $13^\text{th}$ and $87^\text{th}$ percentile for each event-based respectively conventional camera cluster. With these two rectangles $\boldsymbol{r_{eb}}$ and $\boldsymbol{r_{rgb}}$ for each cluster pair, we can, similar to our previous work \cite{Cre.642023672023}, find a suitable scaling $\boldsymbol{s} \in\mathbb{R}^{2}$ and displacement $\boldsymbol{t} \in\mathbb{R}^{2}$ to transform each cluster of the event-based camera with $\boldsymbol{T_{coarse}} \in\mathbb{R}^{3}$: 
\begin{equation}
	\begin{split}
	    \boldsymbol{r_{rgb}}_w = \boldsymbol{r_{rgb}}_{x_{0.87}} - \boldsymbol{r_{rgb}}_{x_{0.13}}, \\
	    \boldsymbol{r_{rgb}}_h = \boldsymbol{r_{rgb}}_{y_{0.87}} - \boldsymbol{r_{rgb}}_{y_{0.13}}, \hspace{1pt}\\
	    \boldsymbol{r_{eb}}_w = \boldsymbol{r_{eb}}_{x_{0.87}} - \boldsymbol{r_{eb}}_{x_{0.13}}, \hspace{9pt} \\
	    \boldsymbol{r_{eb}}_h = \boldsymbol{r_{eb}}_{y_{0.87}} - \boldsymbol{r_{eb}}_{y_{0.13}}, \hspace{10pt}\\
	    \\
		\boldsymbol{s}_{x} = \frac{\boldsymbol{r_{rgb}}_w}{\boldsymbol{r_{eb}}_w}, \hspace{10pt}
		\boldsymbol{s}_{y} = \frac{\boldsymbol{r_{rgb}}_h}{\boldsymbol{r_{eb}}_h}, \\
		\\
		\boldsymbol{r_{eb_{scaled}}}_{x} = \boldsymbol{s}_{x} \cdot \boldsymbol{r_{eb}}_{x_{0.13}}, \\
		\boldsymbol{r_{eb_{scaled}}}_{y} = \boldsymbol{s}_{y} \cdot \boldsymbol{r_{eb}}_{y_{0.13}}, \\
		\\
		\boldsymbol{t}_{x} = (-1 \cdot \boldsymbol{r_{eb_{scaled}}}_{x}) + \boldsymbol{r_{rgb}}_{x_{0.13}}, \\
		\boldsymbol{t}_{y} = (-1 \cdot \boldsymbol{r_{eb_{scaled}}}_{y}) + \boldsymbol{r_{rgb}}_{y_{0.13}}, \\
		\\	
		\Rightarrow \boldsymbol{T_{coarse}} = \left[ \begin{array}{rrr}
			\boldsymbol{s}_{x} & 0 & \boldsymbol{t}_{x} \\ 
			0 & \boldsymbol{s}_{y} & \boldsymbol{t}_{y} \\
			0 & 0 & 1 \\ 
		\end{array}\right].
	\end{split}
\end{equation}

After translation and scaling, an optimal point-to-point assignment with the modified Jonker-Volgenant algorithm \cite{Crouse.2016} inside of each cluster can be found. To filter outliers from the point assignments, we calculate the length of each point assignment $i$ and determine the length $l$ of the $70^\text{th}$ percentile. If the assignment length $l$ is greater than the length $l$ of the $70^\text{th}$ percentile, we calculate the extension factor $f \in\mathbb{R}$. If the factor $f$ is greater than a threshold $C \in\mathbb{R}$, e.g., $C=0.5$., the point-to-point assignment $i$ is defined as an outlier: 
\begin{equation}
	\begin{split}
		f_{i} = \frac{l_{i} - l_{p=0.70}}{l_{i}}, \hspace{10pt} f_{i} > C.
	\end{split}
\end{equation}

After filtering the outliers, we consider only cluster pairs, where the minimum number of assignments $M \in\mathbb{Z}$, e.g., $M=1$, is achieved. Based on the remaining point-to-point assignments between event-based and conventional cameras, which we determined for each cluster pair, we calculate the coarse alignment as optimal affine transformation using RANSAC \cite{Fischler.1981}. In the last step, we refine our coarse estimation between the point clouds of event-based and conventional cameras using point-to-point ICP \cite{Besl.1992}.

\subsection{Dataset}
\label{subsection:methodology-dataset}
According to \cite{Chen.2019} and Table \ref{table:event-based-datasets}, a lack of event-based camera datasets with a stationary roadside perspective exists. 
\begin{figure}[b!]
	\centerline{\includegraphics[width=\linewidth]{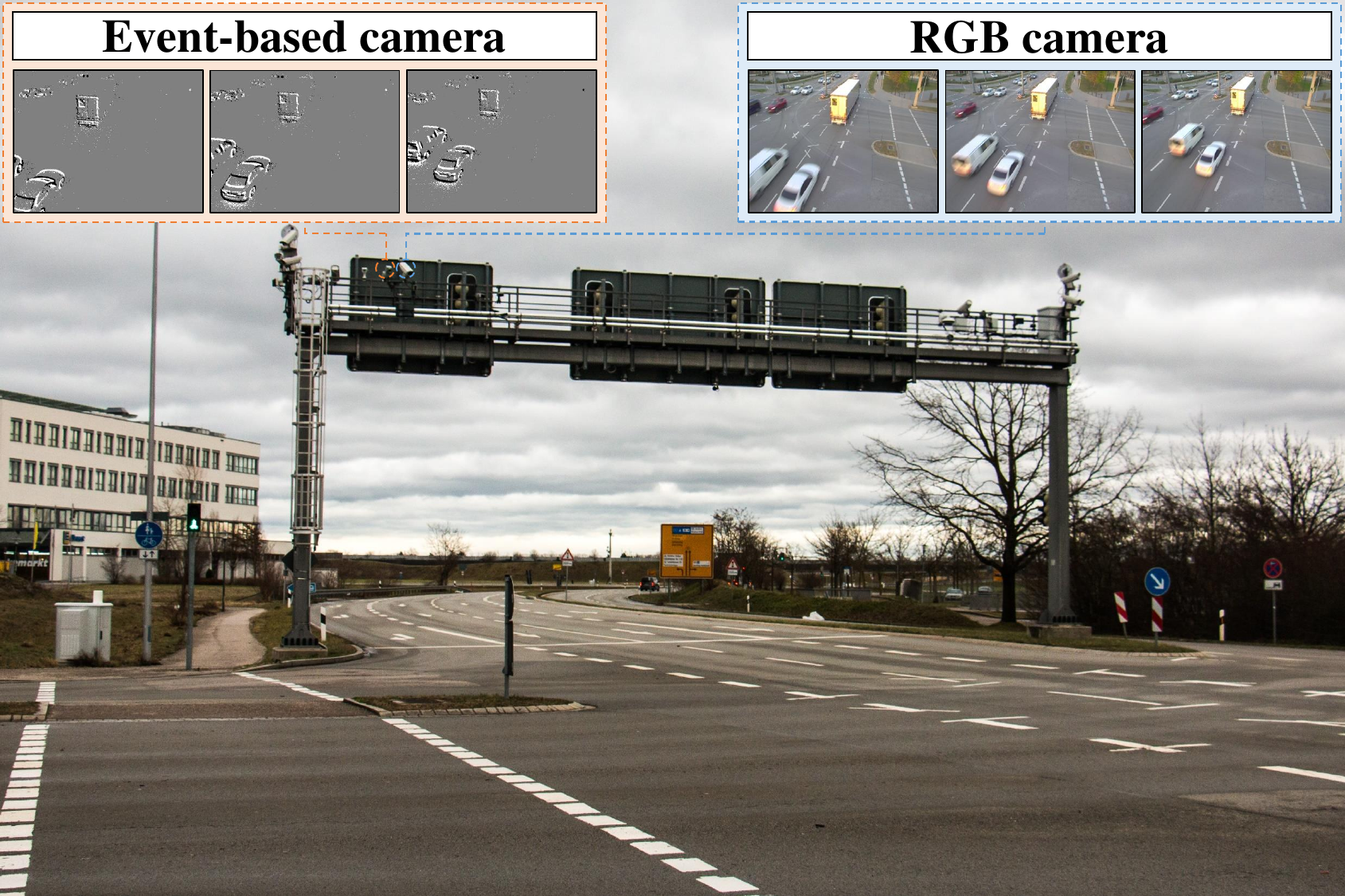}}
	\caption{We recorded the TUMTraf Event Dataset at this intersection in Garching near Munich. Besides event-based and RGB cameras, the gantry contains numerous other sensors, e.g., Lidars, which are the basis for the TUMTraf Dataset family.}
	\label{fig:sensor_setup}
\end{figure}
Therefore, we create our TUMTraf Event Dataset, which includes spatiotemporal synchronized frame triplets of event-based, RGB, and combined RGB-event-based images, see Table \ref{table:dataset_number_labels}. For this, we first record the raw data with our ITS from the Providentia++ project \cite{Krammer.2022, PROVIDENTIAA9TestfeldfurautonomesFahrenunddigitaleErkennungvonFahrzeugen.15022023}, which contains event-based and RGB cameras. As illustrated in Figure \ref{fig:sensor_setup}, the sensors are set up on a gantry at a height of $7$ m located at an intersection in Garching near Munich, Germany. This perspective gives our dataset a unique bird's eye view of the traffic, which minimizes the number of occlusions. The image pairs are recorded time synchronized. The spatial synchronization is achieved using our targetless calibration approach. With it, we create the combined RGB-event-based images with simple blending, see Subsection \ref{subsection:fusion}. The specifications of the sensors used are as follows:
\begin{itemize}
	\item \textbf{RGB camera:} Basler ace acA1920-50gc, $1920 \times 1200$, Sony IMX174, global shutter, color, GigE, with 8 mm lens.
	\item \textbf{Event-based camera:} Imago VisionCam EB, $640 \times 480$, $> 120$ dB dynamic range, 30 000 000 events/s, with 8 mm lens. 
\end{itemize}

To receive accurate detections from the RGB camera, we train the YoloV7 object detector \cite{arXiv.org.07072022b, Wong.14012023, Chen.02012020} on the nuImages dataset \cite{motional.08082023}, which counts $93,000$ images, including rain, snow, and night scenarios from the ego perspective of a vehicle \cite{motional.08082023}. To consider the roadside perspective from a stationary camera on a gantry bridge, we perform transfer learning with the 2D annotations of our TUMTraf Dataset family \cite{Cress.2022, zimmer2023tumtraf}, which also includes snow and night scenarios. With this robust object detector for the conventional camera and the previously calculated extrinsic calibration matrix, we generate pseudo-labels, similar to \cite{Chen.2018b}, using the confidence threshold $C=0.80$. This way, we can obtain any number of accurate pseudo-labels, enabling robust object detection with an event-based camera and early fusion.

We provide two categories of labels: The category ``L-EB'' includes only the labels of moving objects and can be used for the event-based image detector. Here, we automatically analyze each detected object with the optical flow and assign a motion attribute. The other category, ``L-RGB,'' consists of all objects that can, in principle, be recognized by the RGB camera, including moving and non-moving objects. 

To enhance the dataset's quality, we roughly filter the training and validation set by excluding frame triplets where the pseudo-labels are obviously incorrect. This procedure results in an optimized dataset with $2,538$ frame triplets, including synchronized event-based, RGB, and RGB-event-based combined frames for the training set, $580$ frame triplets for the validation set, and $993$ frames for the test set. True to the motto ``Train during the day, detect at night,'' we intentionally select only images from the scenario ``Day'' for the training and validation set. This choice enables us to achieve the best possible quality when generating the pseudo-labels. Image triples from the night scenarios are only used in the test set. This procedure is possible since the event-based camera has the same image content day and night. We also performed meticulous manual fine-tuning of each label in the test set to enable an accurate evaluation. The labels are available in OpenLABEL format \cite{openlabel.31122023}.
\begin{figure}[!b] 
	\centering
	\subfloat[Deactivated anti-flickering filter. \label{fig:eb_raw_flickering}]{%
		\includegraphics[width=0.47\linewidth]{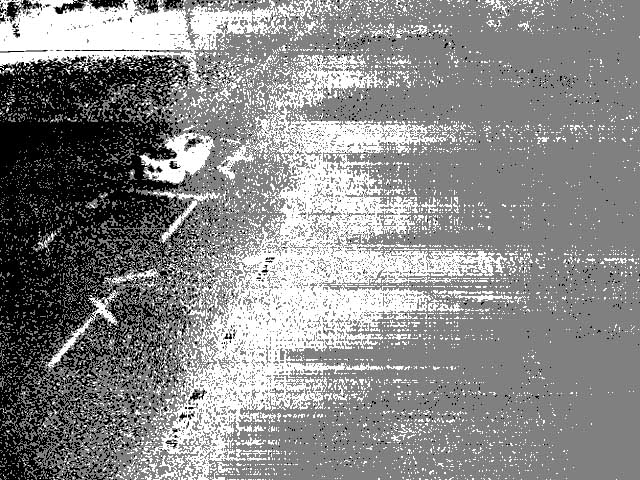}}
	\hfill
	\subfloat[Activated anti-flickering filter.\label{fig:eb_raw}]{%
		\includegraphics[width=0.47\linewidth]{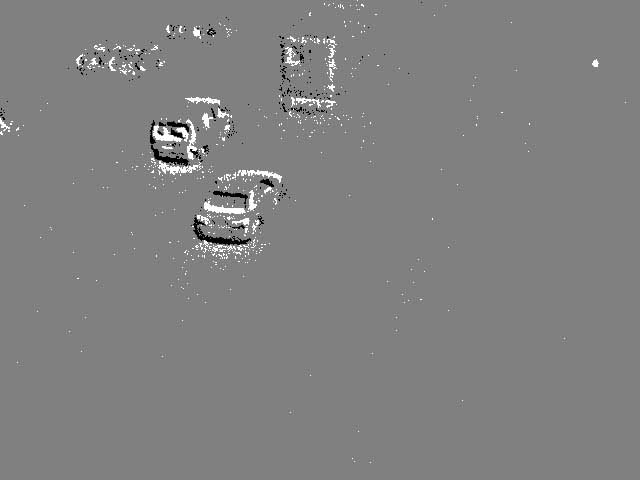}}
	\caption{Street lights cause significant noise at night, which must be eliminated with an anti-flickering filter. This phenomenon is due to the lamps operating with 50 Hz alternating current or pulse width modulation.}
	\label{fig:eb_flickering} 
\end{figure}

\subsection{Detection and fusion}
\label{subsection:fusion}
\begin{figure*}[!h]
	\centerline{\includegraphics[width=\linewidth]{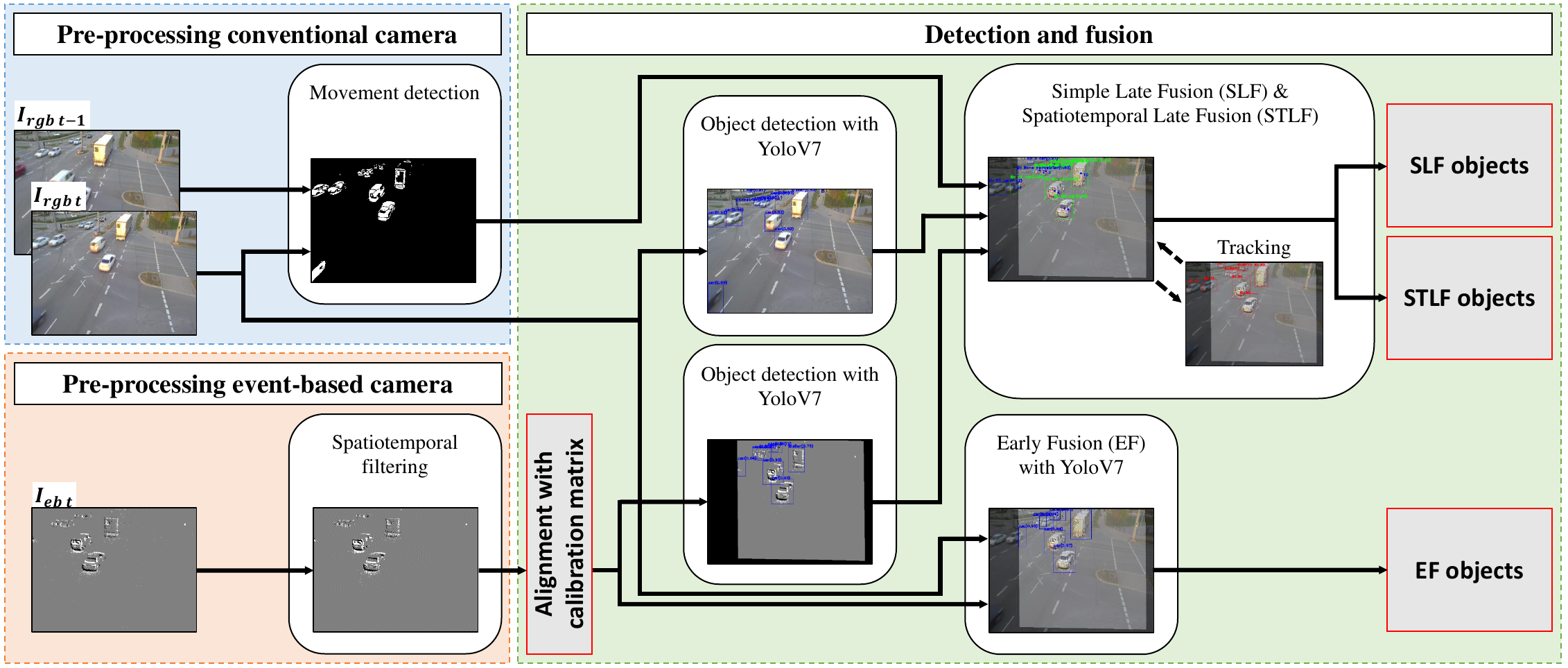}}
	\caption{This figure illustrates the processing pipeline of Early Fusion, Simple Late Fusion, and Spatiotemporal Late Fusion. Early Fusion uses the raw images from the RGB and event-based cameras. On the other hand, the late fusion methods operate on the detections based on the individual images and the motion mask of the RGB camera. In addition, Spatiotemporal Late Fusion utilizes tracking information of each object for its fusion decision.}
	\label{fig:methodology_fusion}
\end{figure*}

This section describes our detection and fusion methods between event-based and conventional cameras. In particular, our methods of Early Fusion (EF), Simple Late Fusion (SLF), and novel Spatiotemporal Late Fusion (STLF) combine the strengths of the two sensor modalities mentioned above. Figure \ref{fig:methodology_fusion} gives an overview of the processing pipeline for our multi-modal image fusion.

As a pre-processing step for the RGB camera, we calculate a motion mask using the grayscaled images $\boldsymbol{I_{t}}$ and $\boldsymbol{I_{t-1}}$. Then, we can determine the per-element absolute difference between both images and we get the motion mask $\boldsymbol{M_{t}}$ by applying a binary threshold function with threshold $T$: 
\begin{equation}
	\begin{split}
		\boldsymbol{D_{t}} = | \boldsymbol{I_{t}} - \boldsymbol{I_{t-1}} |, \\
		\boldsymbol{M_{t}}= \begin{cases}
			255 & \text{if } \boldsymbol{D_{t}} > T  \\ 0 & \text{otherwise}
		\end{cases} 
	\end{split}
\end{equation}
Similar to our calibration approach, we also use a KNN background subtractor \cite{Zivkovic.2006} for refinement. The calculated motion mask allows the fusion component to distinguish between static and moving objects.

Event-based cameras accurately recognize very rapid changes in brightness. If street lights are on, the accumulated grayscale image shows flickering. This phenomenon is due to the lamps operates with $50$ Hz alternating current or pulse width modulation. For this reason, the flickering is removed by the event-based camera driver, see Figure \ref{fig:eb_flickering}. Then, inspired by \cite{Yang.2022}, a spatiotemporal filter is applied as a preprocessing step for data fusion. Here, for each event $e_{k} = (\boldsymbol{u_{k}},t_{k},p_{k})$ at pixel position $\boldsymbol{u_{k}} = (x_{k}, y_{k})^{T}$ at time $t_{k}$, we investigate the spatial-temporal neighborhood $N = (N_{x}, N_{y}, N_{t})$, where
\begin{equation}
	\begin{split}
		N_{x} = [x_{k} - r_{x}, x_{k} + r_{x}], \\
		N_{y} = [y_{k} - r_{y}, y_{k} + r_{y}], \\
		N_{t} = [t_{k} - r_{t}, t_{k}],
	\end{split}
\end{equation}
and $r_{x}, r_{y}, r_{k}$ are the sizes of the neighborhood. The amount of events in the neighborhood $N$ is defined as noise if the total number of events in the neighborhood does not achieve the threshold $E \in\mathbb{Z}$, e.g. $E=30$. This efficient noise suppression ensures that only moving objects are in the image of the event-based camera.

The next step after pre-processing the event-based and conventional camera is the detection and fusion. Here, we develop the methods of Early Fusion, Simple Late Fusion, and a novel Spatiotemporal Late Fusion. For Early Fusion, we simply blend the event-based image $\boldsymbol{I_{eb}}$ and the RGB image $\boldsymbol{I_{rgb}}$ with
\begin{equation}
	\begin{split}
		\boldsymbol{I_{eb\_rgb}} = (1 - \alpha)\boldsymbol{I_{eb}} + \alpha \boldsymbol{I_{rgb}},
	\end{split}
\end{equation}
and $\alpha \in\mathbb{R}$, e.g., $\alpha=0.5$. Then, we train with the label categories ``L-EB'' and ``L-RGB'' the YoloV7 detector \cite{arXiv.org.07072022b, Wong.14012023, Chen.02012020} on the fused images $\boldsymbol{I_{eb\_rgb}}$.

The YoloV7 detector\cite{arXiv.org.07072022b, Wong.14012023, Chen.02012020} for the RGB camera, trained with ``L-RGB,'' is applied to the RGB images for the SLF and STLF. Analogous to this, the detector for the event-based camera, trained with ``L-EB,'' is applied to the aligned event-based images. In the second step, we use the previously calculated motion mask $\boldsymbol{M_{t}}$ and, thus, we find an optimal assignment using the modified Jonker-Volgenant algorithm \cite{Crouse.2016} between objects detected by the event-based camera and moving objects detected by the RGB camera. If the Euclidean distance between the objects of each pair is greater than a threshold $L \in\mathbb{R}$, e.g., $L=50.0$, we reject the assignment. Otherwise, we create a fused object $O_{f}$ and declare it as the output object of the fusion component. In principle, the RGB camera provides more texture information and is more precise in determining the object class. Consequently, the fused object receives the following properties with the weight $\alpha \in\mathbb{R}$, e.g., $\alpha=0.4$:
\begin{equation}
	\begin{split}
		O_{f_{\text{class}}} = O_{rgb_{\text{class}}} \hspace{92pt} \\
		O_{f_{\text{position}}} = (1 - \alpha) O_{rgb_{\text{position}}} + \alpha O_{eb_{\text{position}}}, \\
		O_{f_{\text{size}}} = (1 - \alpha) O_{rgb_{\text{size}}} + \alpha O_{eb_{\text{size}}}. \hspace{17pt}
	\end{split}
\end{equation}

In the third step, we take all fused and unfused objects of the event-based and RGB camera as output objects for the Simple Late Fusion so that moving and non-moving objects are considered. Here, we noticed that in difficult visibility conditions, e.g., night, the false positive rate for the RGB camera detector increases noticeably, even with confidence thresholds $C=0.70$. Therefore, the YoloV7 detector \cite{arXiv.org.07072022b, Wong.14012023, Chen.02012020} for the RGB images has to operate with a relatively high confidence threshold. 

On the other hand, due to the noise-free grayscale mask, the detector of the event-based camera produces significantly fewer to no false positives, even in night images with confidence threshold $C=0.30$, see Figure \ref{fig:qualitative_fusion}. For this reason, we develop a novel Spatiotemporal Late Fusion, where we track each fused object and potential output object with SORT \cite{Bewley.2016} and assign a unique tracking ID. So, we can identify these objects in multiple frames over time. If an object was previously detected by the event-based camera or by a combination of event-based and RGB cameras, we classify this object as trustworthy. Subsequently, only objects from the RGB camera with a confidence threshold, e.g., greater than $C=0.77$, or trustworthy objects are considered by STLF as output objects. This method allows us to significantly reduce the number of false positives.

\section{Evaluation}
\label{section:evaluation}
In this section, we present the results of our improved targetless extrinsic calibration method based on our previous work \cite{Cre.642023672023}. In addition, we perform an ablation study with the DBSCAN \cite{EsterMartinandKriegelHansPeterandSanderJorgandXuXiaowei.1996} algorithm to discuss suitable parameters for a successful targetless calibration. Furthermore, we evaluate the performance of the event-based object detector based on our TUMTraf Event Dataset. Here, we compare our results (EB) to an event-based object detector in the use case of roadside event-based cameras, which we trained on the famous DSEC-Detection Dataset \cite{Gehrig.2021, Gehrig.22112022, DSEC-Detection.27022024} (EB-DSEC). Lastly, we analyze in several traffic scenarios the strengths and limitations of our presented fusion approaches: early fusion based on ``L-EB'' labels (EF-1), early fusion based on ``L-RGB'' labels (EF-2), Simple Late Fusion (SLF), and Spatiotemporal Late Fusion (STLF). To classify these results, we carry out comparisons with the image fusion framework based on convolutional neural network (IFCNN) \cite{Zhang.2020}, and an early fusion based on the DSEC-Detection Dataset \cite{Gehrig.2021, Gehrig.22112022, DSEC-Detection.27022024} (EF-DSEC). As a further ablation study, we also examined the ``Confidence Threshold'' hyperparameter for the method STLF.

\subsection{Extrinsic Calibration}
The main advantage of our improved targetless calibration between event-based and RGB cameras is the ability to handle multiple moving objects. We calculated the reprojection error based on ground truth data to evaluate the accuracy, which we manually created with the same tool as in our previous work \cite{Cre.642023672023}. For comparability to our previous work, we used the identical test sequences 1--3: sequences 1 and 3 contain a single moving car, and sequence 2 includes a crossing van with very slow oncoming traffic. We recorded these sequences with the same cameras as in the TUMTraf Event Dataset; however, we used a $16$ mm lens in our previous work. Furthermore, Sequence 4, which is part of the test set in the TUMTraf Event Dataset, includes bi-directional fast-moving traffic participants with different vehicle classes. We want to mention that our previous approach, fortunately, did not detect the slow movement of oncoming traffic in most frames in Sequence 2. Once multiple motions were detected, our previous approach could no longer determine a meaningful transformation matrix due to its coarse alignment procedure.   

\begin{table}[t]
	\caption{We measured per frame the accuracy of our calculated extrinsic calibration and the manually created ground truth using the reprojection error in pixels. The sequences 1--3 are from our previous work \cite{Cre.642023672023}. Sequence 4 is part of the TUMTraf Event Dataset and contains numerous moving objects. Compared to our previous work, we achieved similar accuracy in all sequences, including the complex traffic scenarios.}
	\begin{center}
		\begin{tabular}{|r|R{1.5cm}|R{1.5cm}|R{1.5cm}|R{1.5cm}|}
			\hline
			& \multicolumn{4}{c|}{\textbf{Rep. Error (Rep. Error GT)}} \\
			\hline
			\textbf{\#} & \textbf{Sequence 1} & \textbf{Sequence 2} & \textbf{Sequence 3} & \textbf{Sequence 4} \\
			\hline
			1 & 10.16 (\textbf{1.15}) & \underline{8.69} (3.30) & \textbf{4.76} (1.55) & 17.41 (\underline{1.62}) \\
			2 & 10.76 (1.97) & 11.09 (\underline{2.09}) & \underline{6.42} (\underline{1.42}) & 15.60 (\textbf{1.36}) \\
			3 & \textbf{3.37} (1.65) & 14.49 (\textbf{1.73}) & 8.20 (1.99) & 7.79 (1.90) \\
			4 & 13.62 (2.76) & \textbf{8.67} (5.16) & 6.69 (\textbf{1.11}) & \textbf{6.72} (2.54) \\
			5 & 6.05 (\underline{1.57}) & 13.51 (2.14) & 12.49 (1.84) & \underline{7.70} (2.84) \\
			6 & \underline{5.41} (1.79) & 12.69 (2.17) & 7.85 (1.69) &  8.18 (3.40) \\
			\hline
		\end{tabular}
		\label{table:quantitative_calibration_evaluation}
	\end{center}
\end{table}

Table \ref{table:quantitative_calibration_evaluation} shows the reprojection error of our extrinsic calibration and the error of the ground truth data in sequences 1--4. The results are roughly comparable to our previous work: We achieved in sequences 2 and 3 with a single moving vehicle a reprojection error of up to $3.37$ px. In the more challenging Sequence 4, which contains several small, independently moving objects, we achieved a reprojection error of up to $6.72$ px. Figure \ref{fig:qualitative_calibration} shows the qualitative effectiveness of our targetless calibration, even with independently moving objects in complex traffic scenarios. Therefore, our improvement increases the flexibility of our previous work \cite{Cre.642023672023}.

The DBSCAN \cite{EsterMartinandKriegelHansPeterandSanderJorgandXuXiaowei.1996} algorithm finds clusters based on the spatial density of a dataset and is essential for a successful targetless calibration. Its input parameters are the maximum allowed distance $\epsilon$ between two data samples of a cluster and the minimum number of necessary data samples $s_{min}$ to create a cluster. For our use case, the defined clusters from the event-based camera must correspond to those from the RGB camera. DBSCAN \cite{EsterMartinandKriegelHansPeterandSanderJorgandXuXiaowei.1996} is ideal for this task because we don't have to specify the total number of clusters, and second, the algorithm is quite robust against outliers. An RGB camera offers significantly more texture information depending on the object than an event-based camera. Therefore, the generated clusters can differ significantly between both sensor modalities, even with the same parameters $\epsilon$ and $s_{min}$. Here, we noticed that parameter $s_{min}$ helps classify disturbing artifacts from the edge image (e.g., shadows) as noise, thus excluding them. In addition, parameter $\epsilon$ should be set depending on the area the moving object occupies in the image. If there is only one moving object, the parameter can also be set as high as possible. Figure \ref{fig:dbscan_ablation_study} shows three examples where we performed DBSCAN \cite{EsterMartinandKriegelHansPeterandSanderJorgandXuXiaowei.1996} in each example with two settings $S_1$ and $S_2$. We defined the settings as follows:  
\begin{equation}
	\begin{split}
	S_1 = \{ \epsilon^{eb} = 150, {s_{min}}^{eb} = 2, \hspace{16pt}\\
	\epsilon^{rgb} = 150, {s_{min}}^{rgb} = 2 \}, 
	\end{split}
\end{equation}
\begin{equation}
	\begin{split}
	S_2 = \{ \epsilon^{eb} = 70, {s_{min}}^{eb} = 2, \hspace{20pt}\\
	\epsilon^{rgb} = 40, {s_{min}}^{rgb} = 2 \}. \hspace{4pt}
	\end{split}
\end{equation}

As seen in examples 1--3, we achieved convincing calibration results if the clusters of the event-based and RGB cameras correspond. It is crucial to ensure no significant deviations, as in Example 2, $S_2$: Here, the DBSCAN \cite{EsterMartinandKriegelHansPeterandSanderJorgandXuXiaowei.1996} clustering divides the vehicle in the RGB image into two clusters, but the clustering in the event-based image represents the vehicle in only one cluster. As the targetless calibration algorithm scales the associated clusters to a uniform size, meaningful cluster matching is impossible in those cases. 

All in all, we used the DBSCAN \cite{EsterMartinandKriegelHansPeterandSanderJorgandXuXiaowei.1996} setting $S_2$ to calibrate the event-based and RGB images from the TUMTraf Event Dataset. Since the result of Sequence 4, frame \#3, with a reprojection error of $7.79$ px, provides the subjectively best extrinsic calibration result, all further data fusion experiments were carried out with this calculated transformation. The intrinsic parameters and the distortion models of the event-based and RGB cameras have been calibrated beforehand.

\begingroup
\begin{figure*}[h!]
	\centering
	\setlength{\tabcolsep}{2pt} 
	\fboxsep=0.001pt
	\fboxrule=0.05pt
	\begin{tabular}{rcccccccc}
		& Seq. 1 - \#2 & Seq. 1 - \#3 & Seq. 2 - \#2 & Seq. 2 - \#3 & Seq. 3 - \#1 & Seq. 3 - \#2 & Seq. 4 - \#2 & Seq. 4 - \#3 \\
		\rot{\hspace{13pt}RGB} & 
		\includegraphics[width=2.08cm]{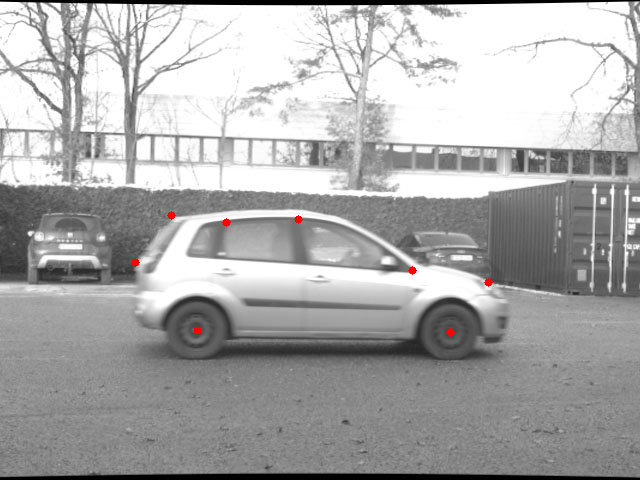} &
		\includegraphics[width=2.08cm]{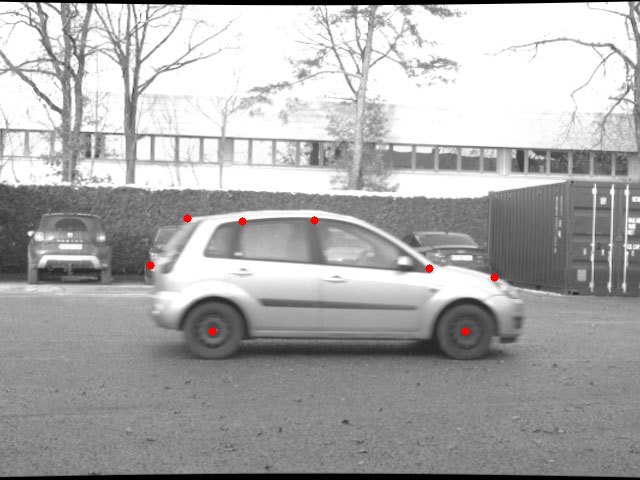} &
		\includegraphics[width=2.08cm]{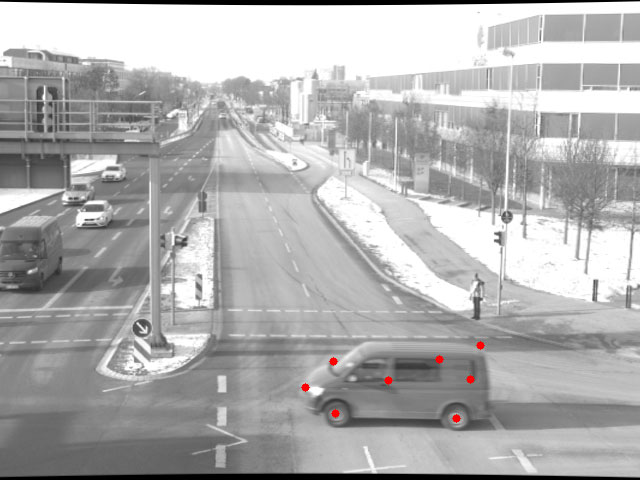} &
		\includegraphics[width=2.08cm]{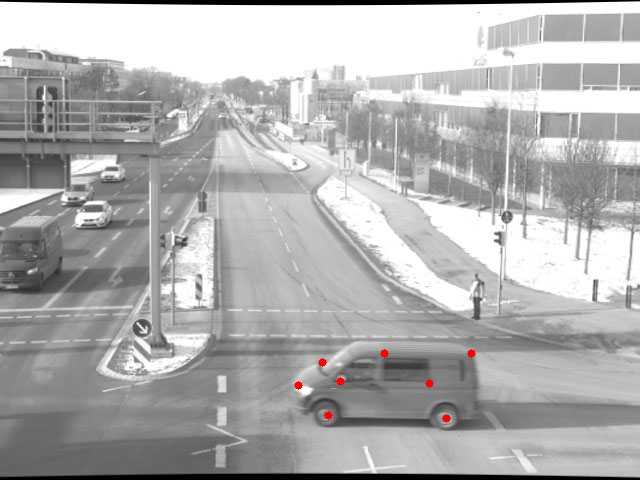} &
		\includegraphics[width=2.08cm]{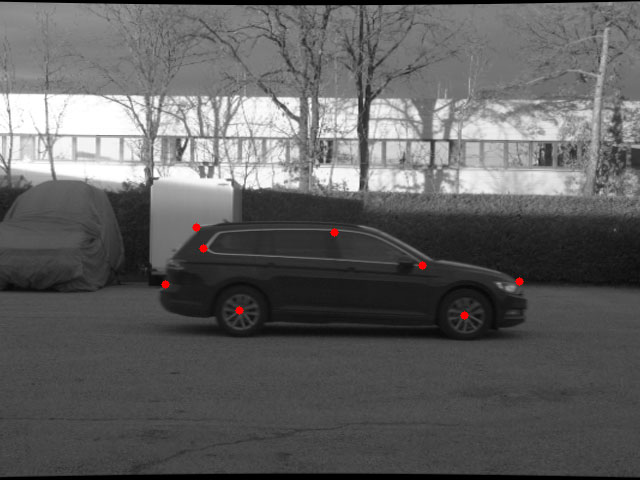} &
		\includegraphics[width=2.08cm]{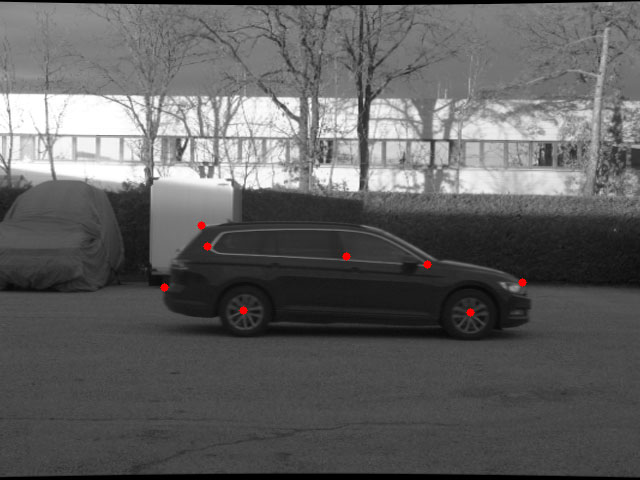} &
		\includegraphics[width=2.08cm]{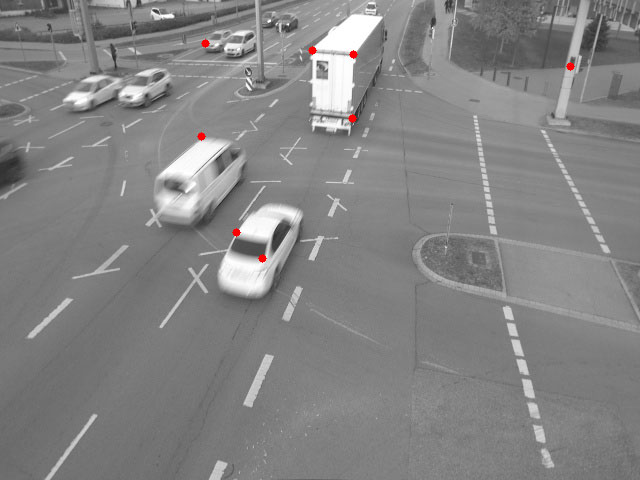} &
		\includegraphics[width=2.08cm]{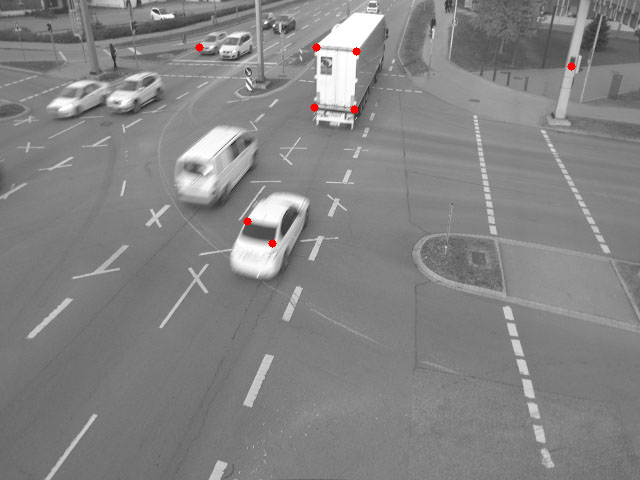} \\
		\rot{\hspace{16pt}EB} & 
		\includegraphics[width=2.08cm]{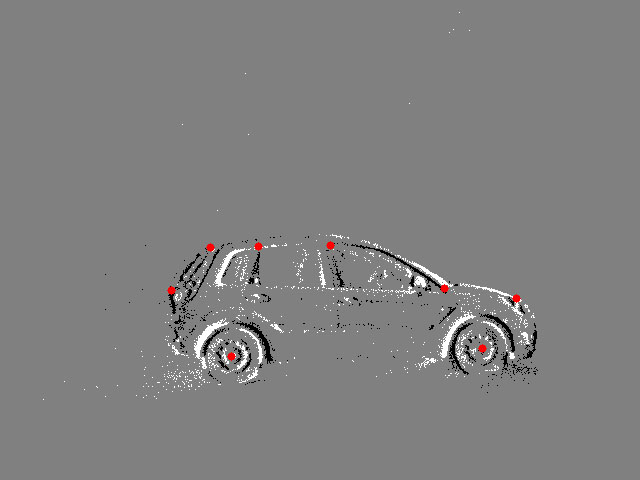} &
		\includegraphics[width=2.08cm]{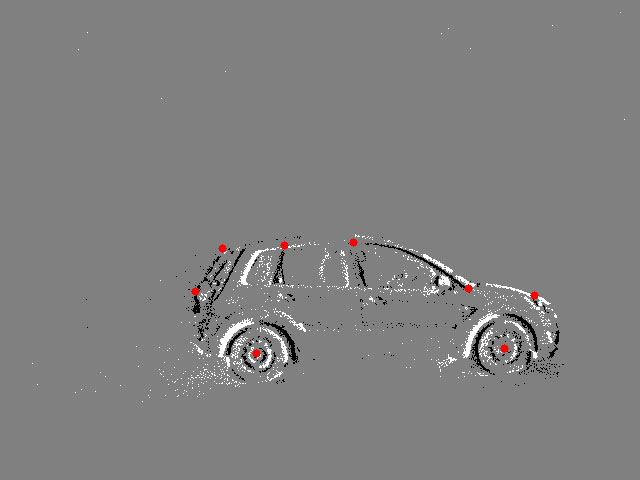} &
		\includegraphics[width=2.08cm]{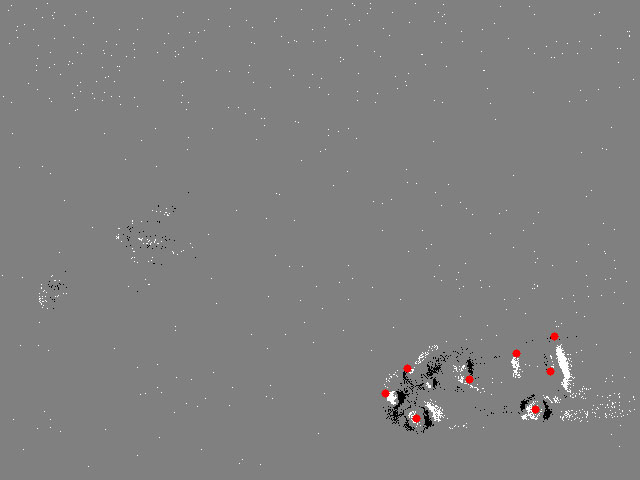} &
		\includegraphics[width=2.08cm]{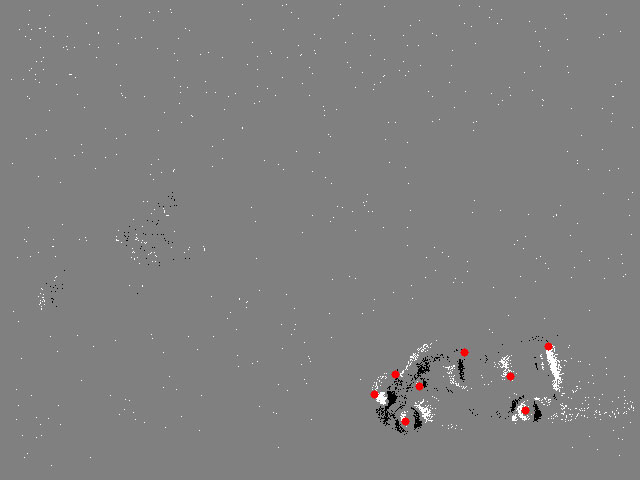} &
		\includegraphics[width=2.08cm]{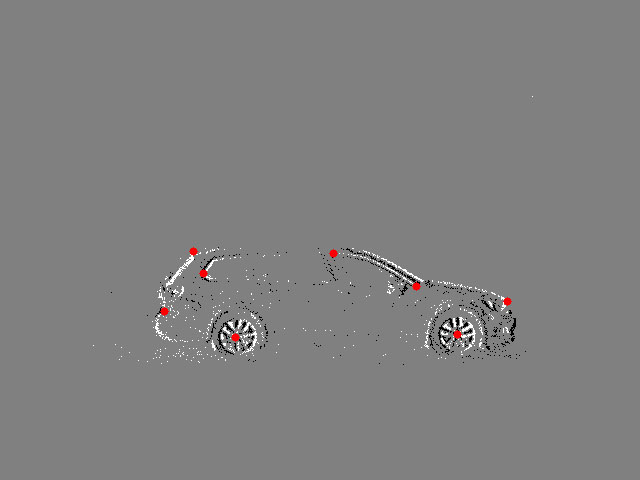} &
		\includegraphics[width=2.08cm]{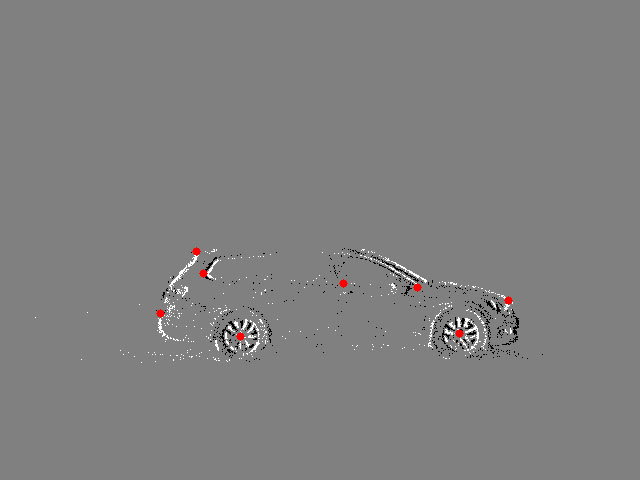} &
		\includegraphics[width=2.08cm]{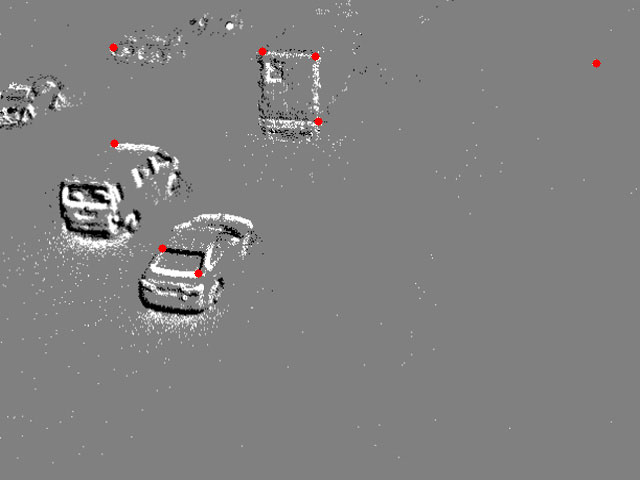} &
		\includegraphics[width=2.08cm]{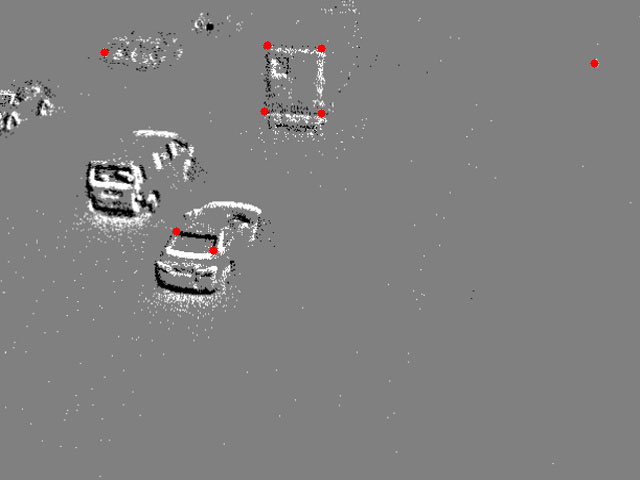} \\
		\rot{\hspace{16pt}GT} & 
		\includegraphics[width=2.08cm]{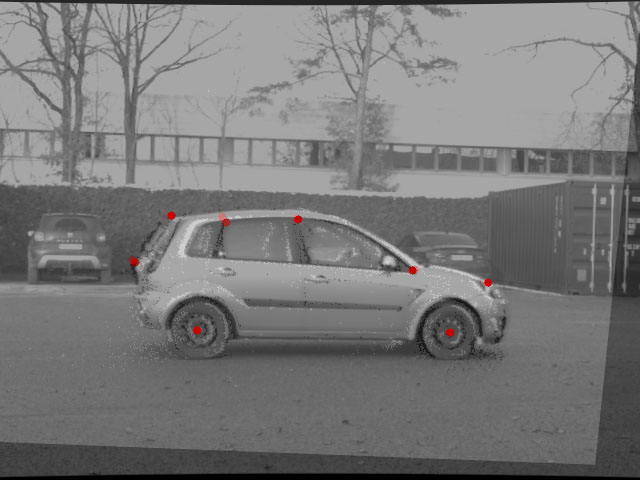} &
		\includegraphics[width=2.08cm]{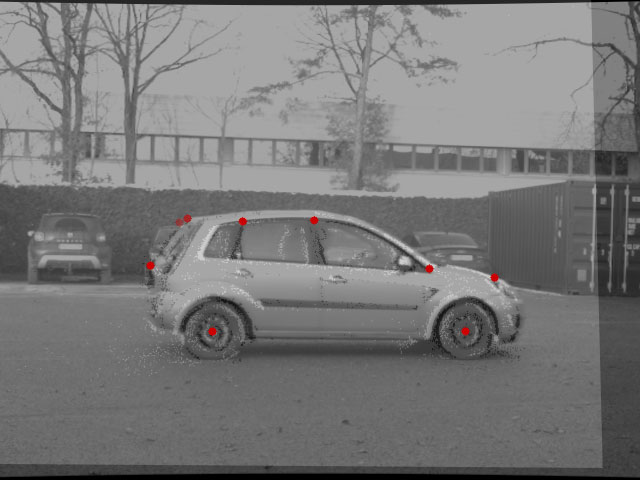} &
		\includegraphics[width=2.08cm]{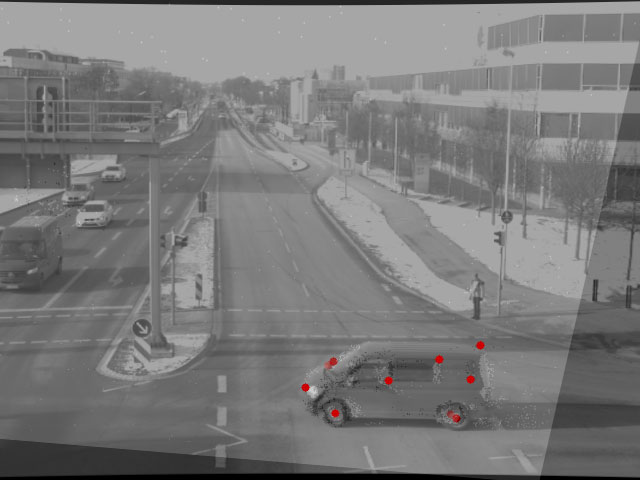} &
		\includegraphics[width=2.08cm]{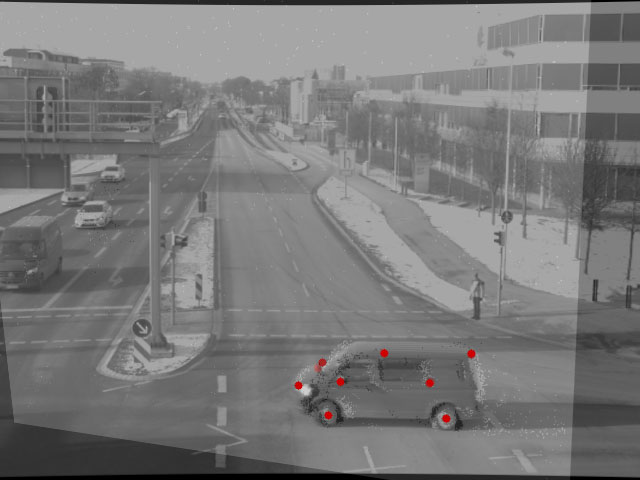} &
		\includegraphics[width=2.08cm]{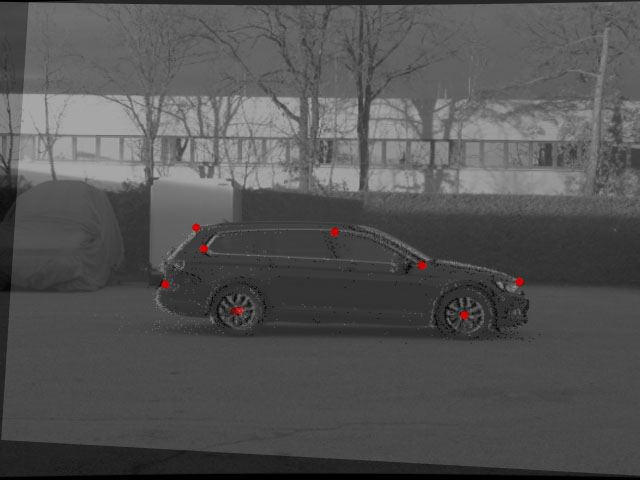} &
		\includegraphics[width=2.08cm]{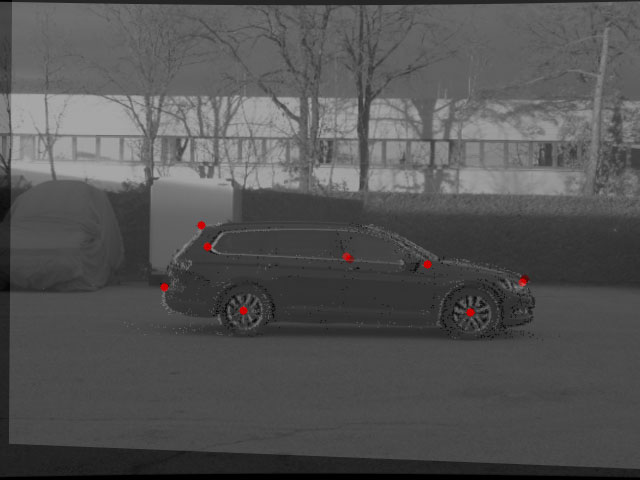} &
		\includegraphics[width=2.08cm]{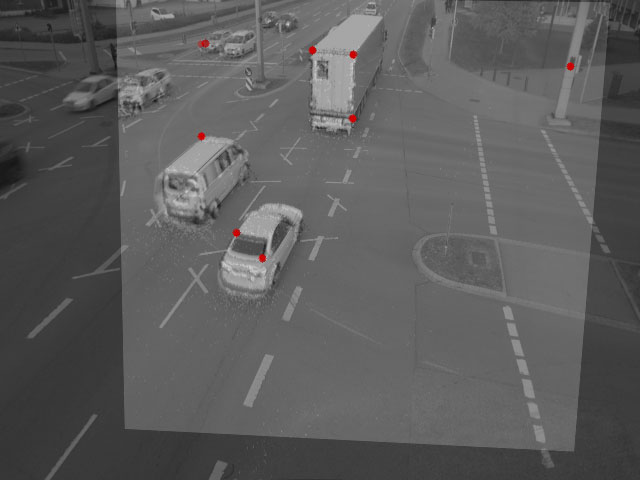} &
		\includegraphics[width=2.08cm]{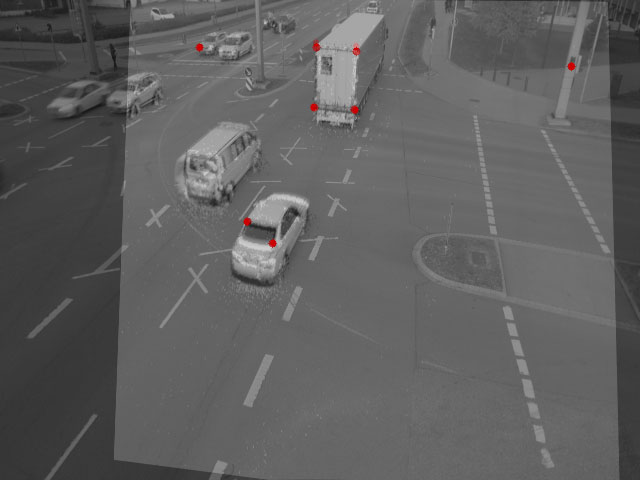} \\
		\rot{\hspace{15pt}ICP} & 
		\fcolorbox{black}{white}{\includegraphics[width=2.08cm]{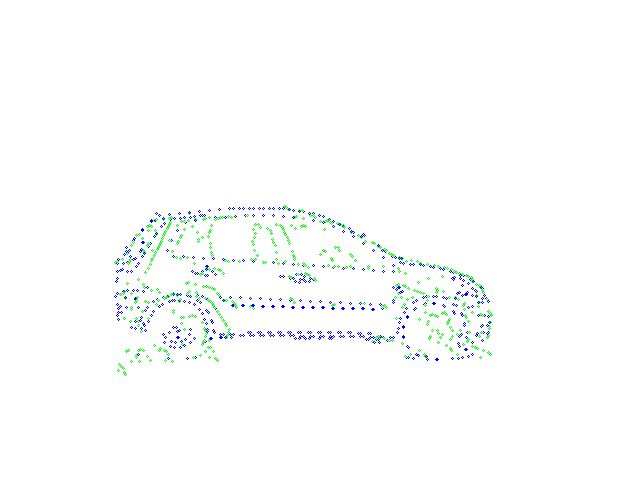}} &
		\fcolorbox{black}{white}{\includegraphics[width=2.08cm]{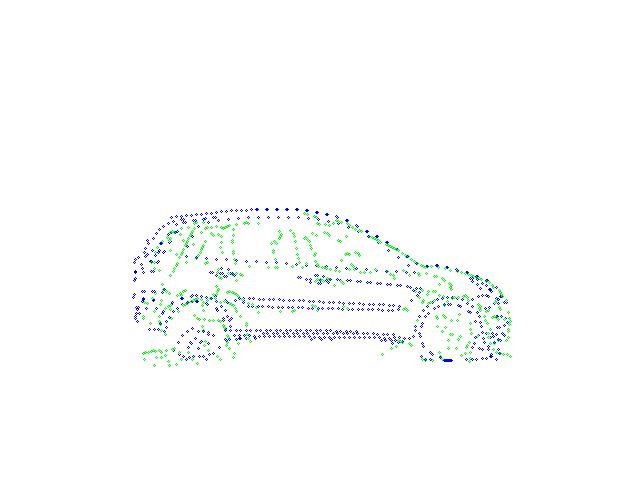}} &
		\fcolorbox{black}{white}{\includegraphics[width=2.08cm]{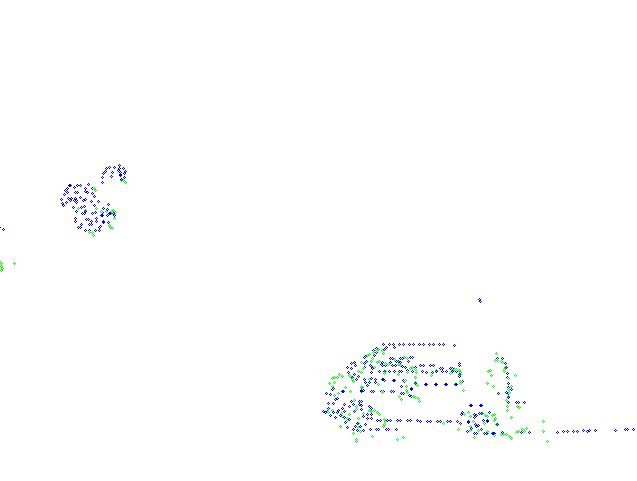}} &
		\fcolorbox{black}{white}{\includegraphics[width=2.08cm]{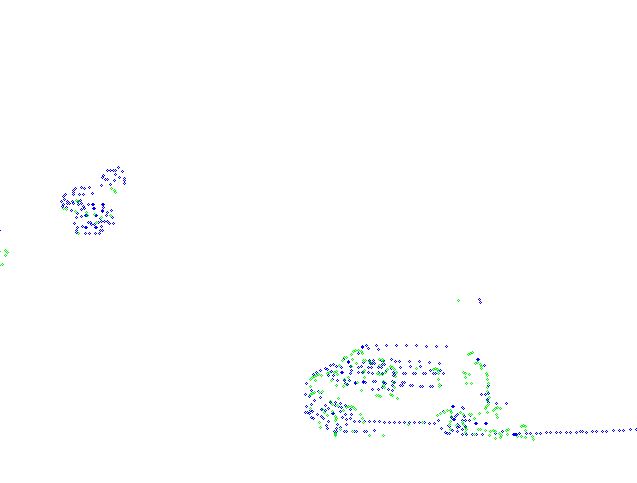}} &
		\fcolorbox{black}{white}{\includegraphics[width=2.08cm]{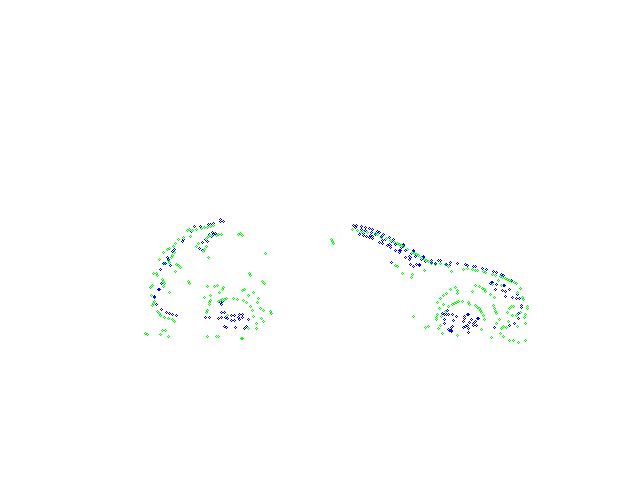}} &
		\fcolorbox{black}{white}{\includegraphics[width=2.08cm]{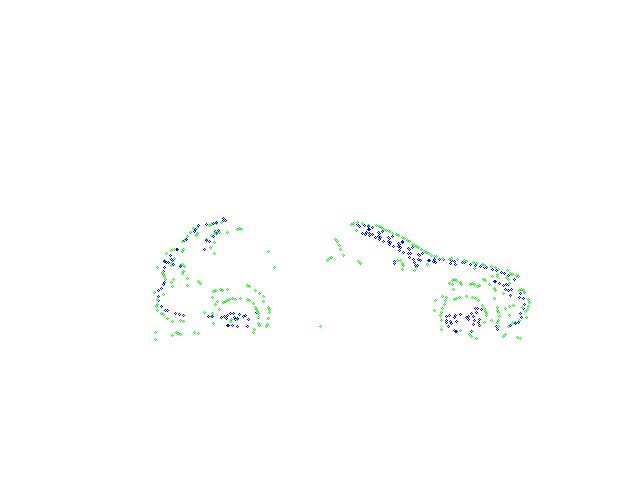}} &
		\fcolorbox{black}{white}{\includegraphics[width=2.08cm]{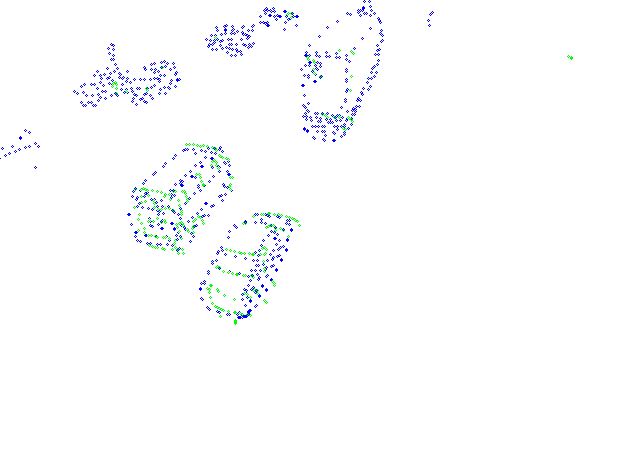}} &
		\fcolorbox{black}{white}{\includegraphics[width=2.08cm]{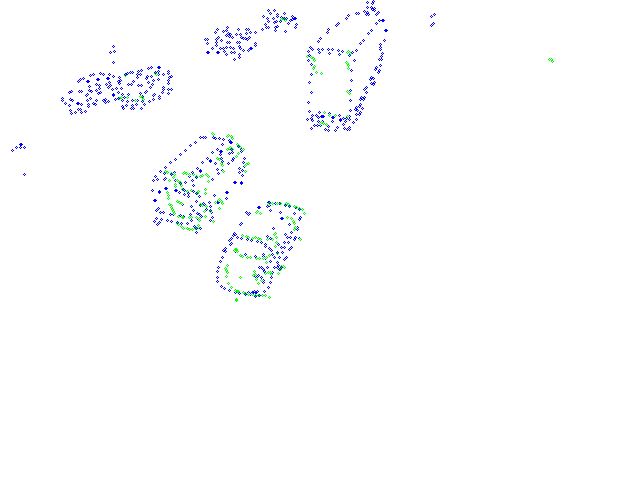}} \\
		\rot{\hspace{8pt}Results} & 
		\includegraphics[width=2.08cm]{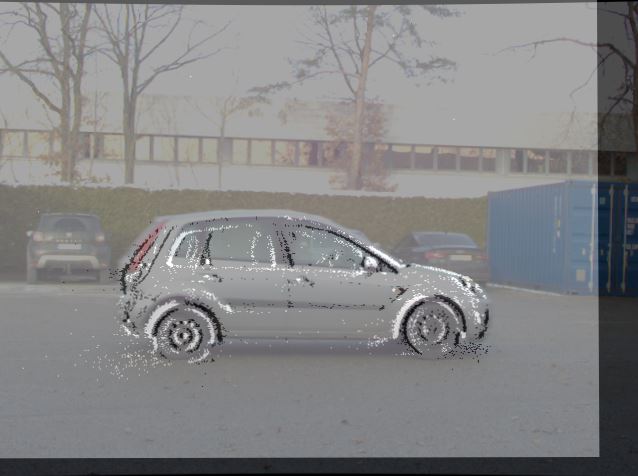} &
		\includegraphics[width=2.08cm]{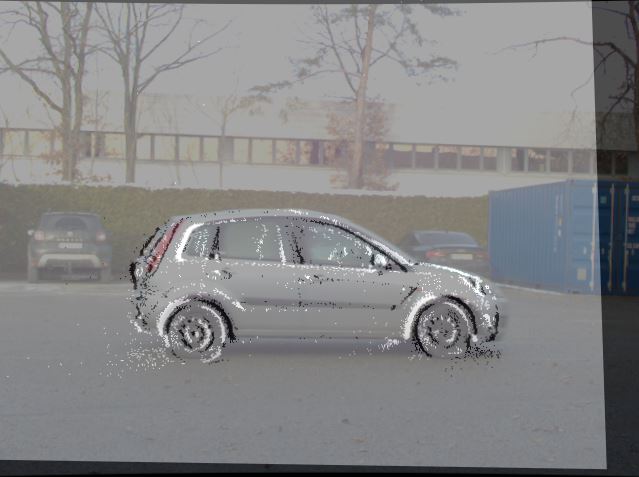} &
		\includegraphics[width=2.08cm]{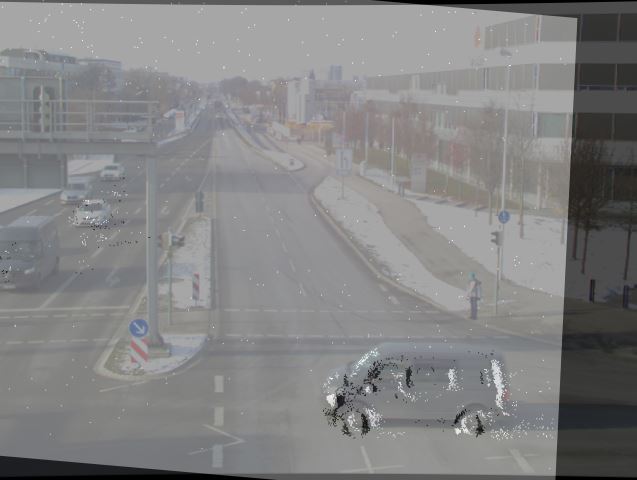} &
		\includegraphics[width=2.08cm]{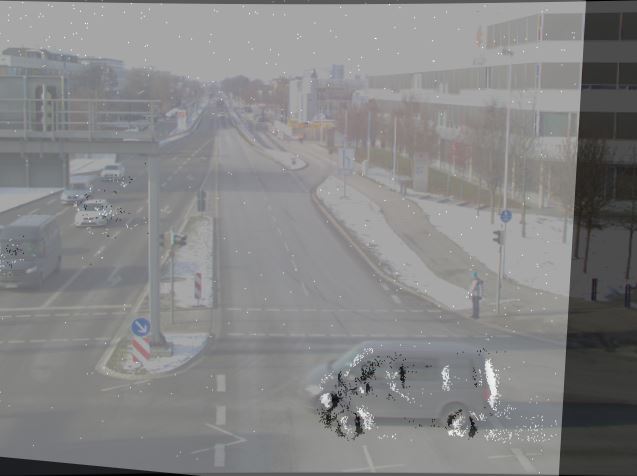} &
		\includegraphics[width=2.08cm]{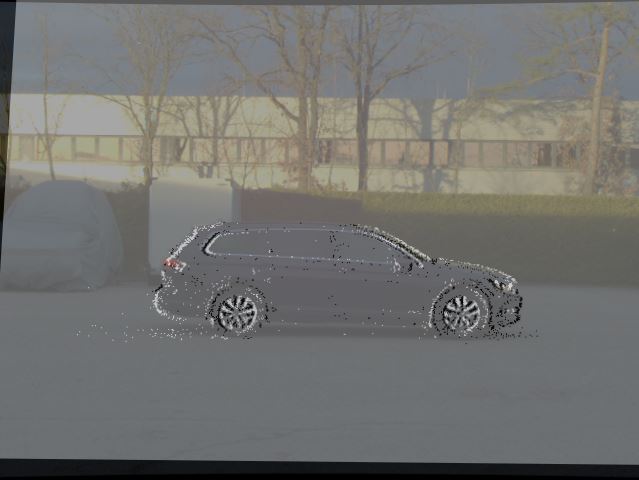} &
		\includegraphics[width=2.08cm]{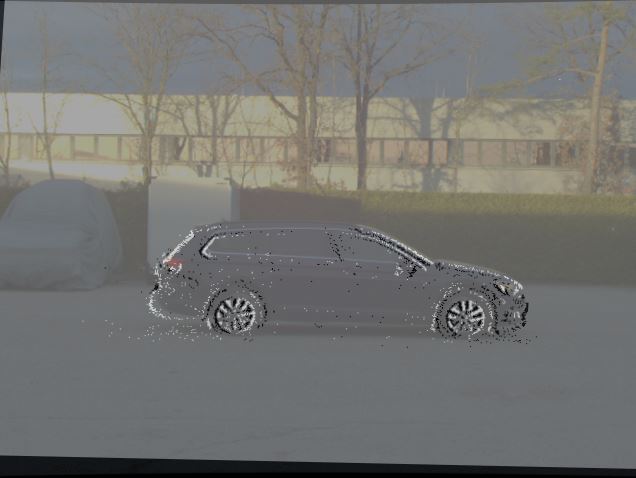} &
		\includegraphics[width=2.08cm]{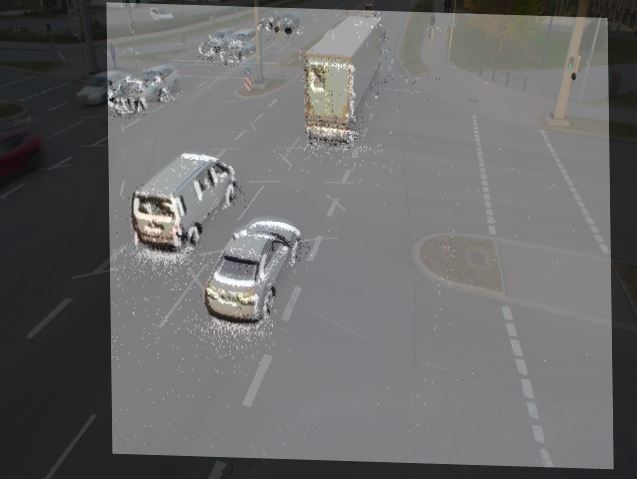} &
		\includegraphics[width=2.08cm]{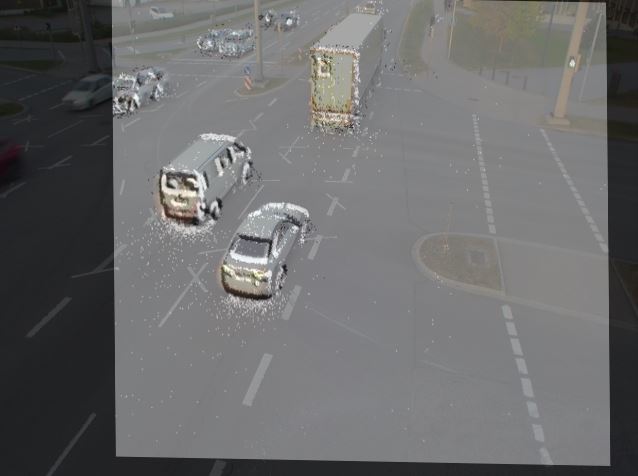} \\
	\end{tabular}
	\caption{The main goal of targetless extrinsic calibration is the association of features, e.g., edges, in both modalities. By manually selecting key points, we marked common points in the images from the event-based and RGB cameras and thus created a ground truth. As can be seen, our improved targetless calibration is not only convincing for single-moving objects, but the approach can also carry out valid camera calibration in complex traffic scenarios with multiple-moving objects.}
	\label{fig:qualitative_calibration}
\end{figure*}
\endgroup

\begingroup
\begin{figure*}[h!]
	\centering
	\setlength{\tabcolsep}{2pt} 
	\fboxsep=0.001pt
	\fboxrule=0.05pt
	\begin{tabular}{rcccccc}
		& \multicolumn{2}{c}{Example 1} & \multicolumn{2}{c}{Example 2} & \multicolumn{2}{c}{Example 3} \\
		& Setting $S_1$ & Setting $S_2$ & Setting $S_1$ & Setting $S_2$ & Setting $S_1$ & Setting $S_2$ \\
		\rot{\hspace{15pt}Results} & 
		\includegraphics[width=2.82cm]{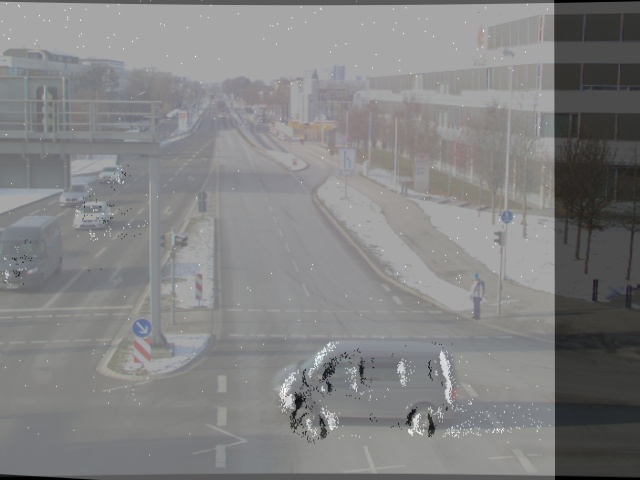} &
		\includegraphics[width=2.82cm]{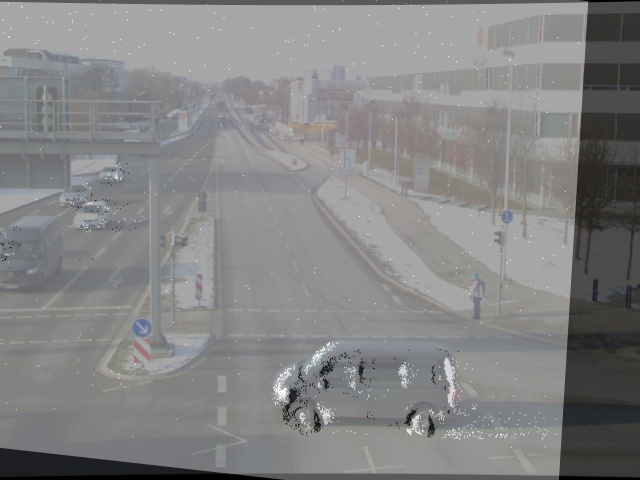} &
		\includegraphics[width=2.82cm]{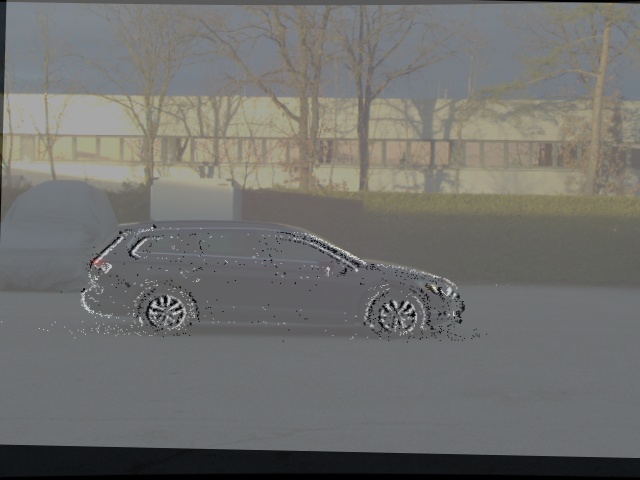} &
		\includegraphics[width=2.82cm]{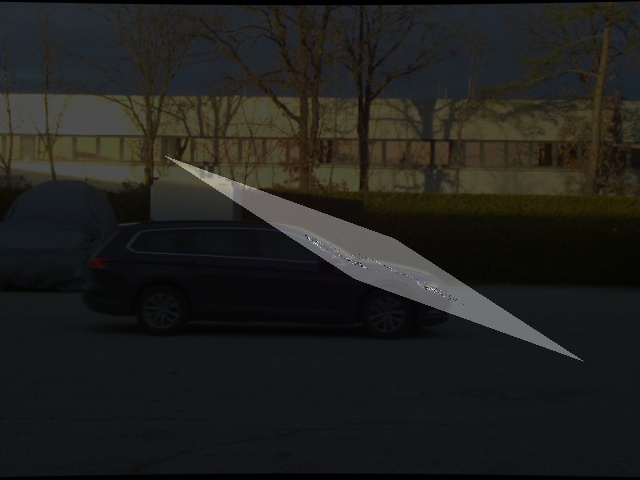} &
		\includegraphics[width=2.82cm]{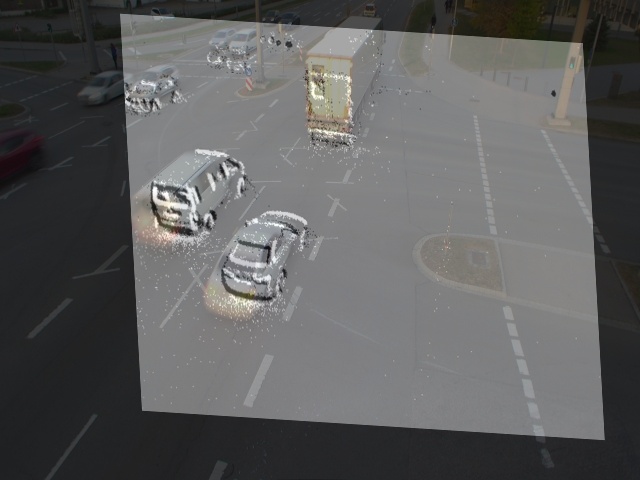} &
		\includegraphics[width=2.82cm]{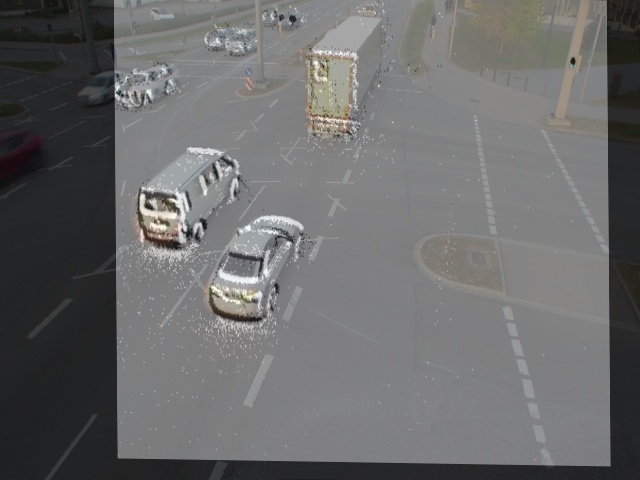} \\
		\rot{\hspace{3pt}Cluster RGB} & 
		\fcolorbox{black}{white}{\includegraphics[width=2.82cm]{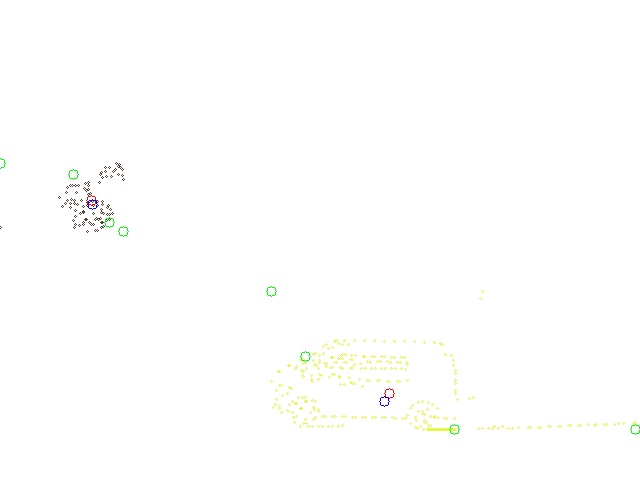}} &
		\fcolorbox{black}{white}{\includegraphics[width=2.82cm]{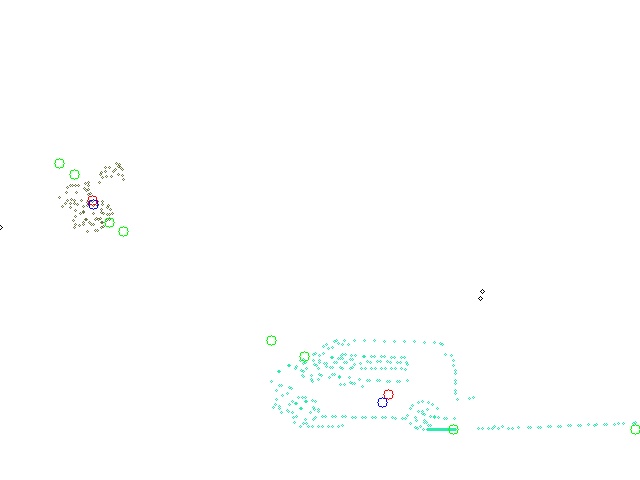}} &
		\fcolorbox{black}{white}{\includegraphics[width=2.82cm]{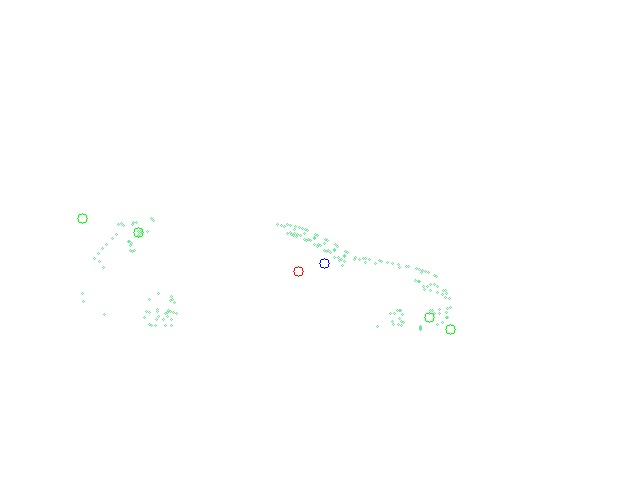}} &
		\fcolorbox{black}{white}{\includegraphics[width=2.82cm]{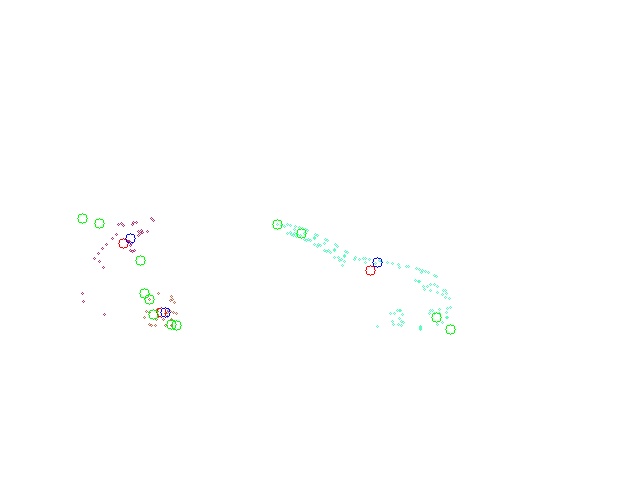}} &
		\fcolorbox{black}{white}{\includegraphics[width=2.82cm]{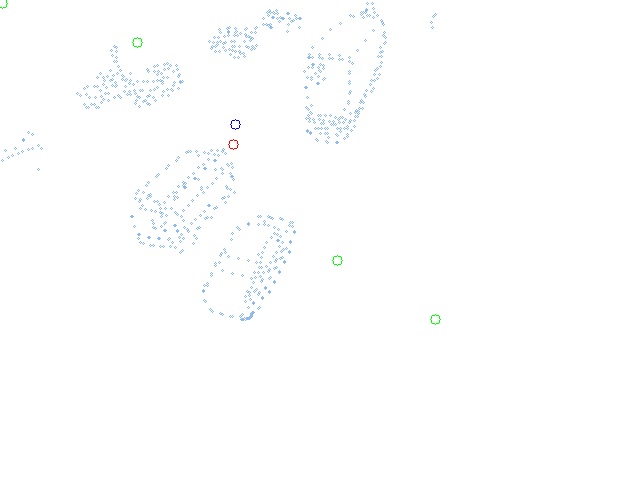}} &
		\fcolorbox{black}{white}{\includegraphics[width=2.82cm]{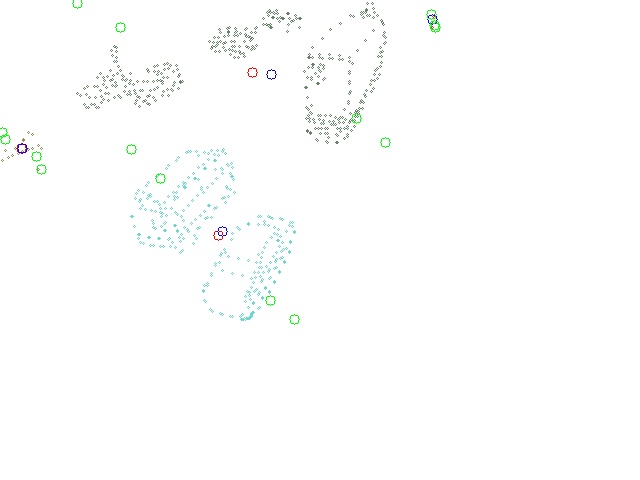}} \\
		\rot{\hspace{7pt}Cluster EB} & 
		\fcolorbox{black}{white}{\includegraphics[width=2.82cm]{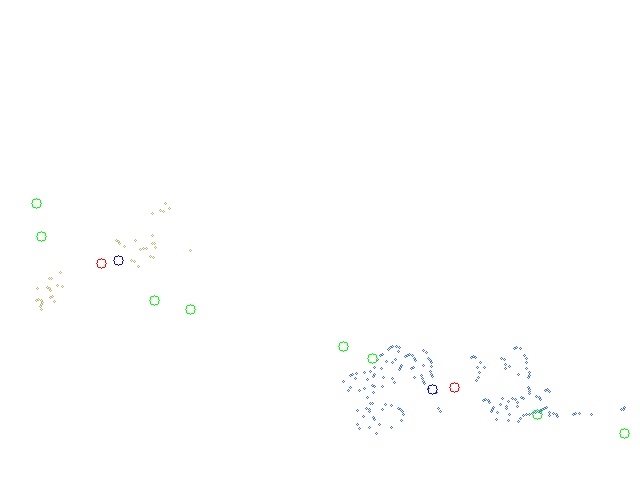}} &
		\fcolorbox{black}{white}{\includegraphics[width=2.82cm]{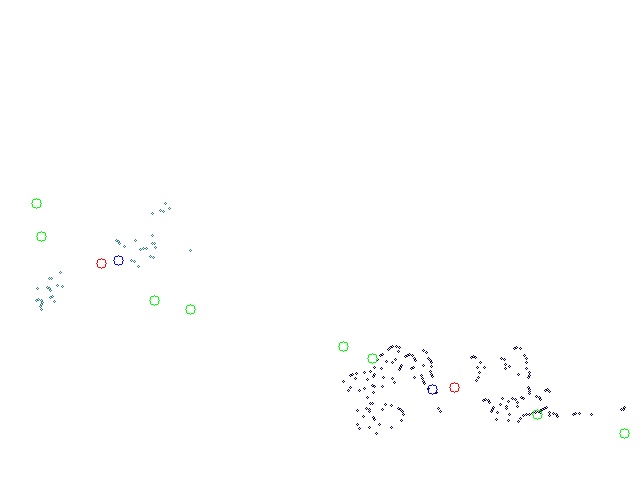}} &
		\fcolorbox{black}{white}{\includegraphics[width=2.82cm]{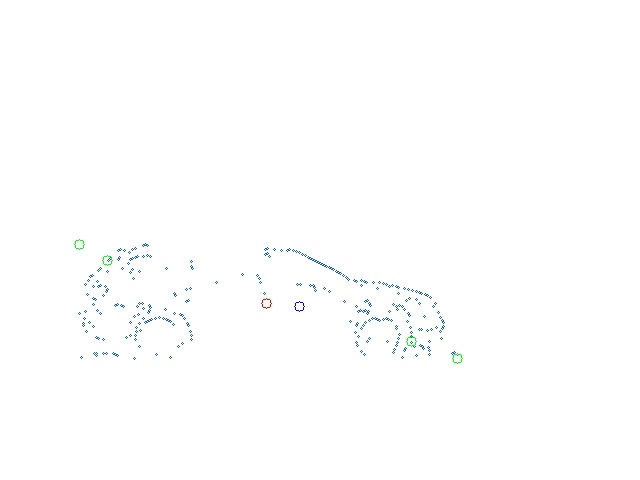}} &
		\fcolorbox{black}{white}{\includegraphics[width=2.82cm]{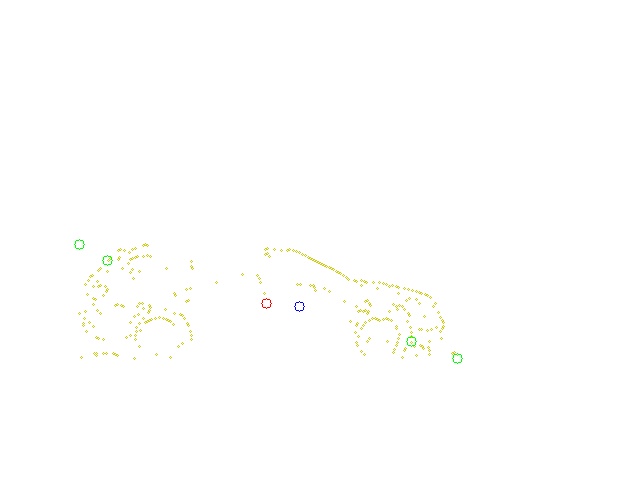}} &
		\fcolorbox{black}{white}{\includegraphics[width=2.82cm]{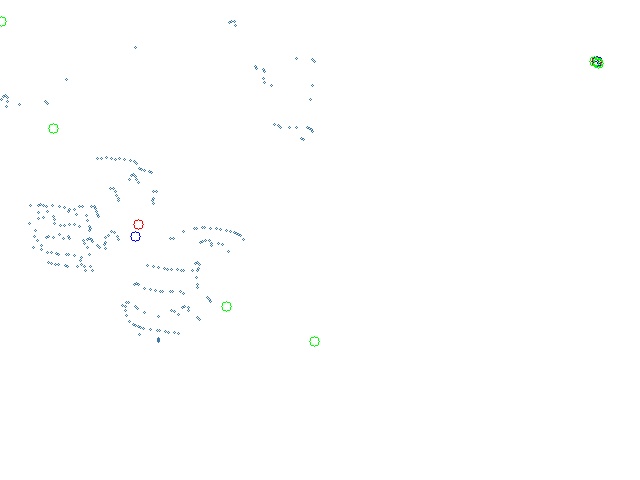}} &
		\fcolorbox{black}{white}{\includegraphics[width=2.82cm]{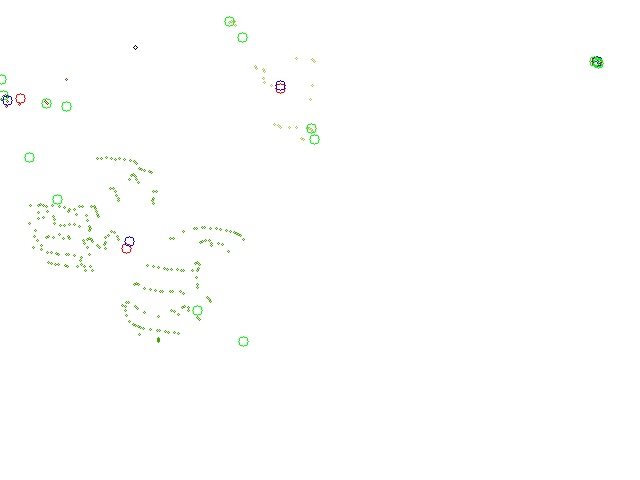}} \\
		\rot{\hspace{5pt}Cluster Ass.} & 
		\fcolorbox{black}{white}{\includegraphics[width=2.82cm]{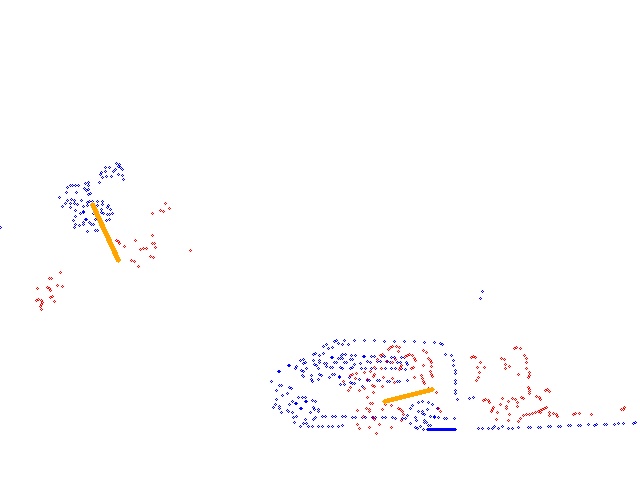}} &
		\fcolorbox{black}{white}{\includegraphics[width=2.82cm]{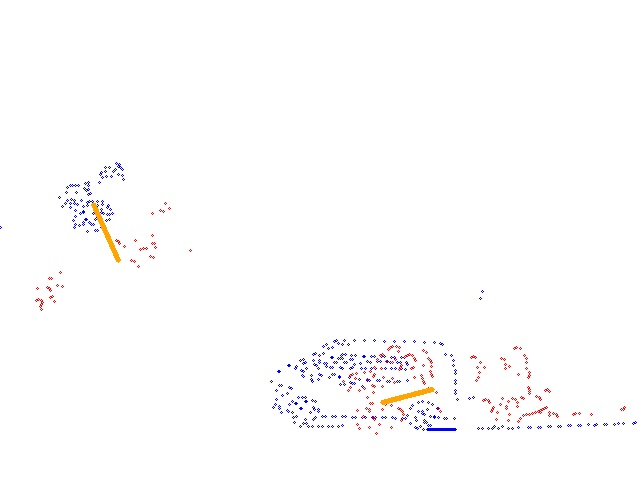}} &
		\fcolorbox{black}{white}{\includegraphics[width=2.82cm]{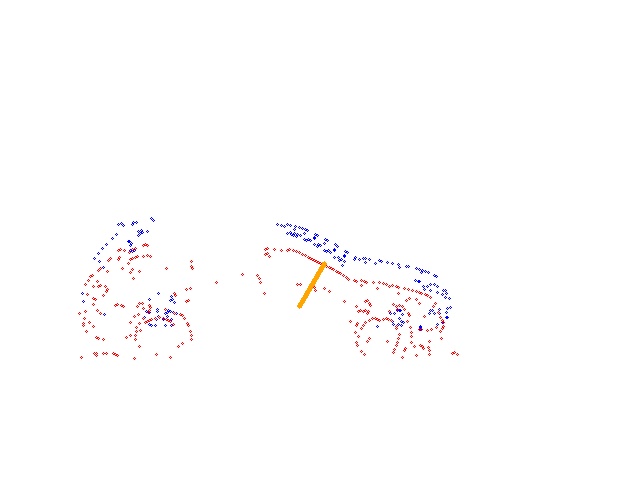}} &
		\fcolorbox{black}{white}{\includegraphics[width=2.82cm]{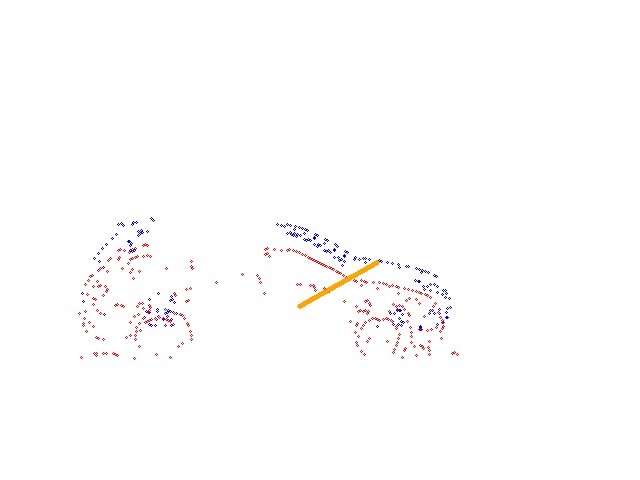}} &
		\fcolorbox{black}{white}{\includegraphics[width=2.82cm]{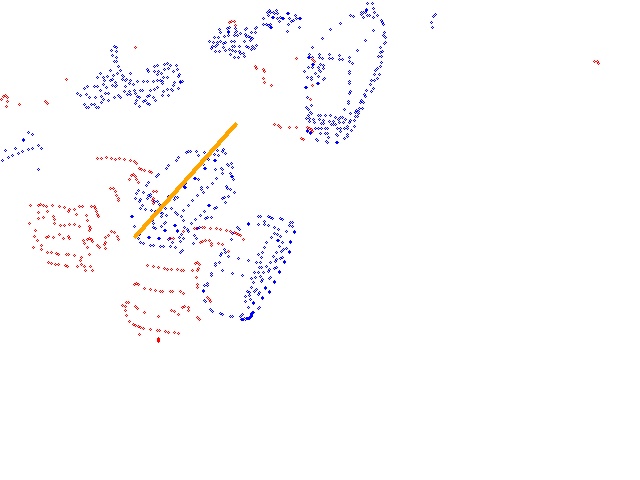}} &
		\fcolorbox{black}{white}{\includegraphics[width=2.82cm]{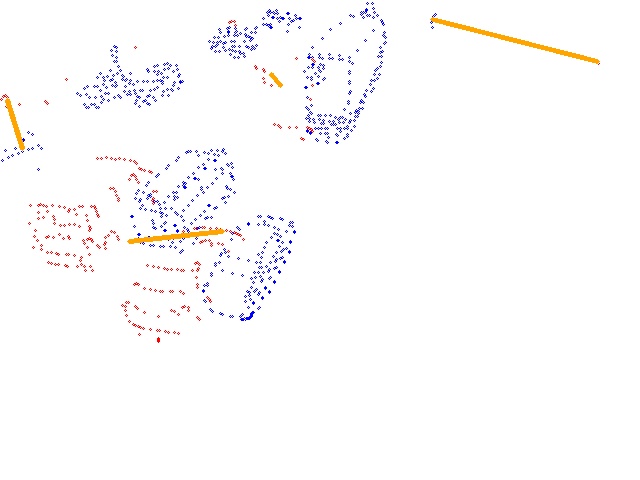}} \\
		\rot{\hspace{23pt}ICP} & 
		\fcolorbox{black}{white}{\includegraphics[width=2.82cm]{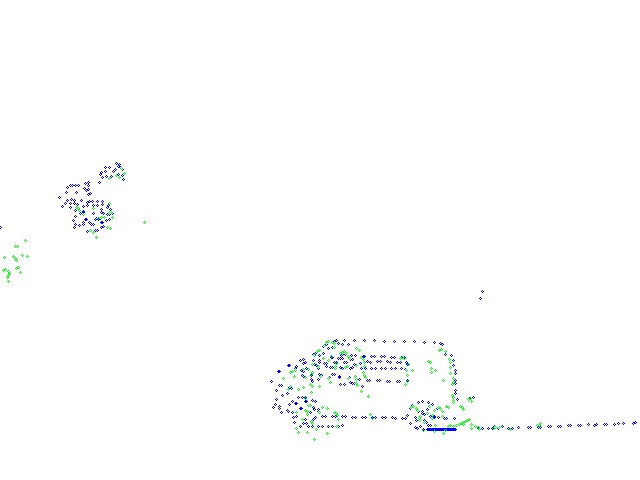}} &
		\fcolorbox{black}{white}{\includegraphics[width=2.82cm]{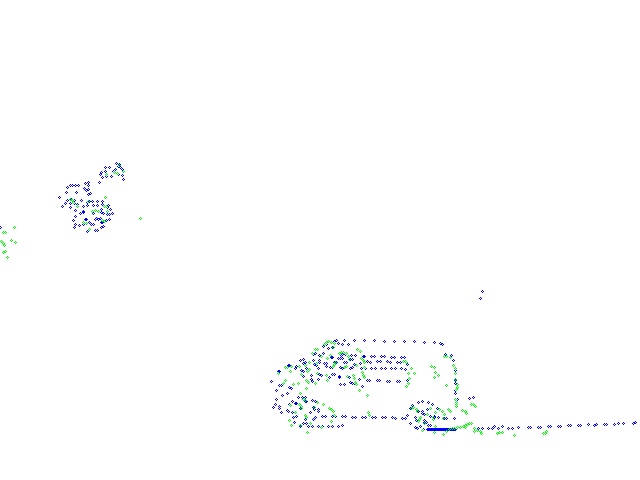}} &
		\fcolorbox{black}{white}{\includegraphics[width=2.82cm]{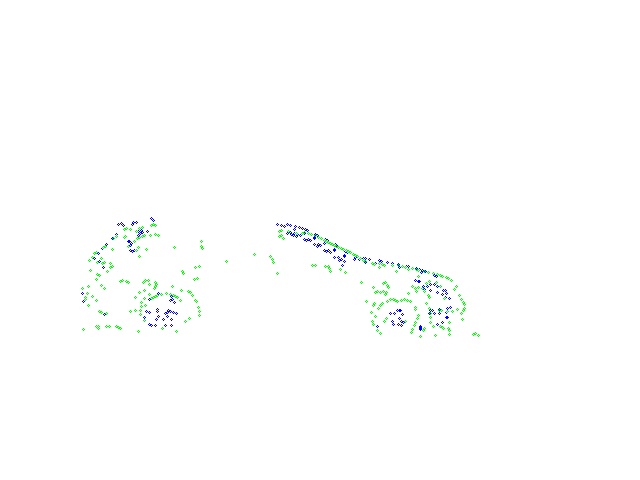}} &
		\fcolorbox{black}{white}{\includegraphics[width=2.82cm]{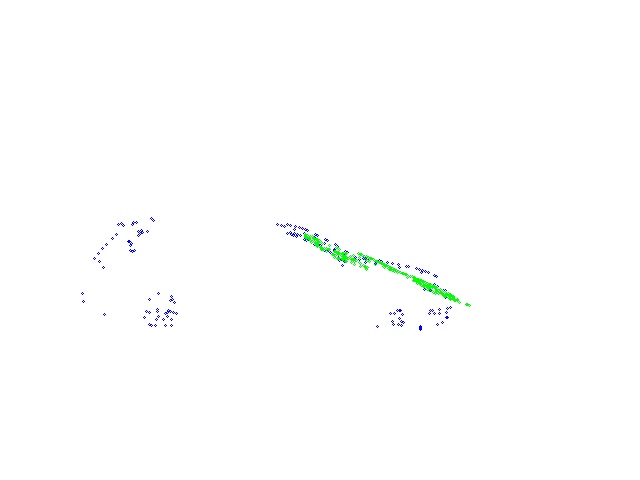}} &
		\fcolorbox{black}{white}{\includegraphics[width=2.82cm]{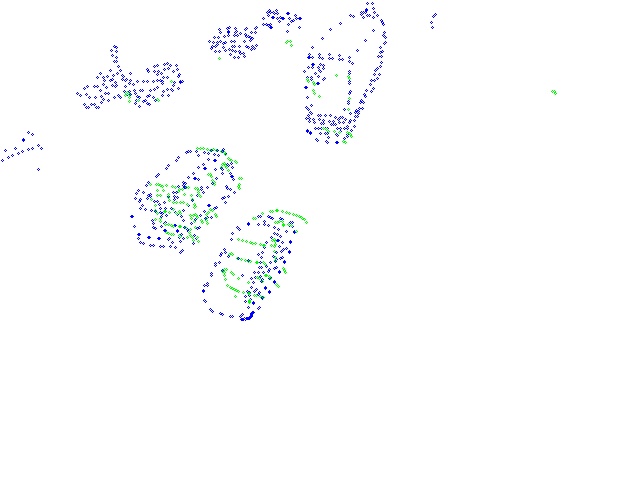}} &
		\fcolorbox{black}{white}{\includegraphics[width=2.82cm]{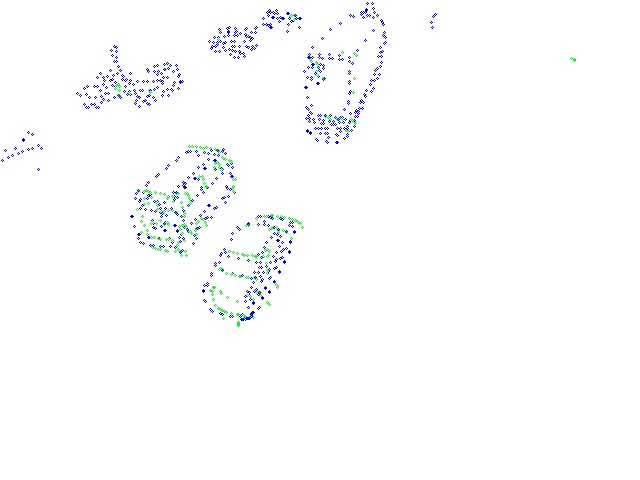}} \\		
	\end{tabular}
	\caption{The most influential in our targetless calibration are the DBSCAN \cite{EsterMartinandKriegelHansPeterandSanderJorgandXuXiaowei.1996} parameters of the maximum allowed distance $\epsilon$ between two samples within a cluster and the minimum number of samples $s_{min}$, which are necessary to define a cluster. Our use case is to generate almost equal clusters in both sensor modalities. To investigate these parameters, we apply the settings $S_1$ and $S_2$ to three examples. This figure shows the final results, the emerging clusters, the cluster assignments, and the ICP output. High values for $\epsilon$ allow a correct assignment for a single object, see $S_1$ in Example 2. However, smaller values allow a more fine-grained and accurate assignment. Furthermore, the value $s_{min}$ can successfully suppress noise during clustering.}
	\label{fig:dbscan_ablation_study}
\end{figure*}
\endgroup

\subsection{Runtime analysis}
Multi-modal sensor fusion ideally achieves more robust detection results by combining two sensor systems' strengths while eliminating their weaknesses. Since the fusion and detection is potentially calculated for every frame on an ITS, we would first like to examine the runtime performance. Table \ref{table:runtime_performance} shows the runtime of the detection and fusion processing pipeline. Early Fusion requires $15$ ms, while Simple Late Fusion respectively Spatiotemporal Late Fusion requires, in a sequential view, $43.6$ ms per frame. Interestingly, the most runtime requires pre-processing with $101$ ms, which mainly includes the event-based camera filtering method and the application of the intrinsic and extrinsic calibration matrices. In future work, GPU parallelization could significantly accelerate the pre-processing overhead and the sequential processing in the late fusion methods. The measurements were made on an Intel Core i9-9900KF CPU, NVIDIA GeForce RTX 2080 SUPER - 8GB VRAM, and 32 GB RAM.

\begin{table}[t]
	\caption{Since fusion and detection are calculated for every frame on an ITS, the runtime performance is highly relevant. Preprocessing, essentially the extrinsic and intrinsic calibration of the camera images, requires the most computing time.}
	\begin{center}
		\begin{tabular}{|r|r|}
			\hline
			\textbf{Component} & \textbf{ Runtime} \\
			\hline
			Preprocessing images & 101 ms \\
			Motion calculation & 10 ms \\
			CNN YoloV7 for RGB & 17 ms \\
			CNN YoloV7 for eb & 15 ms \\
			CNN YoloV7 for EF & 15 ms \\
			STLF / SLF & 1.6 ms \\
			\hline 
			\textbf{Total} & 159.6 ms \\			
			\hline
		\end{tabular}
		\label{table:runtime_performance}
	\end{center}
\end{table}

\subsection{Detection and fusion}
The detector for RGB images serves as input for the late fusion methods, generates pseudo-labels for the training of the event-based camera detector, and represents a baseline for the fusion methods. As described in Subsection \ref{subsection:methodology-dataset}, we first trained the YoloV7 CNN \cite{arXiv.org.07072022b, Wong.14012023, Chen.02012020} on the nuImages \cite{motional.08082023} dataset. Since the TUMTraf Dataset family considers roadside perspective, we finetuned the CNN to our TUMTraf Highway \cite{Cress.2022} and TUMTraf Intersection Dataset \cite{zimmer2023tumtraf} using transfer learning. So, we significantly increased the detection performance from $0.36$ mAP to $0.85$ mAP on our combined TUMTraf test set. 

Based on the RGB detector mentioned above and the extrinsic calibration, we generated the TUMTraf Event Dataset with pseudo-labels. We trained several object detectors for early and late fusion. The training with ``L-EB'' labels on event-based frames creates a robust detector for the event-based camera (EB), which we used for our late fusion approaches SLF and STLF. However, we trained the early fusion methods EF-1 and EF-2 on the combined RGB-event-based images. Here, we used the labels ``L-EB'' for EF-1 and ``L-RGB'' for EF-2. We want to emphasize that our training is based on the optimized variant of our training and validation set. Furthermore, we perform the evaluation on the carefully manually corrected test set. To measure the results, we used the toolkits of \cite{Wong.14012023} and \cite{Padilla.2021}.

\begin{table}[t!]
	\caption{This table shows the average precision, precision, and recall (AP, P, R) of the event-based detector based on our TUMTraf Event Dataset (EB). We conducted experiments during the day, at night with street lights on (N-1), and at night with street lights off (N-2). We achieved $0.26$ mAP during the day and up to $0.54$ mAP at night. Here, we used the ``L-EB'' labels, representing moving objects and, therefore, being visible to the event-based camera. Furthermore, we compared our detector to a trained detector based on the DSEC-Detection Dataset \cite{Gehrig.2021, Gehrig.22112022, DSEC-Detection.27022024} (EB-DSEC). Due to the ego perspective of EB-DSEC, object detection on our test set from the bird's eye perspective is impossible. This result underlines the relevance of our TUMTraf Event Dataset.}
	\begin{center}
		\begin{tabular}{|r|r|r|r|r|}
			\hline
			\multicolumn{2}{|c|}{\textbf{Testset}} & \textbf{Class} & \textbf{EB-DSEC} & \textbf{EB} \\
			\hline
			\multirow{8}{*}{\rot{\textbf{Day}}} & \multirow{8}{*}{\rot{\textbf{Full}}} & Pedestrian & 0.00 0.00 0.00 & \textbf{0.01} 0.10 0.08  \\
			& & Bicycle & 0.00 0.00 0.00 & 0.00 0.00 0.00 \\
			& & Car & \underline{0.04} 0.80 0.04 & \textbf{0.44} 0.61 0.61  \\
			& & Motorcycle & 0.00 0.00 0.00 & 0.00 0.00 0.00 \\
			& & Bus  & 0.00 0.00 0.00 & \textbf{0.92} 0.73 0.95  \\
			& & Truck & 0.00 0.00 0.00 & \textbf{0.10} 0.28 0.13  \\
			& & Trailer & 0.00 0.00 0.00 & \textbf{0.35} 0.83 0.37  \\
			\cline{3-5}
			& & \textbf{ $\varnothing$ Total} & \underline{0.01} 0.11 0.01 & \textbf{0.26} 0.36 0.31  \\
			\hline
			\multirow{3}{*}{\rot{\textbf{N-1}}} & \multirow{3}{*}{\rot{\textbf{Full}}} & Car & \underline{0.01} 0.22 0.01 & \textbf{0.64} 0.83 0.74  \\ 
			& & Bus & 0.00 0.00 0.00 & \textbf{0.44} 0.91 0.46  \\
			\cline{3-5}
			& & \textbf{$\varnothing$ Total} & \underline{0.01} 0.11 0.01 & \textbf{0.54} 0.87 0.60  \\
			\hline
			\multirow{2}{*}{\rot{\textbf{N-2}}} & \multirow{2}{*}{\rot{\textbf{Full}}} & Car & 0.00 0.00 0.00 & \textbf{0.33} 0.88 0.33  \\ 
			\cline{3-5}
			& & \textbf{$\varnothing$ Total} & 0.00 0.00 0.00 & \textbf{0.33} 0.88 0.33  \\
			\hline
		\end{tabular}
		\label{table:eb_performance}
	\end{center}
\end{table}

First, we analyze the performance of our event-based object detector (EB) with ``L-EB'' as ground truth. The quantitative results are in Table \ref{table:eb_performance}, and the qualitative results are in Figure \ref{fig:qualitative_fusion}. In scenario ``Day,'' we achieved satisfactory results with the large object classes, e.g., car ($0.44$ AP) or bus ($0.92$ AP). These classes contain sufficient texture information and allow adequate detection. However, we recognized worse performance with optically small objects that don't include enough features, e.g., pedestrians or bicycles. Another reason may be the low occurrence of these classes in the training set. Scenario ``N-1'' is a traffic scene at night with street lights on with buses and cars. The event-based detector still delivered satisfactory results: We could detect cars with $0.64$ AP. Even in scenario ``N-2,'' a night with street lights off, we could detect cars with $0.33$ AP and a precision of $0.88$. At this point, the advantages of the event-based camera stand out clearly: $\text{1.) The}$ high dynamic range of the event-based camera allows the detection of objects under extreme illumination conditions. $\text{2.) Since}$ the stationary event-based camera provides a clear image mask, false positives in the background area are significantly reduced. Nevertheless, the recall of the class car compared to the day scenario dropped significantly from $0.61$ to $0.33$ in absolute darkness. A possible cause could be disturbing artifacts, e.g., light beams from headlights, see Figure \ref{fig:qualitative_fusion}. These artifacts were mostly, but not completely, filtered with the spatiotemporal filtering by \cite{Yang.2022}. A data fusion between an event-based and RGB camera could combine the advantages of both modalities and thus reduce the disadvantage of a lower recall caused by less texture information. 

\begin{table*}[h!]
	\caption{We evaluated our detectors and fusion methods on the TUMTraf Event test set with the ``L-RGB'' labels. We distinguish between day, night with street lights on (N-1), and night with street lights off (N-2). Furthermore, if available, we analyzed the dominant traffic flows, ``Standing,'' ``Vertical,'' and ``Horizontal.'' We used average precision, precision, and recall (AP, P, R) as metrics for the RGB detector, the IFCNN fusion \cite{Zhang.2020}, and the early fusion detectors EF-DSEC, which is trained on the DSEC-Detection Dataset \cite{Gehrig.2021, Gehrig.22112022, DSEC-Detection.27022024}, EF-1, and EF-2. To evaluate our Simple Late Fusion (SLF) and Spatiotemporal Late Fusion (STLF), we used the average precision (AP). Thresholds: Confidence $= 0.3$; IoU $= 0.45$; STLF Confidence $= 0.77$. The performance of the RGB detector, with a confidence threshold of $0.80$, was also examined (RGB-0.80). The poor performance of EF-DSEC and EF-1 and the high precision of the event-based detector are noteworthy. The drop in performance with STLF is due to the more strict fusion logic. However, SLF significantly outperforms the RGB detector in the subsets day and N-2. The best AP values are highlighted.}
	\begin{center}
		\begin{tabular}{|r|r|r||r|r|r|r|r|r|R{0.7cm}|R{0.7cm}|}
			\hline
			\multicolumn{2}{|c|}{\textbf{Testset}} & \textbf{Class} & \textbf{RGB} & \textbf{RGB-0.80} & \textbf{IFCNN} & \textbf{EF-DSEC} & \textbf{EF-1} & \textbf{EF-2} & \textbf{SLF} & \textbf{STLF} \\
			\hline
			\multirow{26}{*}{\rot{\textbf{Day}}} & \multirow{8}{*}{\rot{\textbf{Full}}} & Pedestrian & \underline{0.64} 0.91 0.66  & 0.17 0.97 0.17  & 0.22 0.94 0.22 & 0.03 0.53 0.03 & 0.00 0.00 0.00 & 0.24 0.94 0.25 & \textbf{0.71}  & 0.26 \\
			& & Bicycle & \underline{0.73} 0.71 0.77  & 0.38 0.98 0.38  & 0.04 1.00 0.04 & 0.00 0.00 0.00 & 0.00 0.00 0.00 & 0.25 0.90 0.25 & \textbf{0.74} & 0.55 \\
			& & Car & \underline{0.94} 0.95 0.95  & 0.83 0.99 0.83  & 0.84 0.96 0.84 & 0.21 0.85 0.21 &  0.10 0.37 0.15 & 0.89 0.96 0.90 & \textbf{0.95}  & 0.86 \\
			& & Motorcycle & 0.37 0.56 0.48  & 0.20 0.44 0.19  & 0.05 1.00 0.05 & 0.00 0.00 0.00 & 0.00 0.00 0.00 & 0.10 0.67 0.10 & \textbf{0.75}  & \underline{0.57} \\
			& & Bus  & \underline{0.98} 0.74 1.00  & 0.92 0.91 0.91  & \textbf{0.99} 0.65 1.00 & 0.67 0.78 0.67 & 0.61 0.82 0.61 & 0.94 0.69 0.98 & \underline{0.98}  & 0.97  \\
			& & Truck & \underline{0.58} 0.48 0.76  & 0.36 0.76 0.39  & 0.42 0.35 0.52 &  0.01 0.25 0.02 & 0.05 0.25 0.08 & 0.50 0.47 0.61 & \textbf{0.63} & 0.47  \\
			& & Trailer & \underline{0.60} 0.69 0.74  & 0.24 0.80 0.28  & 0.58 0.82 0.69 &  0.00 0.00 0.00 & 0.26 0.65 0.33 & 0.55 0.80 0.67 & \textbf{0.71} & 0.46  \\
			\cline{3-11}
			& & \textbf{ $\varnothing$ Total} & \underline{0.69} 0.72 0.77  & 0.44 0.84 0.45  & 0.45 0.82 0.48 & 0.13 0.34 0.13 & 0.15 0.30 0.17 & 0.50 0.78 0.54 & \textbf{0.78} & 0.59 \\
			\cline{2-11}
			& \multirow{8}{*}{\rot{Standing}} & Pedestrian & \underline{0.70} 0.95 0.68 & 0.18 0.97 0.18  & 0.23 0.96 0.23 & 0.04 0.56 0.05 & 0.00 0.01 0.00 & 0.27 0.98 0.27 & \textbf{0.77}  & 0.29 \\
			& & Bicycle & \textbf{0.87} 0.72 0.87 & 0.52 0.98 0.53  & 0.06 1.00 0.05 & 0.00 0.00 0.00 & 0.00 0.00 0.00 & 0.34 0.93 0.35 & \textbf{0.87} & \underline{0.71} \\
			& & Car & \underline{0.93} 0.96 0.93 & 0.83 0.98 0.83  & 0.80 0.96 0.81 & 0.26 0.82 0.27 & 0.07 0.35 0.11 & 0.89 0.96 0.89 & \textbf{0.94} & 0.86 \\
			& & Bus  & \underline{0.99} 0.96 1.00 & 0.89 1.00 0.89  & \underline{0.99} 0.93 1.00 &  0.65 0.70 0.65 & 0.62 0.80 0.62 & \underline{0.99} 0.90 1.00 & \textbf{1.00} & \textbf{1.00}  \\
			& & Truck & \underline{0.72} 0.68 0.79 & 0.40 0.92 0.40  & 0.58 0.42 0.64 &  0.01 0.24 0.03 & 0.00 0.10 0.03 & 0.68 0.63 0.71 & \textbf{0.78} & 0.52  \\
			& & Trailer & 0.56 0.70 0.70 & 0.13 0.61 0.20  & \underline{0.61} 0.76 0.76 & 0.00 0.00 0.00 & 0.20 0.52 0.25 & 0.58 0.75 0.74 & \textbf{0.73} & 0.32  \\
			\cline{3-11}
			& & \textbf{ $\varnothing$ Total} & \underline{0.80} 0.83 0.83 & 0.49 0.91 0.50 & 0.55 0.84 0.58 & 0.16 0.39 0.17 & 0.15 0.30 0.17 & 0.63 0.86 0.66 & \textbf{0.85} & 0.62 \\
			\cline{2-11}
			& \multirow{8}{*}{\rot{Vertical}} & Pedestrian & \underline{0.44} 0.78 0.47 & 0.13 0.92 0.12  & 0.21 0.86 0.23 & 0.00 0.00 0.00 & 0.00 0.00 0.00 & 0.14 0.72 0.19 & \textbf{0.53} & 0.19 \\
			& & Bicycle & \textbf{0.48} 1.00 0.47 & 0.06 1.00 0.06 & 0.00 0.00 0.00 & 0.00 0.00 0.00 & 0.00 0.00 0.00 & 0.03 1.00 0.03 & \underline{0.46} & 0.21 \\
			& & Car & \textbf{0.96} 0.95 0.96 & 0.85 0.99 0.86  & 0.87 0.95 0.88 & 0.18 0.85 0.19 &  0.13 0.35 0.18 & \underline{0.90} 0.95 0.91 & \textbf{0.96} & \underline{0.90} \\
			& & Motorcycle & 0.37 0.62 0.47 & 0.20 0.44 0.19 & 0.05 1.00 0.05 & 0.00 0.00 0.00 &  0.00 0.00 0.00 & 0.10 1.00 0.10 & \textbf{0.75} & \underline{0.57} \\
			& & Truck & \underline{0.42} 0.37 0.68 & 0.30 0.57 0.37 & 0.12 0.23 0.29 & 0.00 0.00 0.00 & 0.13 0.44 0.18 & 0.19 0.28 0.42 & \textbf{0.46} & 0.39  \\
			& & Trailer & \underline{0.67} 0.75 0.70 & 0.38 1.00 0.38  & 0.60 0.93 0.61 & 0.00 0.00 0.00 & 0.36 0.80 0.42 & 0.56 0.92 0.58 & \textbf{0.72} & 0.63  \\
			\cline{3-11}
			& & \textbf{ $\varnothing$ Total} & \underline{0.56} 0.75 0.62 & 0.32 0.82 0.33  & 0.31 0.66 0.34 & 0.03 0.14 0.03 & 0.10 0.26 0.13 & 0.32 0.81 0.37 & \textbf{0.65} & 0.48 \\
			\cline{2-11}
			& \multirow{4}{*}{\rot{Horiz.}} & Pedestrian & \textbf{0.13} 1.00 0.13 & 0.00 0.00 0.00  & 0.00 0.00 0.00 & 0.00 0.00 0.00 & 0.00 0.00 0.00 & 0.03 1.00 0.03 & \textbf{0.13}  & 0.00 \\
			& & Car & \textbf{0.95} 0.98 0.95 & 0.78 1.00 0.78 & 0.85 1.00 0.85 & 0.19 0.93 0.19 & 0.13 0.58 0.16 & \underline{0.86} 0.99 0.86 & \textbf{0.95} & 0.79 \\
			& & Bus  & \textbf{1.00} 1.00 1.00 & 0.96 1.00 0.95 & \textbf{1.00} 1.00 1.00 & 0.71 1.00 0.70 & 0.60 0.86 0.60 & 0.96 1.00 0.95 & \underline{0.99} & 0.95  \\
			\cline{3-11}
			& & \textbf{ $\varnothing$ Total} & \textbf{0.69} 0.99 0.69 & 0.58 0.67 0.58 & \underline{0.62} 0.67 0.62 & 0.30 0.64 0.30 & 0.24 0.48 0.25 & \underline{0.62} 1.00 0.61 & \textbf{0.69} & 0.58 \\
			\hline
			\multirow{9}{*}{\rot{\textbf{N-1}}} & \multirow{3}{*}{\rot{\textbf{Full}}} & Car & \textbf{0.71} 0.91 0.56  & 0.22 0.98 0.22  & 0.47 0.79 0.53 & 0.02 0.75 0.02 & 0.03 0.31 0.08 & 0.38 0.74 0.46 & \underline{0.70} & 0.41 \\ 
			& & Bus & \textbf{0.80} 0.62 0.83  & 0.57 0.96 0.57 & 0.39 0.45 0.55 &  0.00 0.00 0.00 & 0.10 1.00 0.10 & 0.27 0.58 0.33 & \underline{0.62} & 0.43  \\
			\cline{3-11}
			& & \textbf{$\varnothing$ Total} & \textbf{0.76} 0.77 0.70  & 0.40 0.97 0.40  & 0.43 0.62 0.54 & 0.01 0.38 0.01 & 0.07 0.66 0.09 & 0.33 0.66 0.40 & \underline{0.66} & 0.42  \\
			\cline{2-11}
			& \multirow{3}{*}{\rot{Ver.}} & Car & \textbf{0.72} 0.88 0.72  & 0.20 0.99 0.20  & 0.48 0.83 0.52 & 0.02 0.76 0.02 & 0.02 0.27 0.07 & 0.43 0.80 0.49 & \underline{0.71} & 0.38 \\
			& & Bus & 0.05 0.07 0.50 & 0.00 0.00 0.00 & 0.04 0.06 0.18 & 0.00 0.00 0.00 & 0.00 0.00 0.00 & \underline{0.17} 0.10 0.17 & 0.13 & \textbf{0.28}  \\
			\cline{3-11}
			& & \textbf{$\varnothing$ Total} & \underline{0.39} 0.48 0.61  & 0.10 0.50 0.10  & 0.26 0.45 0.35 & 0.01 0.38 0.01 & 0.01 0.14 0.04 & 0.30 0.45 0.33 & \textbf{0.42} & 0.33  \\
			\cline{2-11}	
			& \multirow{3}{*}{\rot{Hor.}} & Car & \textbf{0.70} 0.93 0.54  & 0.33 0.94 0.34  & 0.51 0.66 0.60 &  0.03 1.00 0.02 & 0.09 0.59 0.14 & 0.19 0.43 0.31 & \underline{0.66} & 0.58 \\
			& & Bus & \textbf{0.91} 0.88 0.86 & 0.67 0.96 0.67 & 0.48 0.64 0.60 & 0.00 0.00 0.00 & 0.12 1.00 0.11 & 0.35 0.93 0.36 & \underline{0.71} & 0.45  \\
			\cline{3-11}
			& & \textbf{$\varnothing$ Total} & \textbf{0.81} 0.91 0.70  & 0.50 0.95 0.50  & 0.49 0.65 0.60 & 0.01 0.50 0.01 & 0.11 0.80 0.12 & 0.27 0.68 0.33 & \underline{0.69} & 0.52  \\
			\hline
			\multirow{6}{*}{\rot{\textbf{N-2}}} & \multirow{2}{*}{\rot{\textbf{Full}}} & Car & \underline{0.43} 0.85 0.38  & 0.13 1.00 0.12  & 0.04 0.46 0.07 &  0.00 0.00 0.00 & 0.01 0.25 0.04 & 0.07 0.38 0.13 & \textbf{0.49} & 0.27 \\ 
			\cline{3-11}
			& & \textbf{$\varnothing$ Total} & \underline{0.43} 0.85 0.38  & 0.13 1.00 0.12  & 0.04 0.46 0.07 & 0.00 0.00 0.00 & 0.01 0.25 0.04 & 0.07 0.38 0.13 & \textbf{0.49} & 0.27 \\
			\cline{2-11}
			& \multirow{2}{*}{\rot{Ver.}} & Car &  \underline{0.42} 0.83 0.38  & 0.14 1.00 0.13  & 0.03 0.43 0.04 &  0.00 0.00 0.00 & 0.02 0.25 0.05 & 0.07 0.36 0.13 & \textbf{0.49} & 0.26 \\
			\cline{3-11}
			& & \textbf{$\varnothing$ Total} &  \underline{0.42} 0.83 0.38  & 0.14 1.00 0.13  & 0.03 0.43 0.04 & 0.00 0.00 0.00 & 0.02 0.25 0.05 & 0.07 0.36 0.13 & \textbf{0.49} & 0.26 \\
			\cline{2-11}
			& \multirow{2}{*}{\rot{Hor.}} & Car & \underline{0.48} 0.87 0.50 & 0.08 1.00 0.07  & 0.13 0.60 0.21 & 0.00 0.00 0.00 & 0.00 0.00 0.00 & 0.11 0.50 0.14 & \textbf{0.61} & 0.35 \\
			\cline{3-11}
			& & \textbf{$\varnothing$ Total} & \underline{0.48} 0.87 0.50 & 0.08 1.00 0.07  & 0.13 0.60 0.21 & 0.00 0.00 0.00 & 0.00 0.00 0.00 & 0.11 0.50 0.14 & \textbf{0.61} & 0.35 \\
			\hline
		\end{tabular}
		\label{table:rgb_eb_performance}
	\end{center}
\end{table*}

\begingroup
\begin{figure*}[!h]
	\centering
	\setlength{\tabcolsep}{2pt} 
	\begin{tabular}{rcccccc}
		& \#1 & \#2 & \#3 & \#4 & \#5 & \#6 \\
		\rot{\hspace{6pt}RGB (Ours)} & 
		\includegraphics[width=2.82cm]{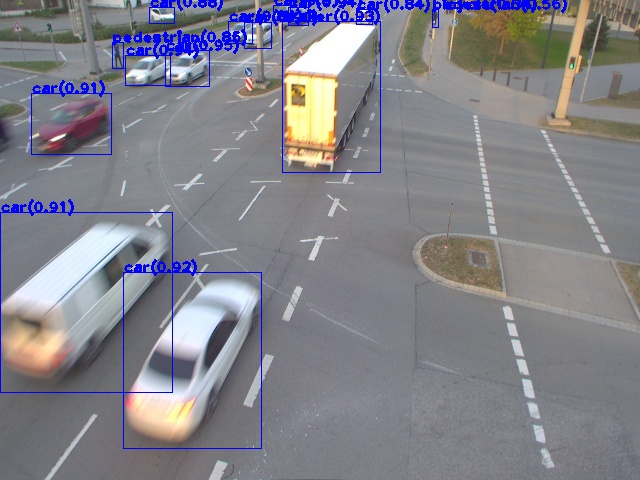} &
		\includegraphics[width=2.82cm]{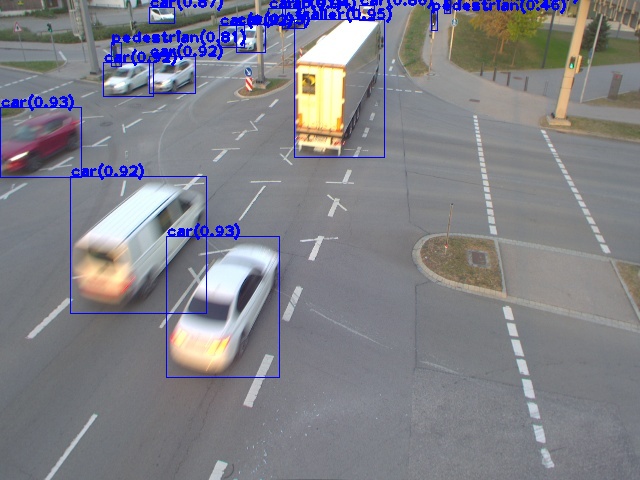} &
		\includegraphics[width=2.82cm]{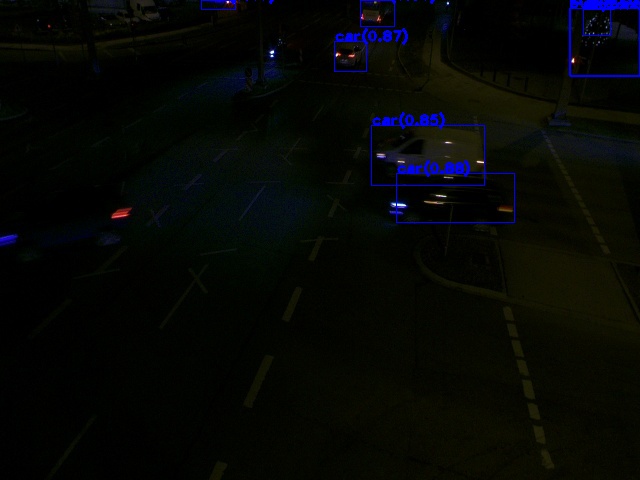} &
		\includegraphics[width=2.82cm]{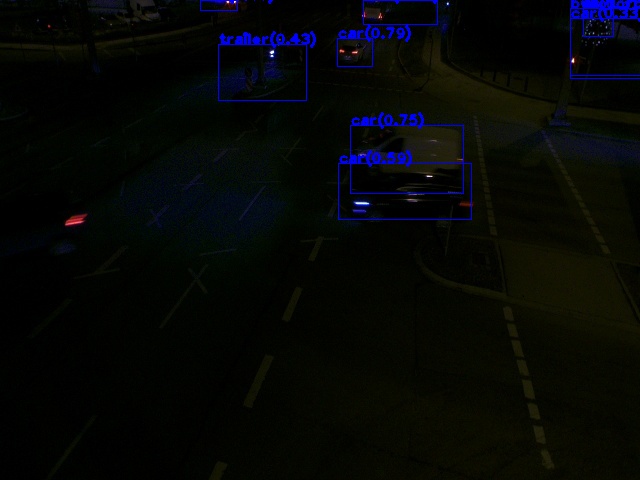} &
		\includegraphics[width=2.82cm]{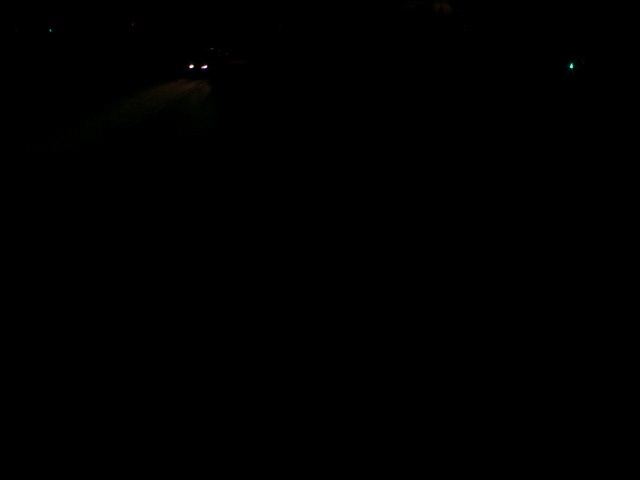} &
		\includegraphics[width=2.82cm]{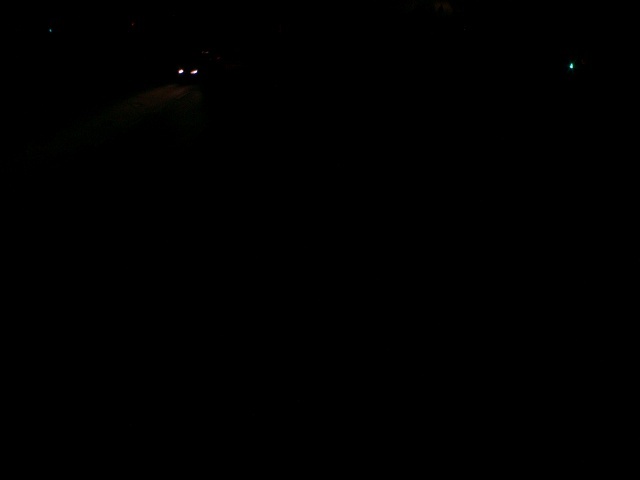} \\
		\rot{\hspace{9pt}EB (Ours)} & 
		\includegraphics[width=2.82cm]{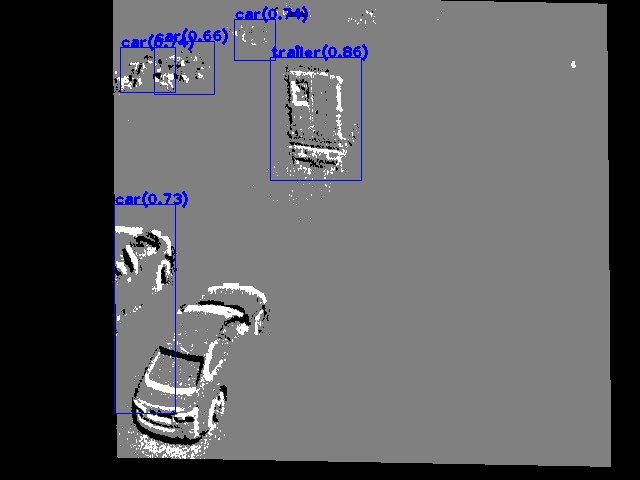} &
		\includegraphics[width=2.82cm]{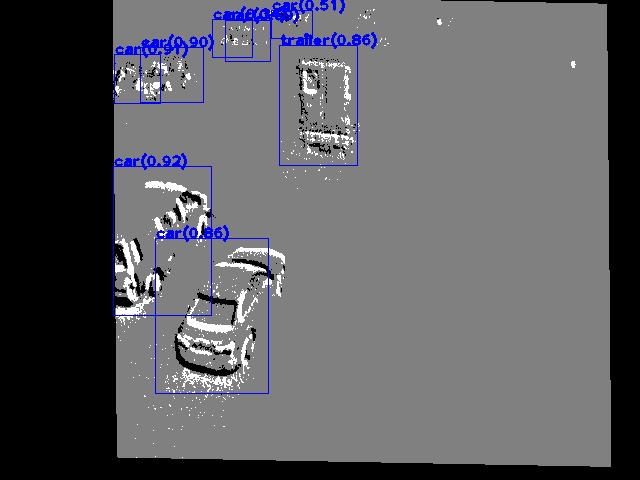} &
		\includegraphics[width=2.82cm]{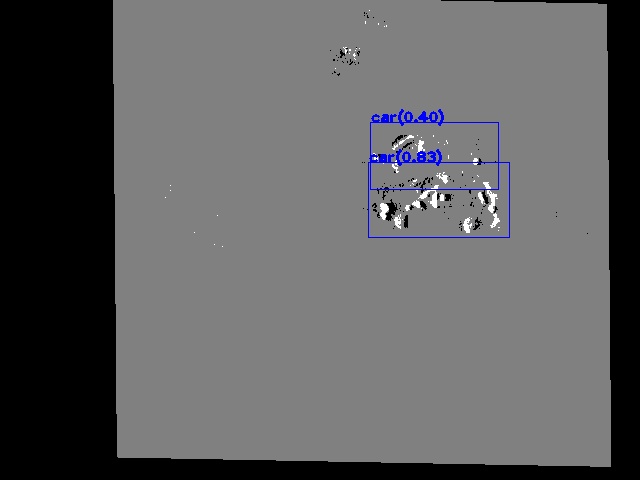} &
		\includegraphics[width=2.82cm]{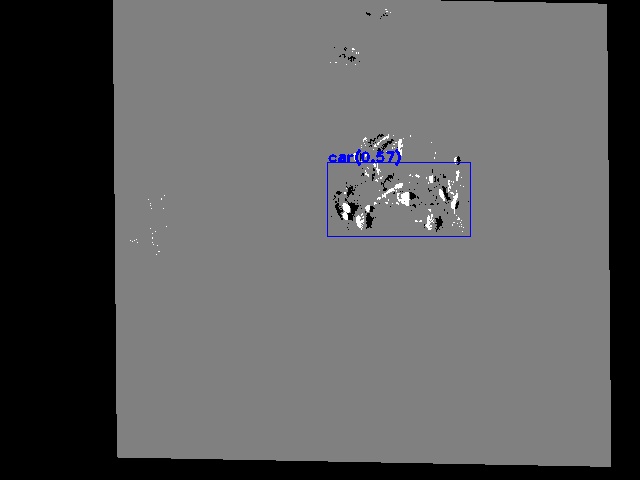} &
		\includegraphics[width=2.82cm]{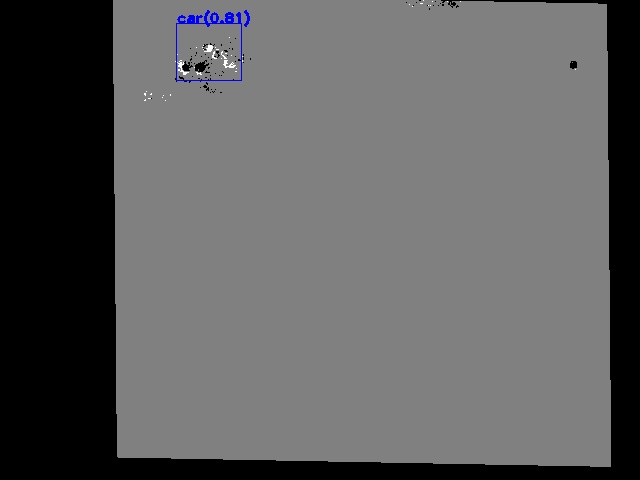} &
		\includegraphics[width=2.82cm]{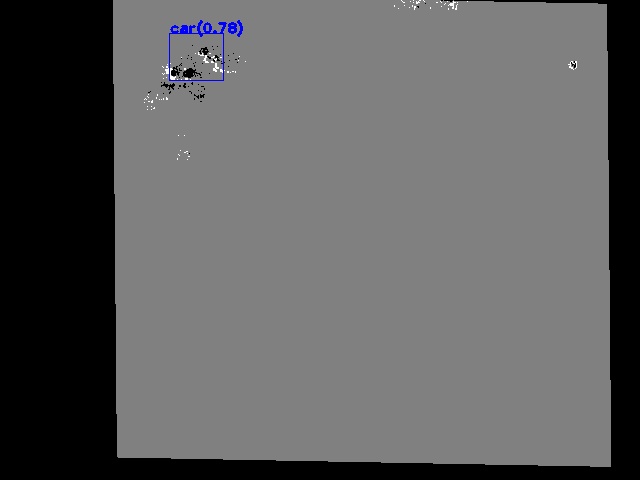} \\
		\rot{\hspace{11pt}EB-DSEC} & 
		\includegraphics[width=2.82cm]{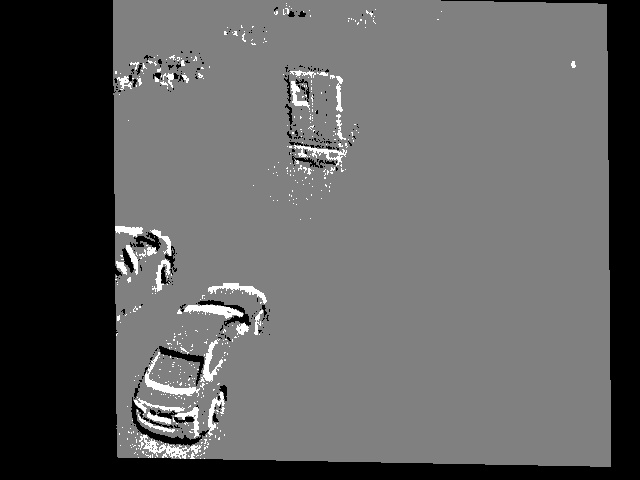} &
		\includegraphics[width=2.82cm]{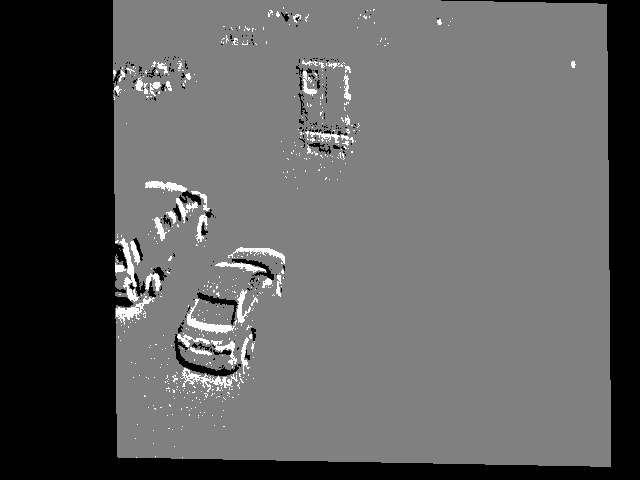} &
		\includegraphics[width=2.82cm]{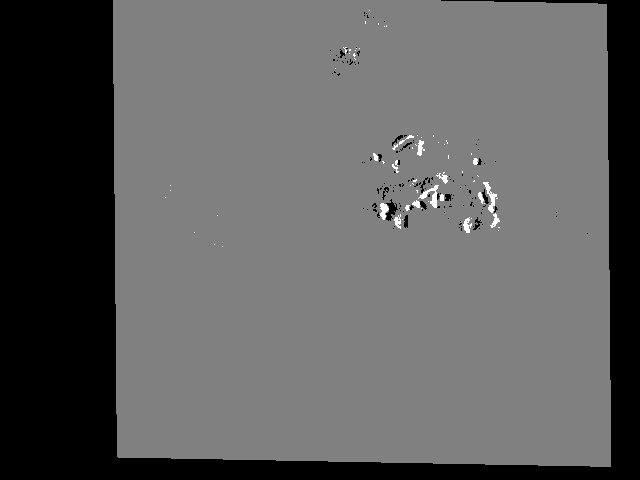} &
		\includegraphics[width=2.82cm]{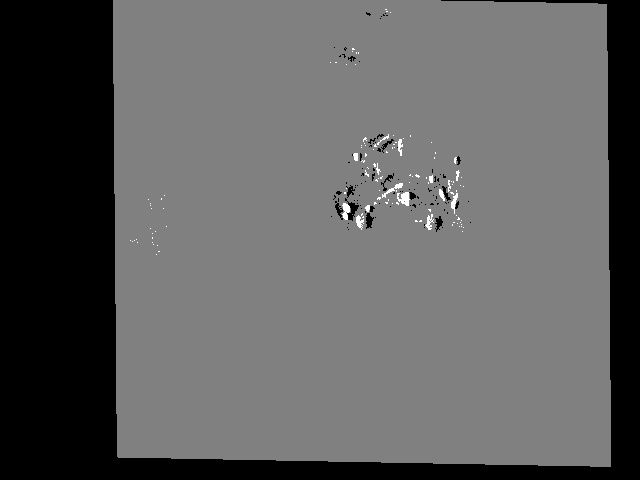} &
		\includegraphics[width=2.82cm]{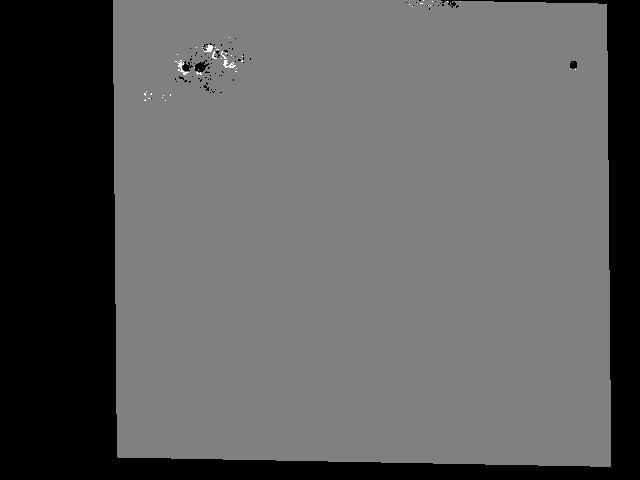} &
		\includegraphics[width=2.82cm]{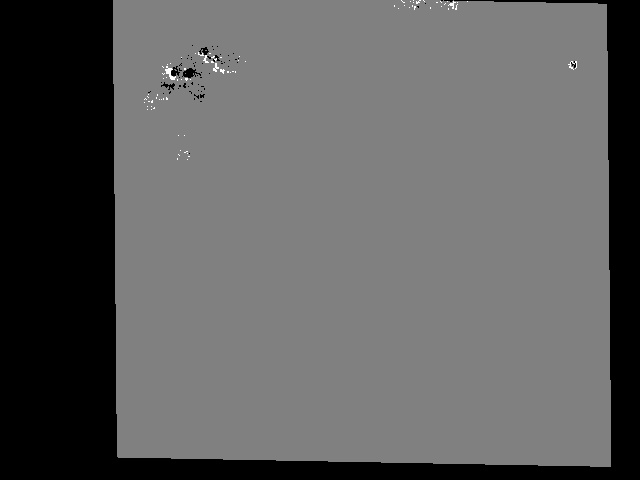} \\
		\rot{\hspace{15pt}IFCNN} & 
		\includegraphics[width=2.82cm]{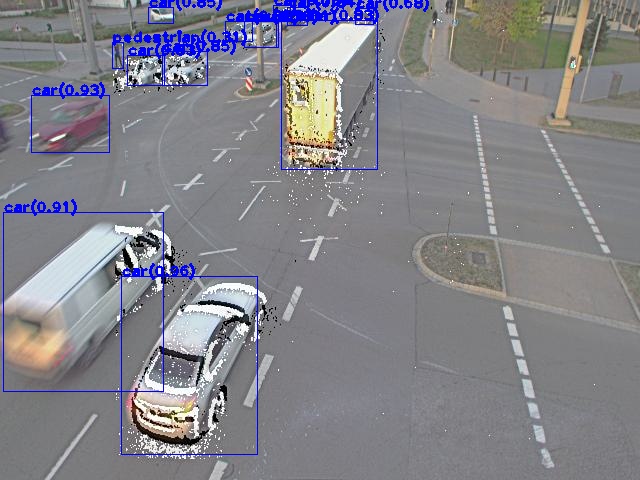} &
		\includegraphics[width=2.82cm]{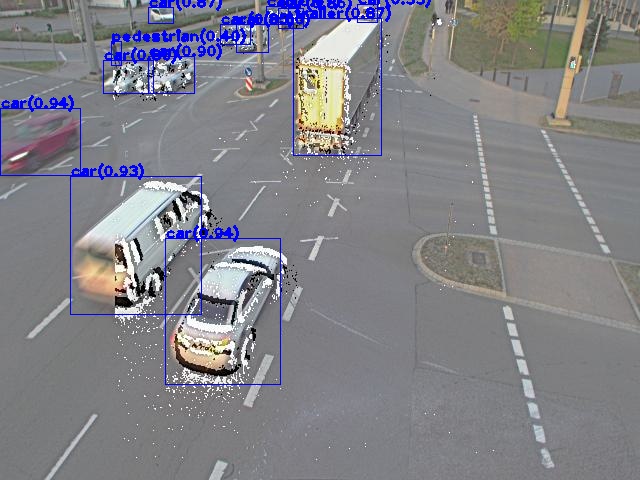} &
		\includegraphics[width=2.82cm]{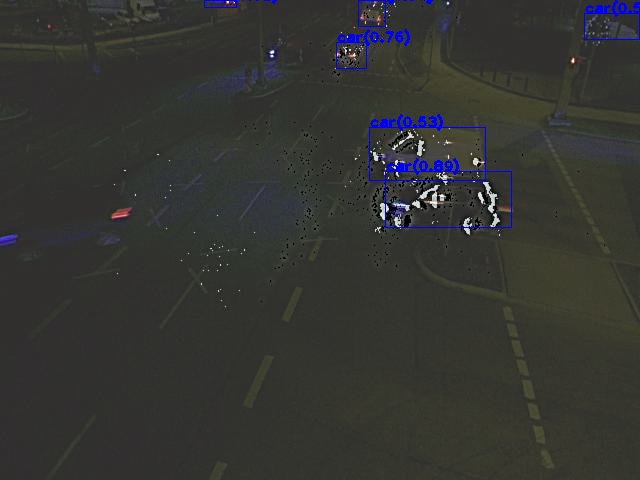} &
		\includegraphics[width=2.82cm]{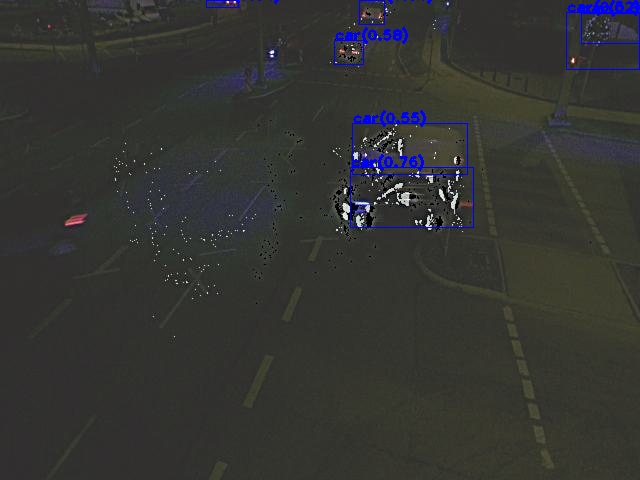} &
		\includegraphics[width=2.82cm]{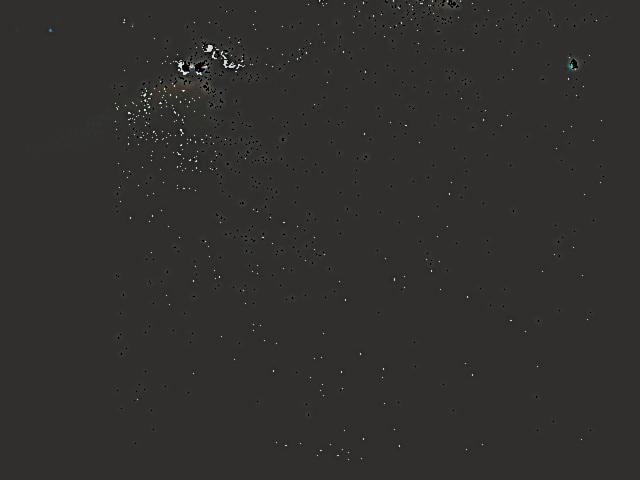} &
		\includegraphics[width=2.82cm]{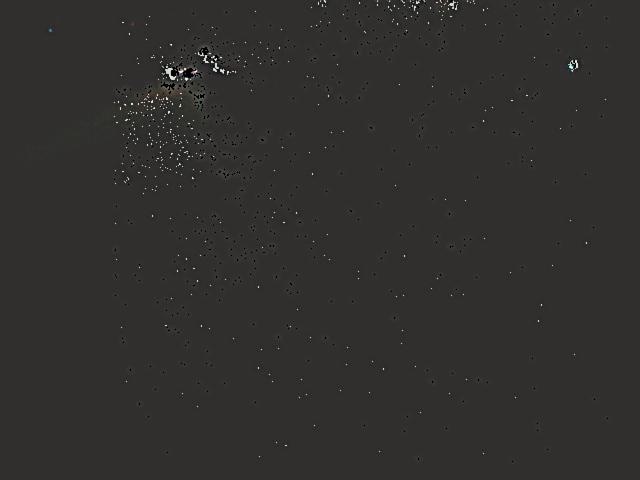} \\
		\rot{\hspace{11pt}EF-DSEC} & 
		\includegraphics[width=2.82cm]{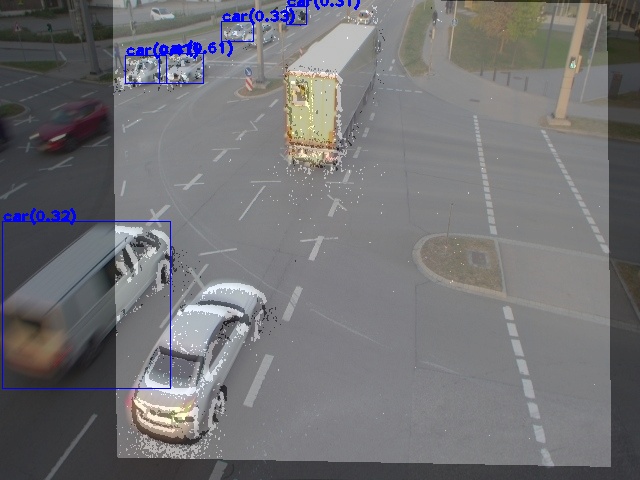} &
		\includegraphics[width=2.82cm]{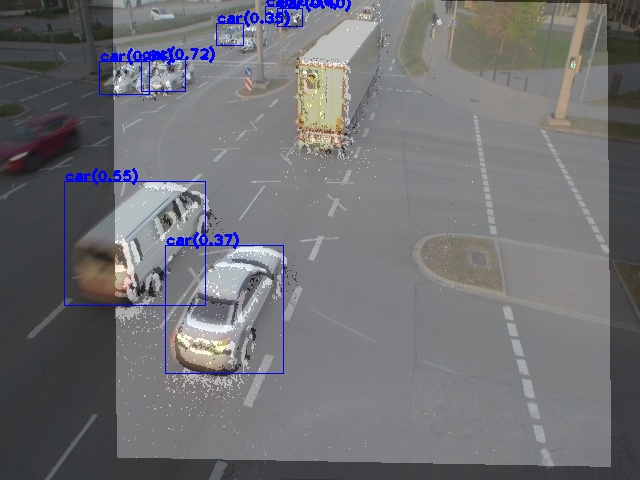} &
		\includegraphics[width=2.82cm]{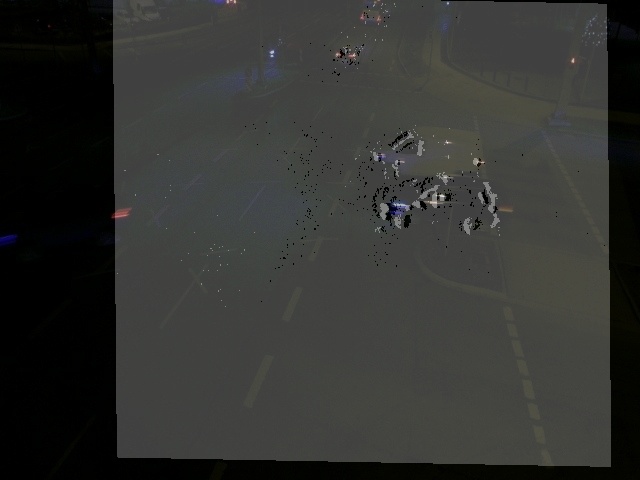} &
		\includegraphics[width=2.82cm]{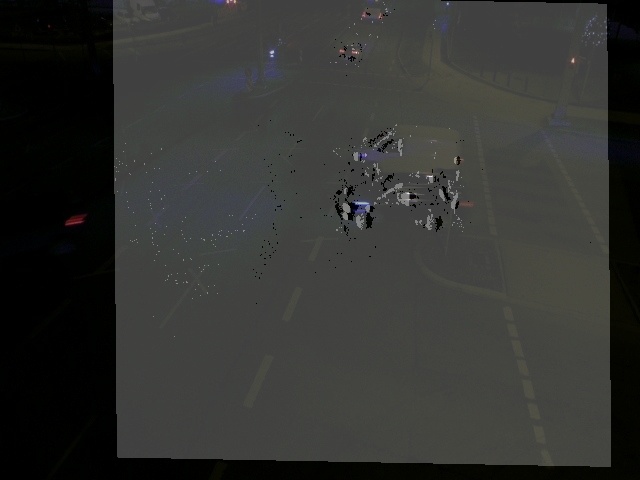} &
		\includegraphics[width=2.82cm]{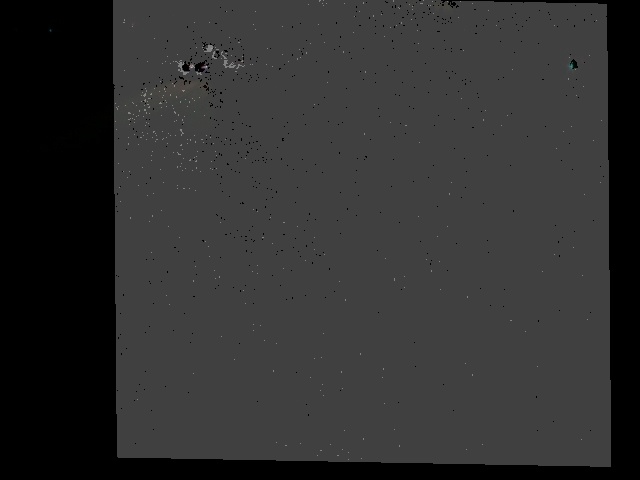} &
		\includegraphics[width=2.82cm]{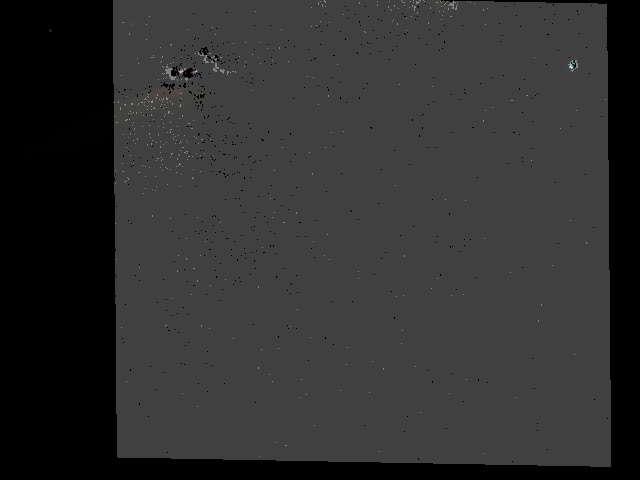} \\
		\rot{\hspace{5pt}EF-1 (Ours)} & 
		\includegraphics[width=2.82cm]{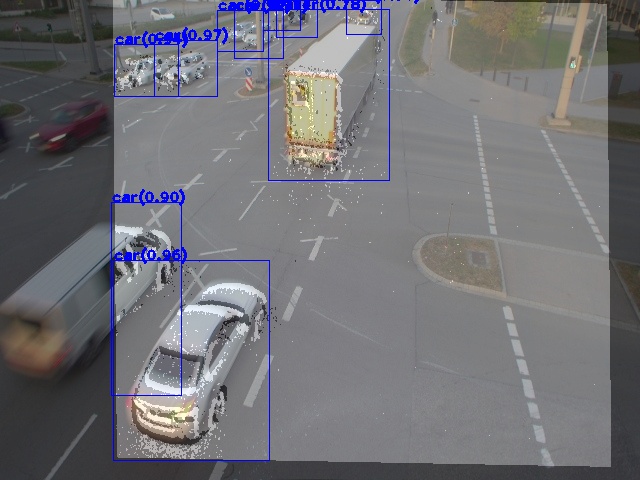} &
		\includegraphics[width=2.82cm]{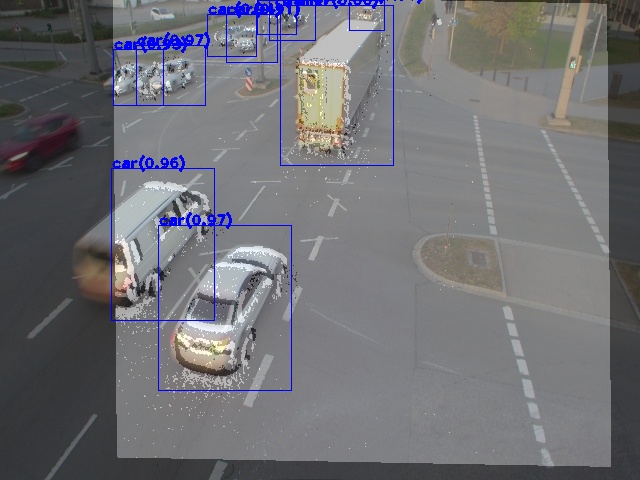} &
		\includegraphics[width=2.82cm]{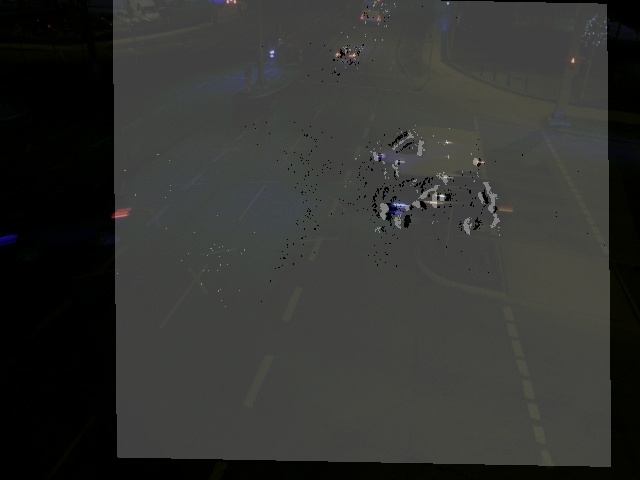} &
		\includegraphics[width=2.82cm]{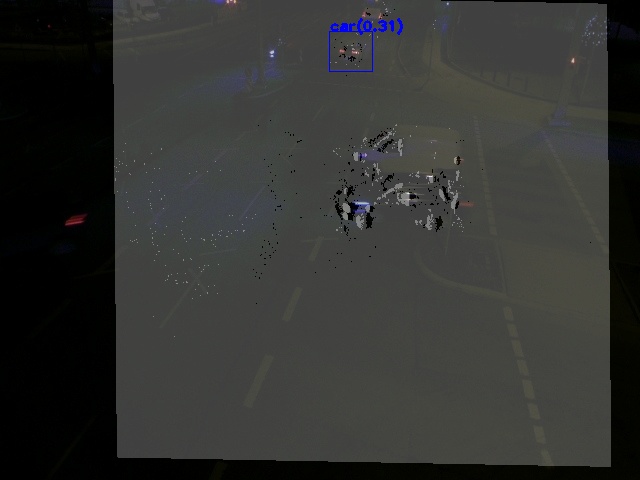} &
		\includegraphics[width=2.82cm]{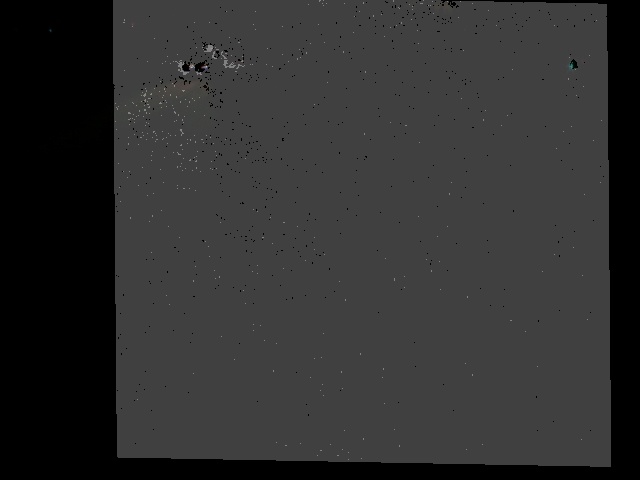} &
		\includegraphics[width=2.82cm]{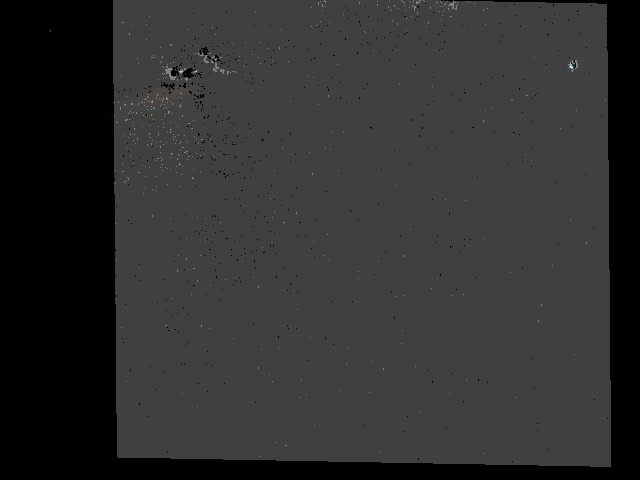} \\
		\rot{\hspace{5pt}EF-2 (Ours)} & 
		\includegraphics[width=2.82cm]{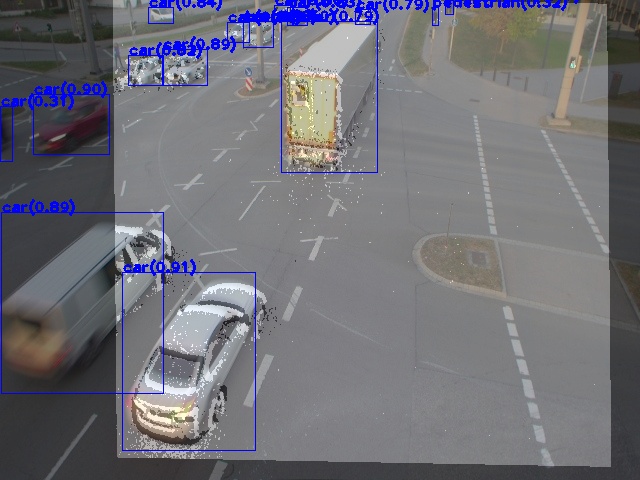} &
		\includegraphics[width=2.82cm]{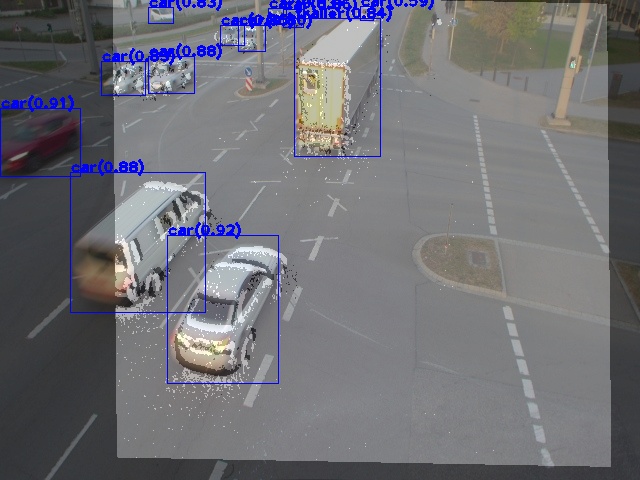} &
		\includegraphics[width=2.82cm]{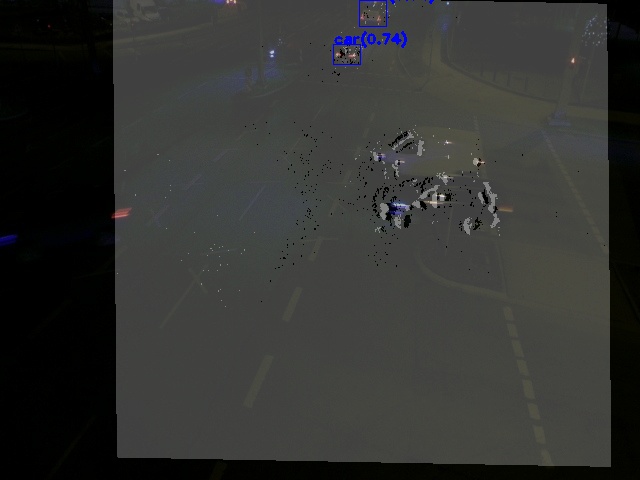} &
		\includegraphics[width=2.82cm]{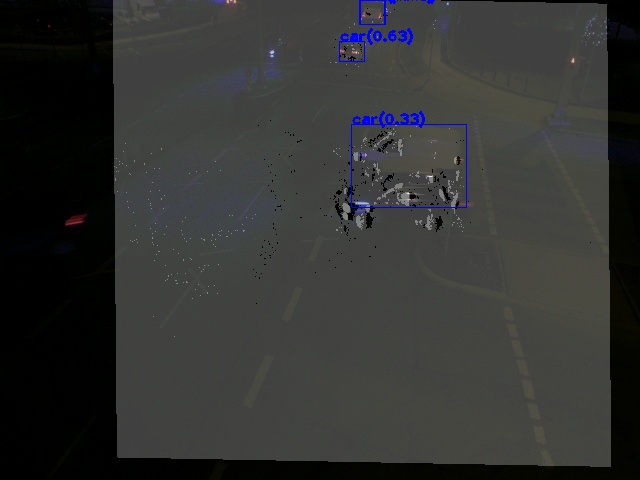} &
		\includegraphics[width=2.82cm]{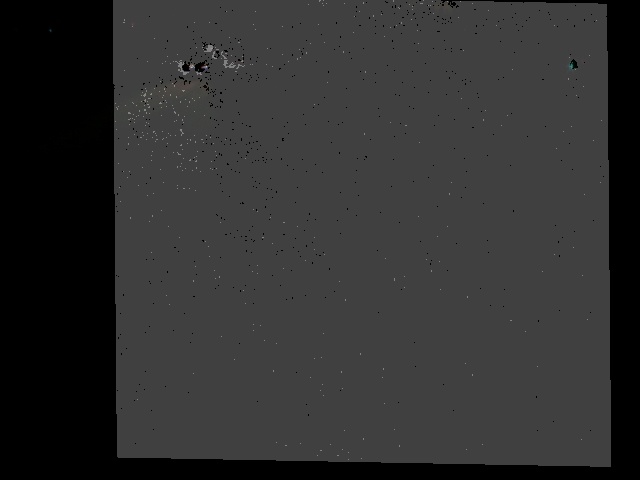} &
		\includegraphics[width=2.82cm]{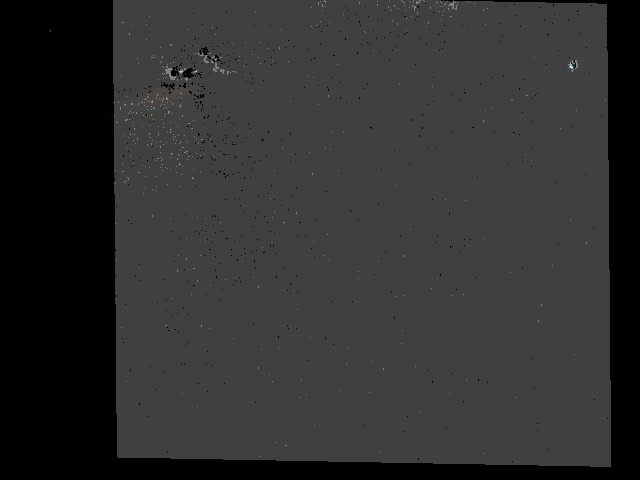} \\
		\rot{\hspace{7pt}SLF (Ours)} & 
		\includegraphics[width=2.82cm]{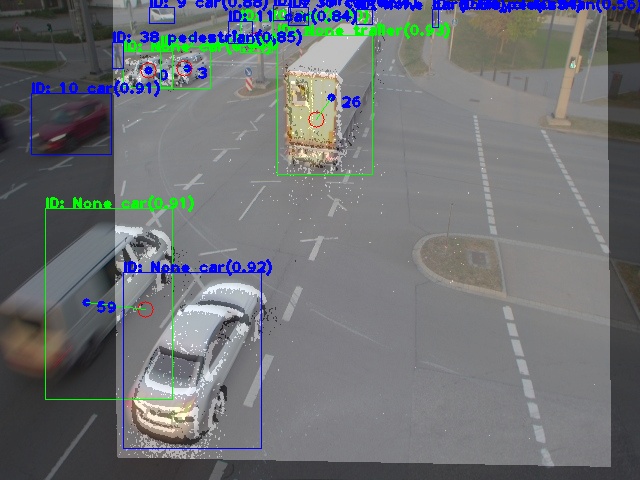} &
		\includegraphics[width=2.82cm]{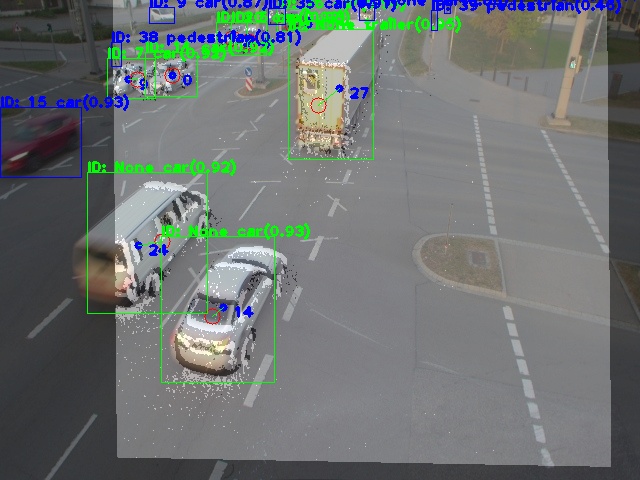} &
		\includegraphics[width=2.82cm]{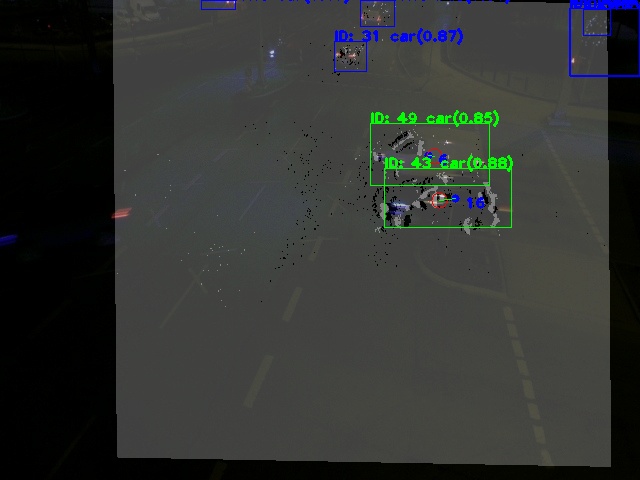} &
		\includegraphics[width=2.82cm]{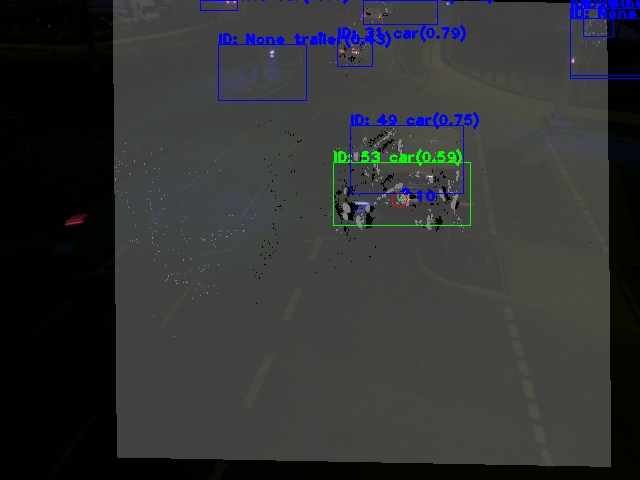} &
		\includegraphics[width=2.82cm]{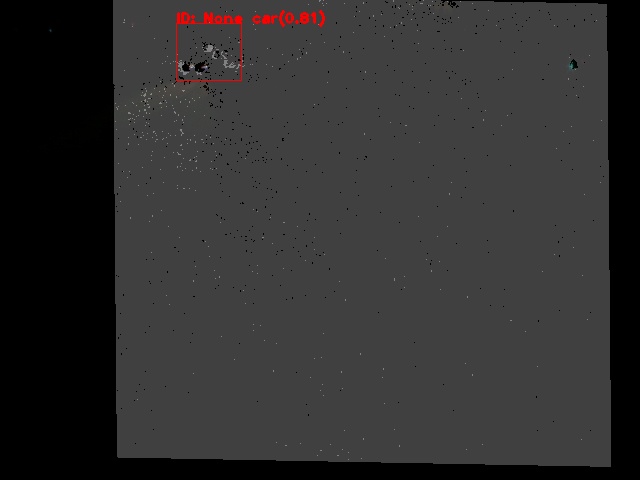} &
		\includegraphics[width=2.82cm]{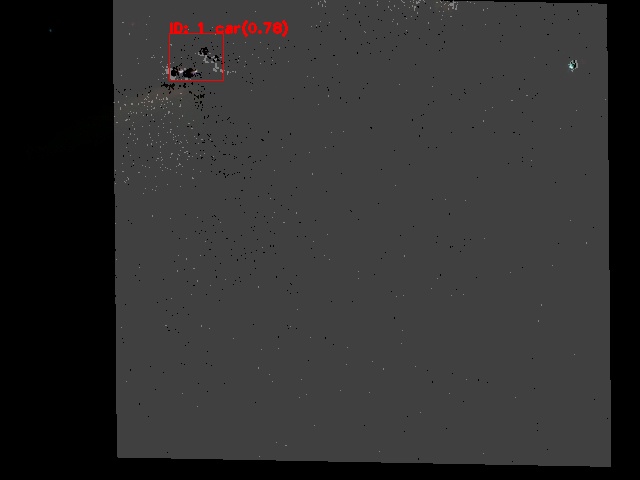} \\
		\rot{\hspace{5pt}STLF (Ours)} & 
		\includegraphics[width=2.82cm]{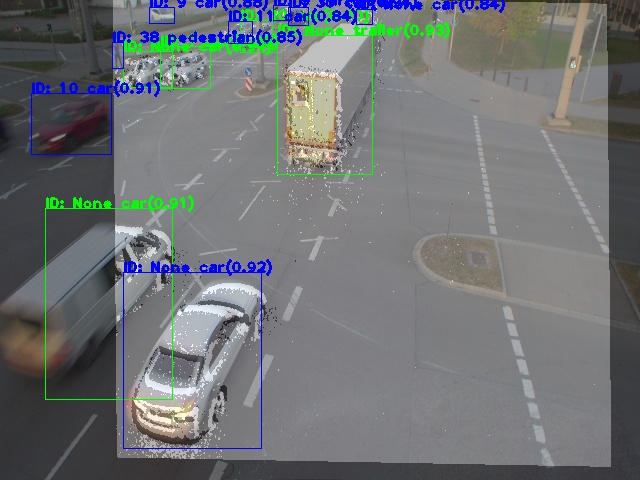} &
		\includegraphics[width=2.82cm]{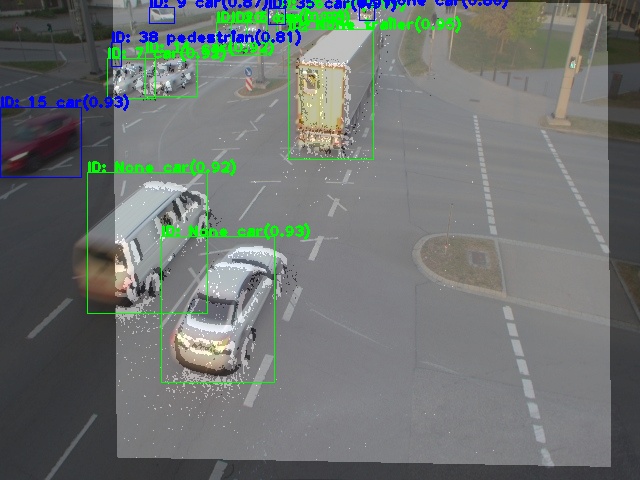} &
		\includegraphics[width=2.82cm]{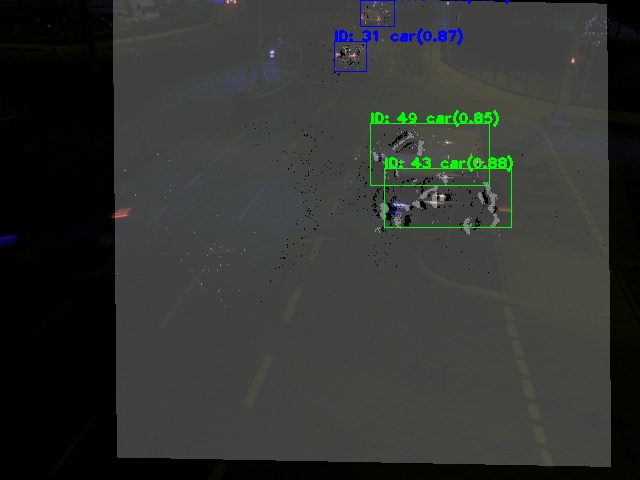} &
		\includegraphics[width=2.82cm]{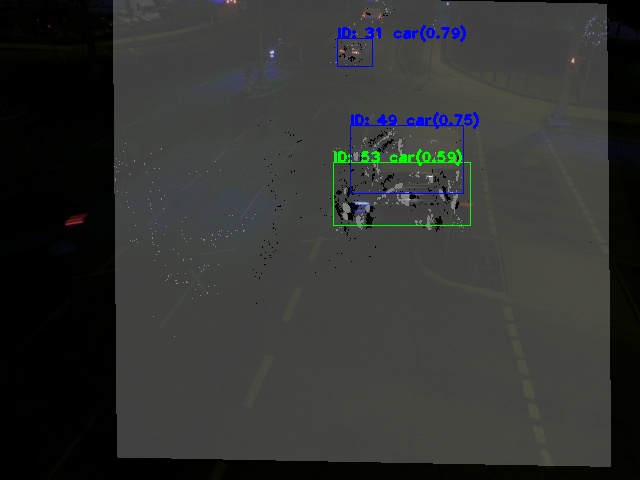} &
		\includegraphics[width=2.82cm]{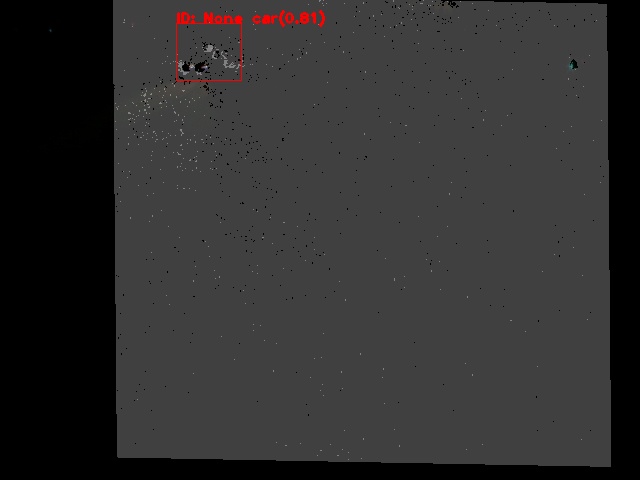} &
		\includegraphics[width=2.82cm]{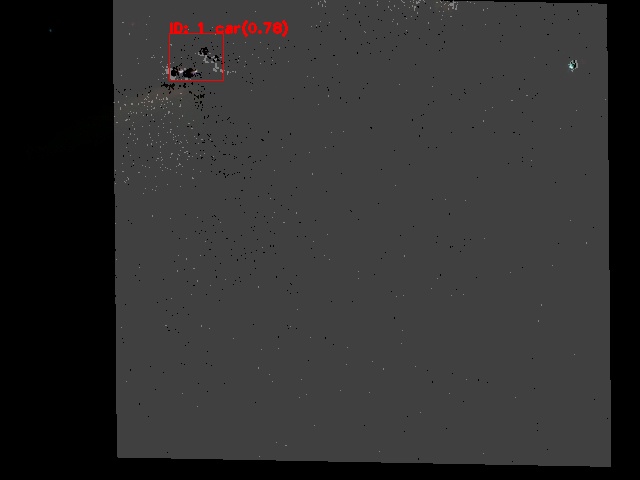} \\
	\end{tabular}
	\caption{This figure compares the results of the RGB and EB detectors and our fusion methods EF-1 and EF-2, SLF, and STLF with EB-DSEC and EF-DSEC, based on the DSEC-Detection Dataset \cite{Gehrig.2021, Gehrig.22112022, DSEC-Detection.27022024} and the IFCNN \cite{Zhang.2020} fusion. The experiments were on the ``Day'' (\#1--\#2), ``night with street lights on'' (\#3--\#4), and ``night with street lights off'' (\#5--\#6). Blue boxes in RGB, EB, IFCN \cite{Zhang.2020}, and EF are detections. In SLF and STLF, green boxes show detections by an event-based and RGB camera; a blue box indicates detection only by the RGB, and a red box only by the event-based camera. STLF successfully filtered false positives caused by the RGB camera in frames \#3 and \#4. The advantages of the event-based camera in darkness are visible. Thresholds: Confidence $ = 0.3$; IoU $ = 0.45$.}
	\label{fig:qualitative_fusion}
\end{figure*}
\endgroup

For comparison, we additionally trained an event-based detector with the DSEC-Detection Dataset {\cite{Gehrig.2021, Gehrig.22112022, DSEC-Detection.27022024} (EB-DSEC). For this purpose, we rendered the event-based data as a grayscale image to have the same color code for event-based data as in the TUMTraf Event Dataset. Interestingly, EB-DSEC cannot reliably detect objects in all three scenarios. We suspect the domain gap of perspective: The training data consider just the ego perspective of a vehicle. The arrangement of the edges differs significantly compared to the roadside perspective from a height of $7$ m; meaningful recognition seems impossible. This result underlines the relevance and the need for the TUMTraf Event Dataset. 

In the next step, we evaluated our early and late fusion methods EF-1 and EF-2, SLF, and STLF. We compared our results with the RGB detector with two confidence thresholds as a baseline and the fusion methods IFCNN \cite{Zhang.2020} and EF-DSEC, where we trained an early-fusion based on the DSEC-Detection Dataset {\cite{Gehrig.2021, Gehrig.22112022, DSEC-Detection.27022024}. To prepare the IFCNN \cite{Zhang.2020} evaluation, we fused our complete TUMTraf Event Dataset with IFCNN \cite{Zhang.2020} and trained and tested a YoloV7 \cite{arXiv.org.07072022b, Wong.14012023, Chen.02012020} detector. Here, we used as ground truth ``L-RGB'' labels on our carefully manually corrected test set, which includes the scenarios ``Day'', ``N-1'' (``night with street lights on''), and ``N-2'' (``night with street lights off''). We also examined various traffic scenarios. For this purpose, if available, we have extracted from the three illumination scenarios the subcategories ``Standing,'' ``Vertical,'' and ``Horizontal,'' which contain a specific dominant traffic flow. The quantitative results are in Table \ref{table:rgb_eb_performance}, and the qualitative results are in Figure \ref{fig:qualitative_fusion}. 

In the scenario ``Day,'' the RGB detector achieved an mAP of $0.69$, particularly with the object classes car an AP of $0.94$ and bus the highest AP of $0.98$. Nevertheless, we also recognized worse performance with optically small objects, e.g., pedestrians ($0.64$ AP), motorcycles ($0.37$ AP), or bicycles ($0.73$ AP). Here, less texture information and less training data in comparison to the class car are attributed to these results. If we applied a high confidence threshold of $0.80$ in the RGB detector, we increased the precision from $0.72$ to $0.84$, but the mAP dropped to $0.44$. The fusion with IFCNN \cite{Zhang.2020} achieved a mAP of $0.45$, for the class car an AP of $0.84$, and for the bus $0.99$. In total, we were worse with IFCNN \cite{Zhang.2020} compared to our RGB baseline: The visually smaller classes pedestrian ($0.22$ AP), bicycle ($0.04$ AP), and motorcycle ($0.05$ AP) couldn't be recognized by IFCNN \cite{Zhang.2020}. The early fusion methods EF-DSEC and EF-1 didn't produce satisfactory results. They could only detect large visual buses with an AP of $0.67$ and $0.61$. The other classes weren't detectable. We suspect that the domain gap in the camera perspective leads to insufficient results for EF-DSEC. However, the achieved mAP of $0.50$ from EF-2 is significantly better than EF-1 and EF-DSEC, and object detection of cars ($0.89$ AP) and buses ($0.94$ AP) was reliable. Furthermore, in Figure \ref{fig:qualitative_fusion}, we noticed that EF-1 cannot detect objects that are not in the view of both sensors. We conclude that considering objects in the training process that are not recognizable by a specific sensor domain is important. In the next step, we want to discuss our late fusion methods: SLF significantly outperformed all other fusion methods with its $0.78$ mAP and the RGB baseline with more than $+0.09$ mAP. The recognition rate is high in all object classes. This result shows the strengths of the event-based and RGB cameras when combined. Furthermore, our late fusion STLF achieved an mAP of $0.59$. Therefore, it outperforms the early fusion methods. The described results of all listed methods can tend to be observed in all traffic scenarios of the scenario ``Day''.

Scenario ``N-1'' was recorded at night with street lights on, including the car and bus object classes. Despite a noticeable drop in recall, we can still detect cars ($0.71$ AP) and buses ($0.80$ AP) with the RGB detector. Interestingly, as shown in Figure \ref{fig:qualitative_fusion}, false positive detections occur significantly more often with the RGB detector. Using a higher confidence threshold could solve this problem but also result in less recall. A fusion between an event-based and RGB camera could help. As in the scenario ``Day,'' EF-DSEC and EF-1 didn't allow proper object detection. Although EF-2 performed significantly better than EF-1, we achieved just $0.38$ AP for the class car and $0.27$ AP for the class bus and, therefore, had significant losses compared to the ``Day.'' In addition, EF-2 didn't achieve the performance of IFCNN \cite{Zhang.2020}, which can detect cars with $0.47$ AP and buses with $0.39$ AP. Compared to the RGB detector, our SLF method achieved equivalent detection performance for the class car with an AP of $0.70$ and an adequate AP of $0.62$ for the class bus. Although the STLF method reached a lower mAP of $0.42$, we could effectively eliminate false positives, see Figure \ref{fig:qualitative_fusion}. 

The scenario ``N-2'' is the most challenging for object detection. We recorded this scenario in complete darkness with no street lights at around 2:00 am. Due to the low traffic volume, this subset only contains passing cars. Compared to the scenario ``Day,'' the performance of the RGB detector dropped for cars from $0.94$ AP to $0.43$ AP. The IFCNN fusion \cite{Zhang.2020}, EF-DSEC, EF-1, and EF-2 didn't allow the detection of passing cars, too. According to Table \ref{table:eb_performance} and Figure \ref{fig:qualitative_fusion}, the event-based detector EB could still deliver adequate or better detection results than the RGB detector. Since our late fusion methods, SLF and STLF, in principle, represent a logical OR of both sensors, we could achieve significant improvements in this challenging illumination scenario: The SLF achieved $0.49$ mAP during the dark night and, therefore, slightly better than the RGB detector. Furthermore, we recognize higher detection performance in horizontal traffic scenarios. The SLF outperformed in this case with more than $+0.13$ mAP the RGB detector and achieved an mAP of $0.61$. However, STLF reaches there a lower mAP value of $0.35$.

Interestingly, in almost all scenarios, the performance of STLF is slightly weaker compared to the SLF method. This effect occurs due to the fact that STLF accepts only objects that either have at least a confidence threshold in the RGB camera detection, e.g., of $C = 0.77$, or have been recognized over time by the event-based camera. In challenging illumination conditions, e.g., at night, where the RGB confidence value is lower, predominantly moving objects are more considered for fusion. Therefore, STLF is particularly suitable for highly dynamic traffic situations, e.g., on a motorway. Figure \ref{fig:comparision_late_fusion_simple_spatiotemporal} shows this effect: Cars are slightly less detected in STLF, but higher precision values are guaranteed. This result is particularly noticeable in scenarios with challenging illumination, where the RGB camera can no longer deliver high confidence values. For example, SLF's minimum possible precision value in scenario ``N-1'' is $0.76$, whereas STLF's is $0.90$. With it, the algorithm prevents false positives, see Figure \ref{fig:qualitative_fusion}. 

\begin{figure*}[h!]
	\begin{tikzpicture}
	\begin{axis}[
		title = {Day},
		xlabel = {Recall}, 
		ylabel= {Precision},
		xmin = 0.0, ymin = 0.74, xmax = 1.0,
		xtick = {0.00, 0.2, ..., 1.00},
		scaled ticks=false,
		ticklabel style={
			/pgf/number format/fixed,
			/pgf/number format/1000 sep={}
		},
	    x tick label style={rotate=60},
		legend style={at={(0.03,0.2)},anchor=west},
		scale only axis,
		width=0.24\linewidth,
		height=5cm,
		]
		
		\addplot[color=black] table [x=r, y=p, col sep=comma] {day-simple-class-2.csv};
		\addlegendentry{SLF}
		
		\addplot[color=red, dash pattern=on 7pt off 3pt] table [x=r, y=p, col sep=comma] {day-filtered-class-2.csv};
		\addlegendentry{STLF}
		
	\end{axis}
\end{tikzpicture}
	\begin{tikzpicture}
	\begin{axis}[
		title = {N-1},
		xlabel = {Recall}, 
		ylabel= {Precision},
		xmin = 0.0, ymin = 0.74, , xmax = 1.0,
	    xtick = {0.00, 0.2, ..., 1.00},
		scaled ticks=false,
		ticklabel style={
			/pgf/number format/fixed,
			/pgf/number format/1000 sep={}
		},
	    x tick label style={rotate=60},
		legend style={at={(0.03,0.2)},anchor=west},
		scale only axis,
		width=0.24\linewidth,
		height=5cm,
		]
		
		\addplot[color=black] table [x=r, y=p, col sep=comma] {night-with-light-simple-class-2.csv};
		\addlegendentry{SLF}
		
		\addplot[color=red, dash pattern=on 7pt off 3pt] table [x=r, y=p, col sep=comma] {night-with-light-filtered-class-2.csv};
		\addlegendentry{STLF}

	\end{axis}
\end{tikzpicture}
	\begin{tikzpicture}
	\begin{axis}[
		title = {N-2},
		xlabel = {Recall}, 
		ylabel= {Precision},
		xmin = 0.0, ymin = 0.74, xmax = 1.0,
		xtick = {0.00, 0.2, ..., 1.00},
		scaled ticks=false,
		ticklabel style={
			/pgf/number format/fixed,
			/pgf/number format/1000 sep={}
		},
	    x tick label style={rotate=60},
		legend style={at={(0.03,0.2)},anchor=west},
		scale only axis,
		width=0.24\linewidth,
		height=5cm,
		]
		
		\addplot[color=black] table [x=r, y=p, col sep=comma] {night-dark-simple-class-2.csv};
		\addlegendentry{SLF}
		
		\addplot[color=red, dash pattern=on 7pt off 3pt] table [x=r, y=p, col sep=comma] {night-dark-filtered-class-2.csv};
		\addlegendentry{STLF}

	\end{axis}
\end{tikzpicture}
	\caption{The effect of Spatiotemporal Late Fusion (STLF) becomes clear on the precision-recall curve for the class ``car'': Because STLF gives preferential treatment to higher confidence values or objects detected by the event-based camera, higher precision values are guaranteed. Since not every non-moving object is recognized in difficult visibility conditions, e.g., at night, the recall drops in such situations. On the other hand, Simple Late Fusion (SLF) enables higher recall values even in difficult visibility conditions with the compromise of lower precision values. We chose the class ``car'' because it is available in all illumination scenarios.}
	\label{fig:comparision_late_fusion_simple_spatiotemporal}
\end{figure*}
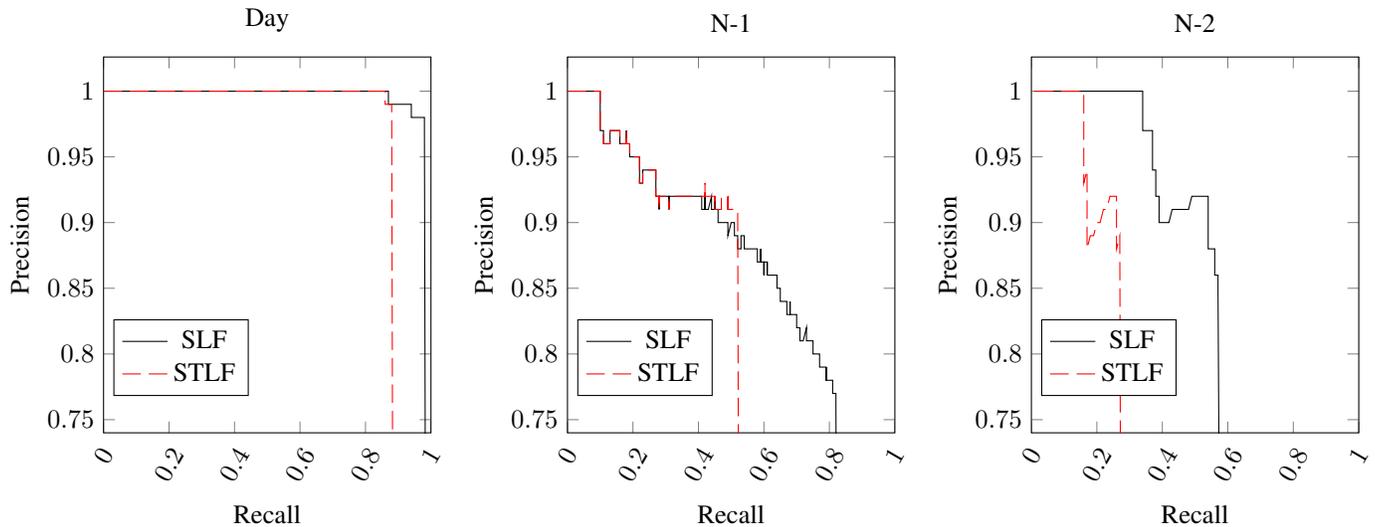 

\begin{figure*}[h!]
	\begin{tikzpicture}
	\begin{axis}[
		title = {Day},
		xlabel = {Confidence Threshold}, 
		ylabel= {Average Precision},
		xmin = 30, ymin = 0.00, xmax = 90, ymax = 1.00,
		xtick = {30, 40, ..., 90},
		scaled ticks=false,
		ticklabel style={
			/pgf/number format/fixed,
			/pgf/number format/1000 sep={}
		},
		x tick label style={rotate=60},
		legend style={at={(0.02,0.26)},anchor=west},
		scale only axis,
		width=0.24\linewidth,
		height=6.5cm,
		]
		
		\addplot[color=cyan, dash pattern=on 8pt off 2pt] coordinates {
			(30, 0.71) (40, 0.65) (50, 0.58) (60, 0.50) (70, 0.40) (80, 0.17) (90, 0.04)
		};
		\addlegendentry{Pedestrian}
		
		\addplot[color=red, dashed] coordinates {
			(30, 0.74) (40, 0.72) (50, 0.69) (60, 0.66) (70, 0.58) (80, 0.46) (90, 0.28)
		};
		\addlegendentry{Bicycle}
		
		\addplot[color=black] coordinates {
			(30, 0.95)	(40, 0.94)	(50, 0.93)	(60, 0.91)	(70, 0.88)	(80, 0.85)	(90, 0.66)	
		};
		\addlegendentry{Car}
		
		\addplot[color=orange, dashed, thick] coordinates {
			(30, 0.75)	(40, 0.71)	(50, 0.66)	(60, 0.66)	(70, 0.62) (80, 0.43) (90, 0.29)			
		};
		\addlegendentry{Motorcycle}
		
		\addplot[color=blue, ultra thick] coordinates {
			(30, 0.98)	(40, 0.98)	(50, 0.98)	(60, 0.98)	(70, 0.96)	(80, 0.95)	(90, 0.95)	
		};
		\addlegendentry{Bus}
		
		\addplot[color=green, densely dotted, ultra thick] coordinates {
			(30, 0.63)	(40, 0.62)	(50, 0.61)	(60, 0.59)	(70, 0.55)	(80, 0.42)	(90, 0.17)
		};
		\addlegendentry{Truck}
		
		\addplot[color=gray, densely dotted, thick] coordinates {
			(30, 0.71)	(40, 0.68)	(50, 0.65)	(60, 0.63)	(70, 0.57)	(80, 0.41)	(90, 0.32)
		};
		\addlegendentry{Trailer}

	\end{axis}
\end{tikzpicture}
	\begin{tikzpicture}
	\begin{axis}[
		title = {N-1},
		xlabel = {Confidence Threshold}, 
		ylabel= {Average Precision},
		xmin = 30, ymin = 0.00, xmax = 90, ymax = 1.00,
		xtick = {30, 40, ..., 90},
		scaled ticks=false,
		ticklabel style={
			/pgf/number format/fixed,
			/pgf/number format/1000 sep={}
		},
		x tick label style={rotate=60},
		legend style={at={(0.02,0.26)},anchor=west},
		scale only axis,
		width=0.24\linewidth,
		height=6.5cm,
		]
				
		\addplot[color=black] coordinates {
			(30, 0.70)	(40, 0.66)	(50, 0.62)	(60, 0.58)	(70, 0.49)	(80, 0.37)	(90, 0.30)
		};
		\addlegendentry{Car}
				
		\addplot[color=blue, ultra thick] coordinates {
			(30, 0.62)	(40, 0.62)	(50, 0.59)	(60, 0.57)	(70, 0.55)	(80, 0.42) (90, 0.17)
		};
		\addlegendentry{Bus}		
		
	\end{axis}
\end{tikzpicture}
	\begin{tikzpicture}
	\begin{axis}[
		title = {N-2},
		xlabel = {Confidence Threshold}, 
		ylabel= {Average Precision},
		xmin = 30, ymin = 0.00, xmax = 90, ymax = 1.00,
		xtick = {30, 40, ..., 90},
		scaled ticks=false,
		ticklabel style={
			/pgf/number format/fixed,
			/pgf/number format/1000 sep={}
		},
		x tick label style={rotate=60},
		legend style={at={(0.02,0.26)},anchor=west},
		scale only axis,
		width=0.24\linewidth,
		height=6.5cm,
		]
		
		\addplot[color=black] coordinates {
			(30, 0.49)	(40, 0.47)	(50, 0.40)	(60, 0.37) (70, 0.31)	(80, 0.21)	(90, 0.19)
			
		};
		\addlegendentry{Car}

	\end{axis}
\end{tikzpicture}
	\caption{This figure shows the average precision (AP) of the Spatiotemporal Late Fusion (STLF) for several confidence thresholds $C$ in the illumination scenarios Day, N-1, and N-2. The detection performance on the day for most classes decreases at $C = 70$. However, there is no performance drop for buses; the performance drop for cars occurs at $C = 80$. In the scenarios N-1, the performance drop starts at $C = 70$. A linear decline of the performance is recognizable in N-2.}
	\label{fig:confidence_treshold_ablation_study}
\end{figure*}
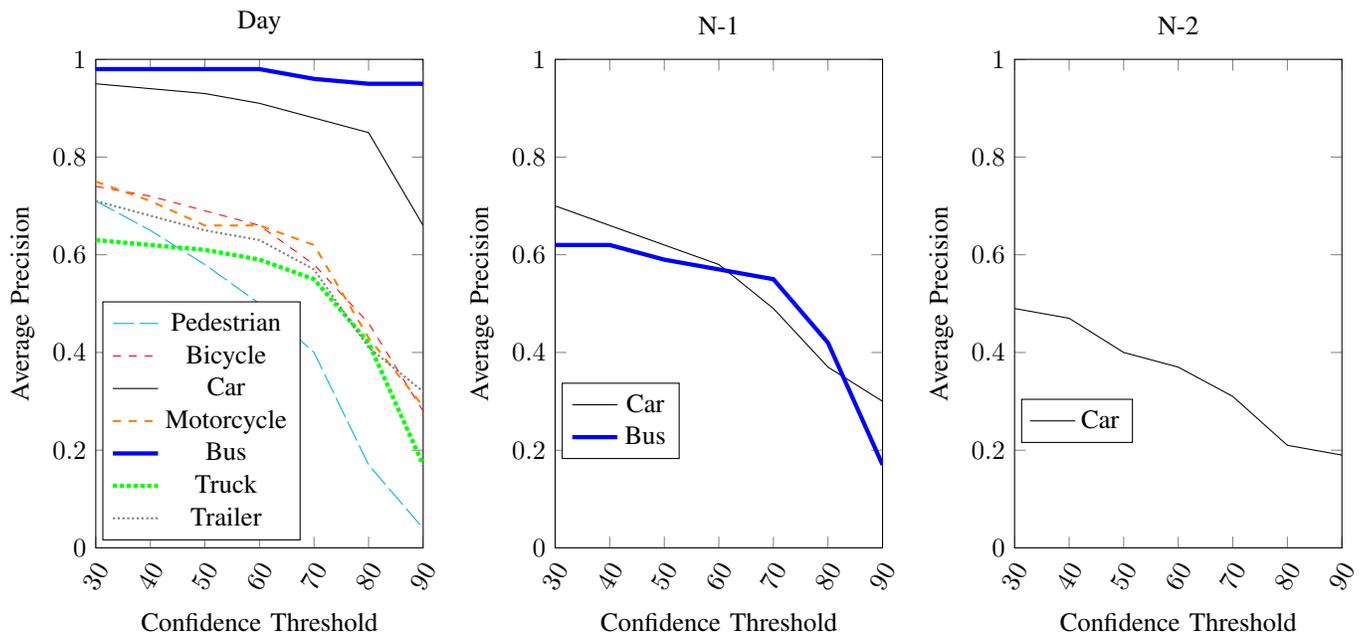

The question arises regarding which confidence threshold $C$ offers the best detection performance with maximum filtering. For investigation, we analyzed the average precision (AP) of each available object class with different confidence values in the illumination scenarios Day, N-1, and N-2, see Figure \ref{fig:confidence_treshold_ablation_study}. We observed that on the Day, the performance drop for cars occurred at $C = 80$. However, there is no drop for the class bus. For most other classes, the collapse started at $C = 70$. These results are plausible because, due to their geometric size, cars and buses are much easier to recognize in the RGB camera than, for example, pedestrians or cyclists. In addition, motorized vehicles generate significantly more movement in the event-based image, which leads to more trustworthy classifications. In scenario N-1, we also recognized a drop in detection performance starting with a confidence threshold of $C = 70$. In absolute darkness, as in scenario N-2, the detections have lower confidence values, and the detection performance decreases almost linearly. According to Figure \ref{fig:qualitative_fusion}, with confidence $C = 77$, we achieved a sufficient compromise between detection performance and reduction of false positives.

In summary, the sensor fusion between event-based and RGB cameras, particularly Simple Late Fusion (SLF), offers comparable or significantly more stable detection performance than a single sensor. Here, the strengths of both sensor types are combined: The RGB camera provides color and detailed texture information, whereas the event-based camera is primarily characterized by a high dynamic range and an extremely fast response time, which reduces motion blur. Compared to a single RGB camera, we could improve the detection performance by $+0.09$ mAP during the day and $+0.13$ mAP during the dark night. Using Spatiotemporal Late Fusion (STLF), we were also able to noticeably reduce the number of false positives. Thus, the sensor fusion with the event-based camera supplements the RGB camera and ensures more stability and reliability in object detection for a roadside ITS, even in highly challenging illumination conditions.

\section{Conclusion}
\label{section:conclusion}
In this work, we have improved our previous approach from \cite{Cre.642023672023} for targetless extrinsic calibration between event-based and RGB cameras. To find image correspondences, we have extended the matching algorithm with DBSCAN clustering \cite{EsterMartinandKriegelHansPeterandSanderJorgandXuXiaowei.1996}. This enables us to handle multiple moving objects. Furthermore, we published the novel $\text{TUMTraf Event Dataset}$, which contains spatiotemporal calibrated event-based and RGB images from stationary roadside cameras in the domain of Intelligent Transportation Systems. In addition, we have created an event-based image detector and the early fusion methods EF-1 and EF-2 with our dataset. Last but not least, we have developed Simple Late Fusion and Spatiotemporal Late Fusion to combine the strengths of event-based and RGB cameras with parallel reduction of their limitations. 

We demonstrated a significant increase in the flexibility of our calibration approach, as it can now also be used in more complex scenarios with independent moving traffic participants. In addition, we have pointed out the advantages and disadvantages of event-based and RGB cameras in the domain of roadside ITS in several traffic scenarios during the day and at night. Here, we showed that our presented sensor fusion algorithms can significantly increase the detection performance compared to an RGB camera of up to +9\% mAP during the day and up to +13\% mAP at night. Furthermore, we compared our results qualitatively and quantitatively with those from other fusion methods and event-based camera datasets. In applying a roadside Intelligent Transportation System, our methods delivered significantly more convincing results, which underlines the relevance of our methods and our TUMTraf Event Dataset.

For future work, we propose significantly expanding the $\text{TUMTraf Event Dataset}$. Thus, more accurate object detection could be possible during the day and at night. This goal can be achieved in a highly scalable manner using so-called pseudo-labeling with our targetless calibration tool. Furthermore, instead of DBSCAN \cite{EsterMartinandKriegelHansPeterandSanderJorgandXuXiaowei.1996} clusterization, we propose deep-learning-based targetless calibration, which is now possible with our $\text{TUMTraf Event Dataset}$.

\section*{Acknowledgment}
This research was funded by the Federal Ministry of Education and Research of Germany in the project AUTOtech.agil, FKZ: 01IS22088U. We would like to express our gratitude for making this paper possible.

\bibliography{master.bib}
\bibliographystyle{IEEEtran}

\begin{IEEEbiography}[{\includegraphics[width=1in,height=1.25in,clip,keepaspectratio]{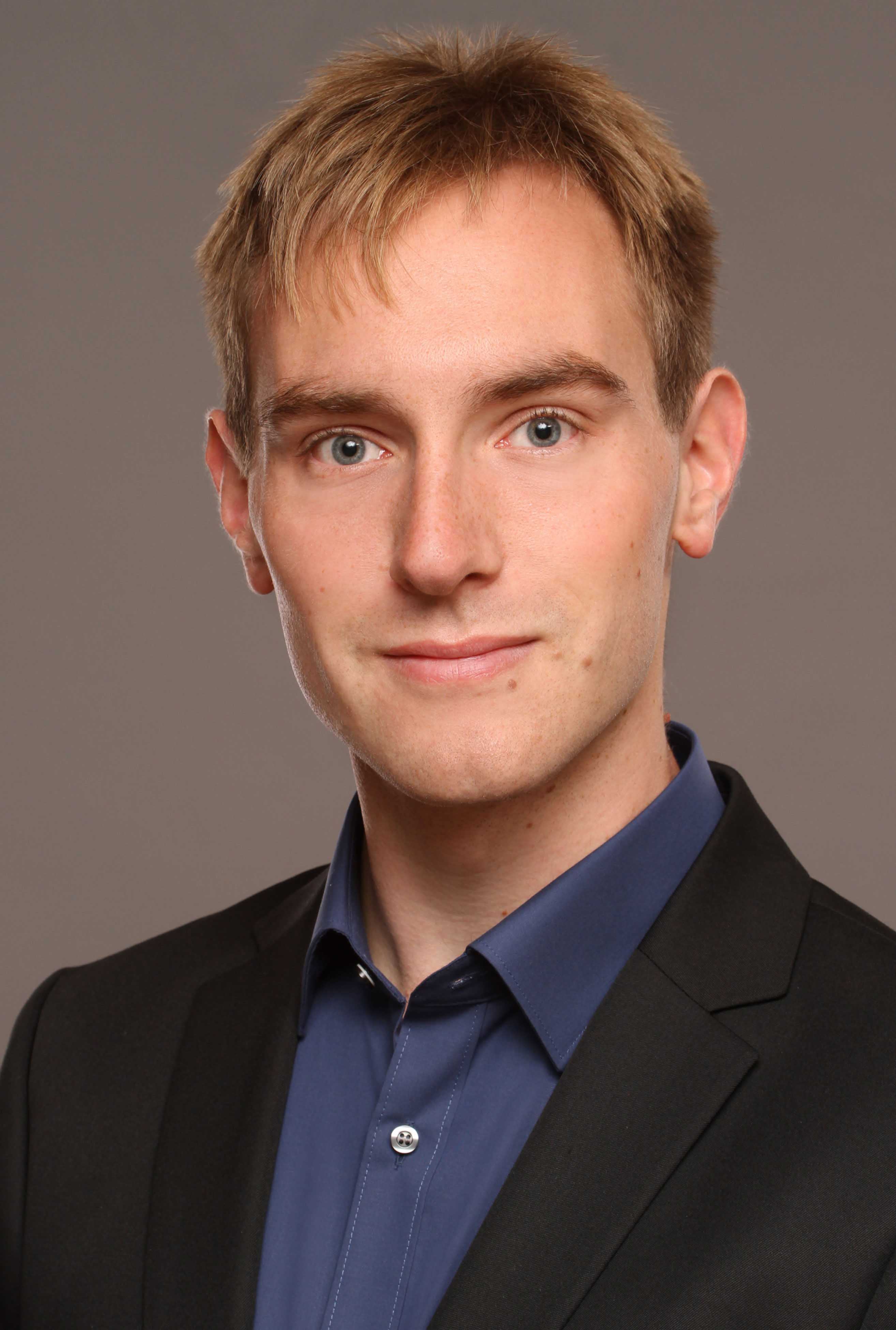}}]{Christian Creß} joined the Chair of Robotics, Artificial Intelligence, and Real-time Systems at the Technical University of Munich (TUM), Germany, in 2020 as a Research Assistant and Ph.D. student, where he is currently researching computer vision and data fusion for multi-modal sensor systems. He completed his M.Sc. in Applied Computer Science at the University of Applied Sciences Kempten in 2016. The Master's thesis was in computer vision and machine learning. His further research interests are artificial intelligence and software architecture. \end{IEEEbiography}

\begin{IEEEbiography}[{\includegraphics[width=1in,height=1.25in,clip,keepaspectratio]{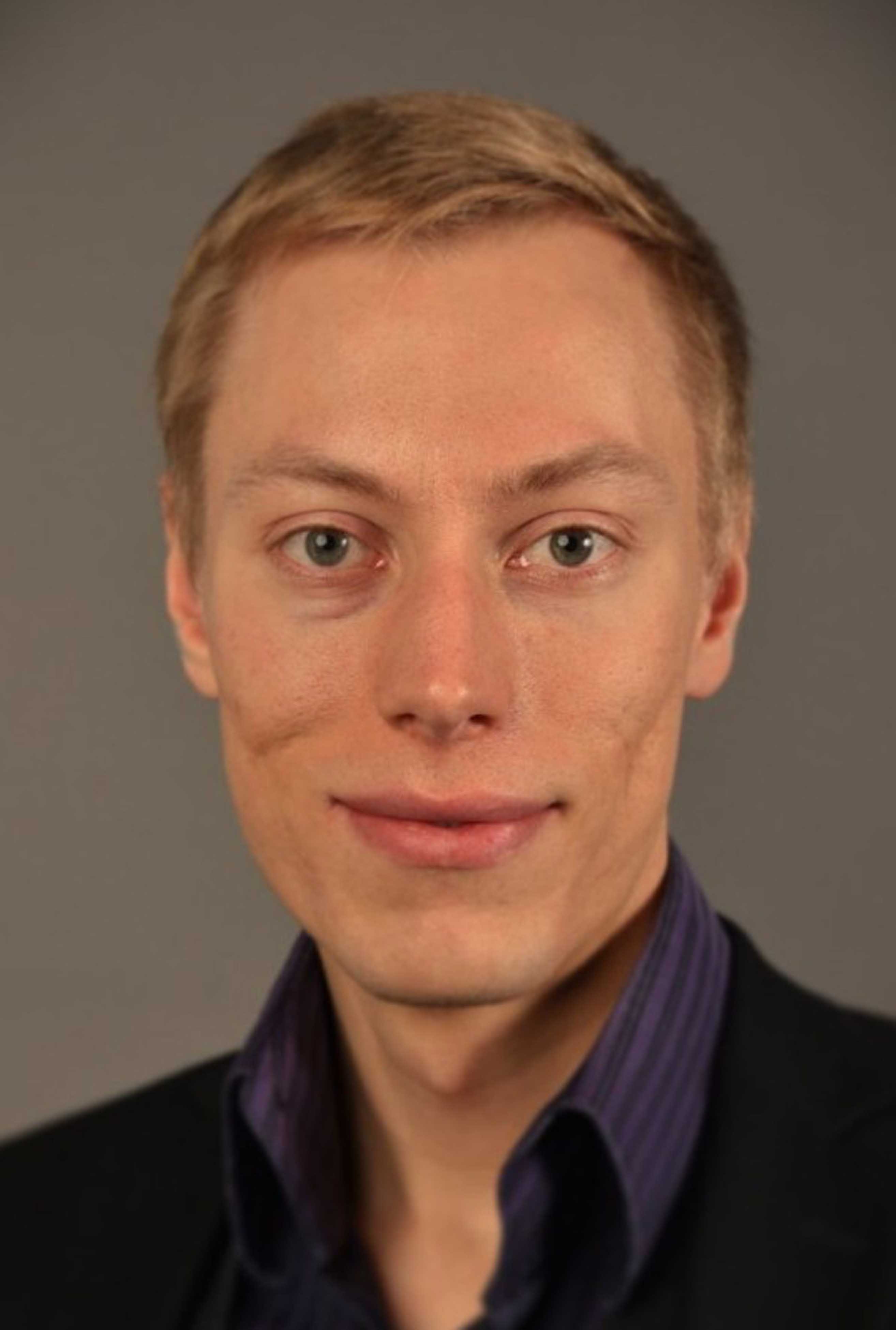}}]{Walter Zimmer} is currently a Ph.D. candidate and Research Assistant at the Chair of Robotics, Artificial Intelligence, and Real-time Systems of the Technical University of Munich (TUM). He received his M.Sc. degree in Computer Science from the TUM in 2018. During his studies, he stayed abroad at the Technical University of Delft (Netherlands) and the University of California, San Diego (UCSD), where he developed perception and autonomous driving algorithms. His research interests are 3D perception, simulation, and autonomous driving. \end{IEEEbiography}

\begin{IEEEbiography}[{\includegraphics[width=1in,height=1.25in,clip,keepaspectratio]{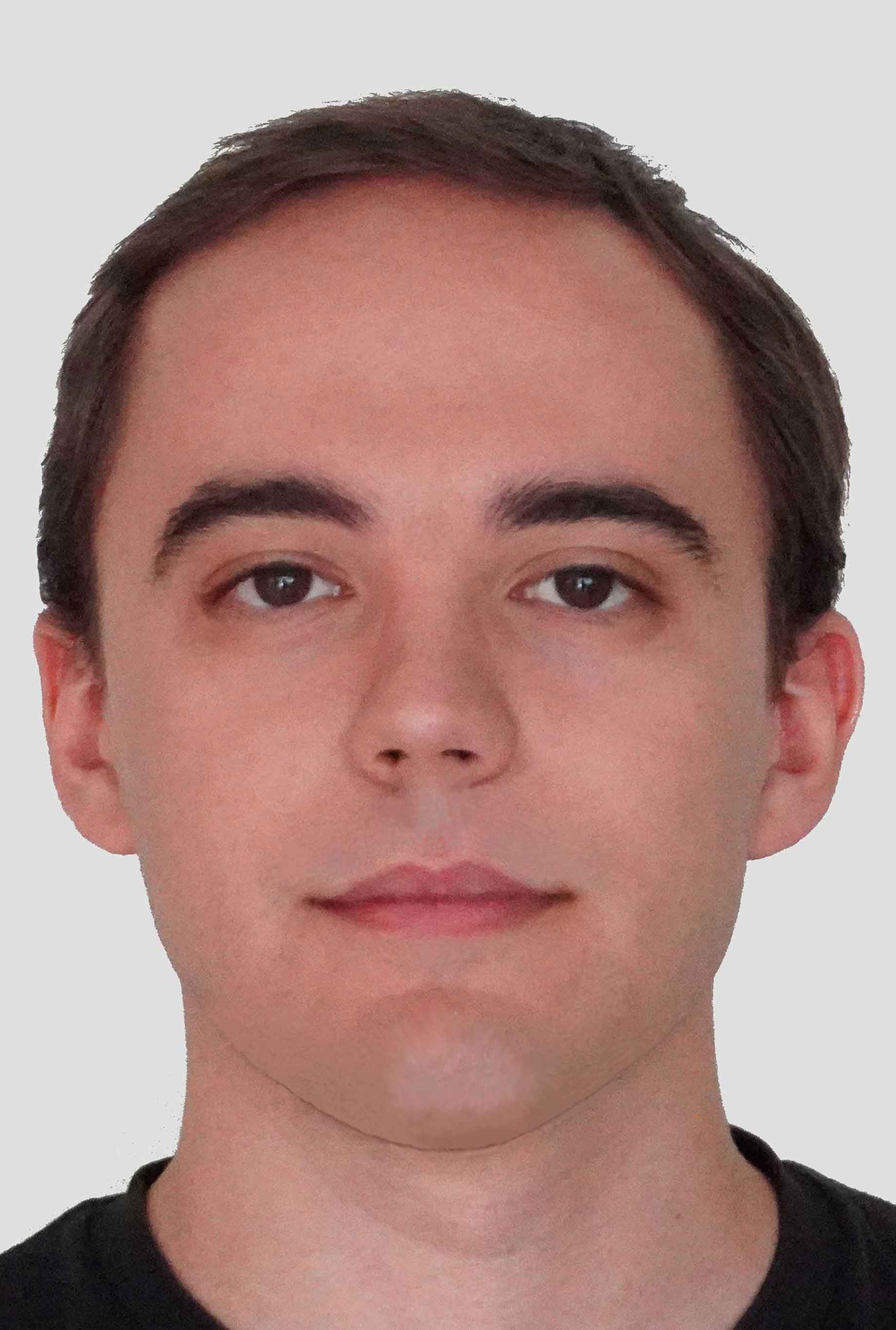}}]{Nils Purschke} is a Ph.D. student at the Technical University of Munich in Germany. He started his Ph.D. in 2023 at the Chair of Robotics, Artificial Intelligence, and Real-time Systems and conducted research in autonomous driving, vehicle software architecture, and artificial intelligence. He completed his Bachelor's degree (2021) and Master's degree (2023) at the Technical University of Darmstadt. In his Master's thesis, he created 3D digital twins. Parallel to his scientific work, he has been working as a software developer in IT security since 2020. \end{IEEEbiography}

\begin{IEEEbiography}[{\includegraphics[width=1in,height=1.25in,clip,keepaspectratio]{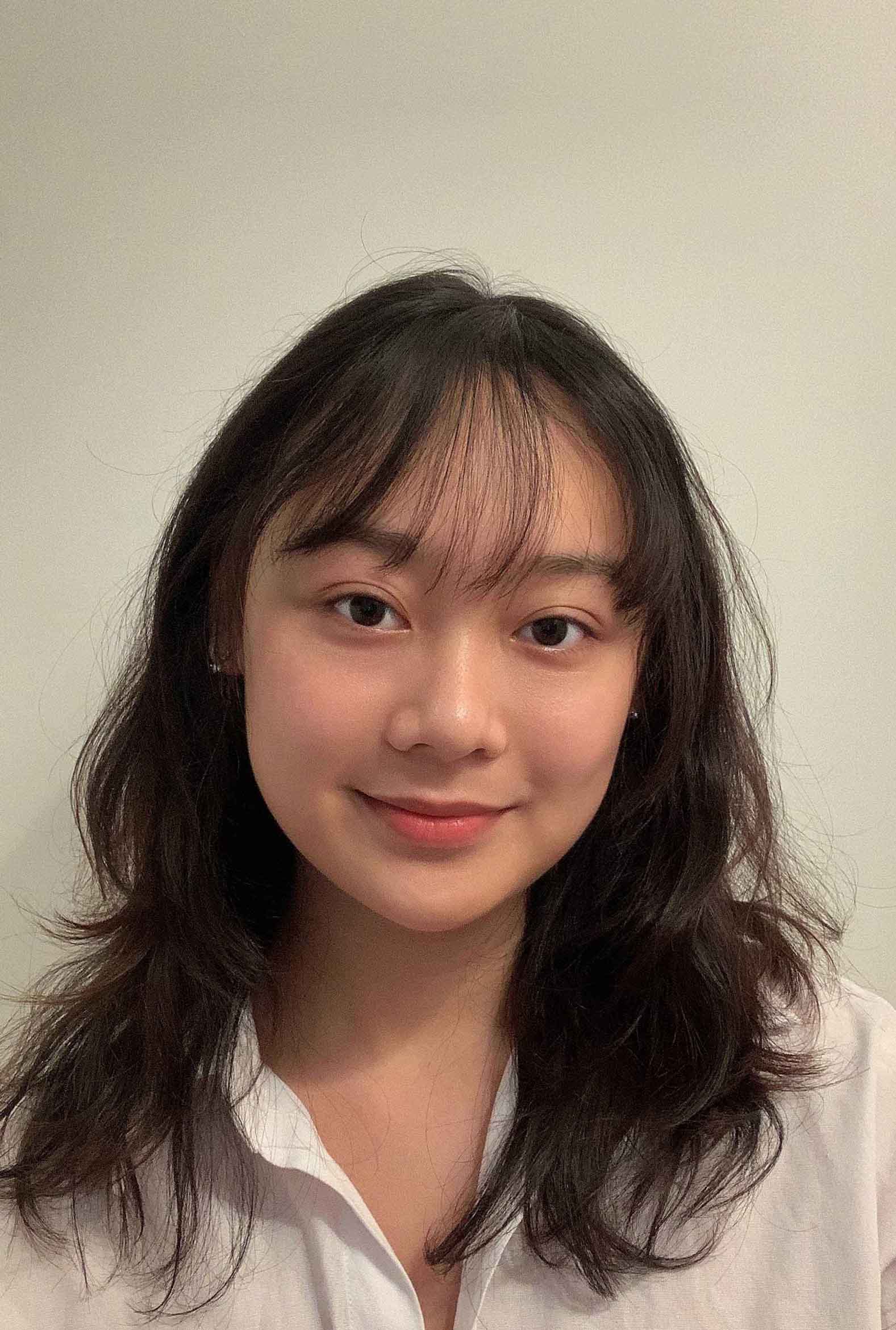}}]{Bach Ngoc Doan} began her B.Sc. in Informatics at the Technical University of Munich (TUM), Germany, in 2020. She is currently working on her Bachelor's thesis at the Chair of Robotics, Artificial Intelligence, and Real-time Systems at TUM. Her thesis is in the field of computer vision and focuses on monocular 3D perception for autonomous driving. \end{IEEEbiography}

\begin{IEEEbiography}[{\includegraphics[width=1in,height=1.25in,clip,keepaspectratio]{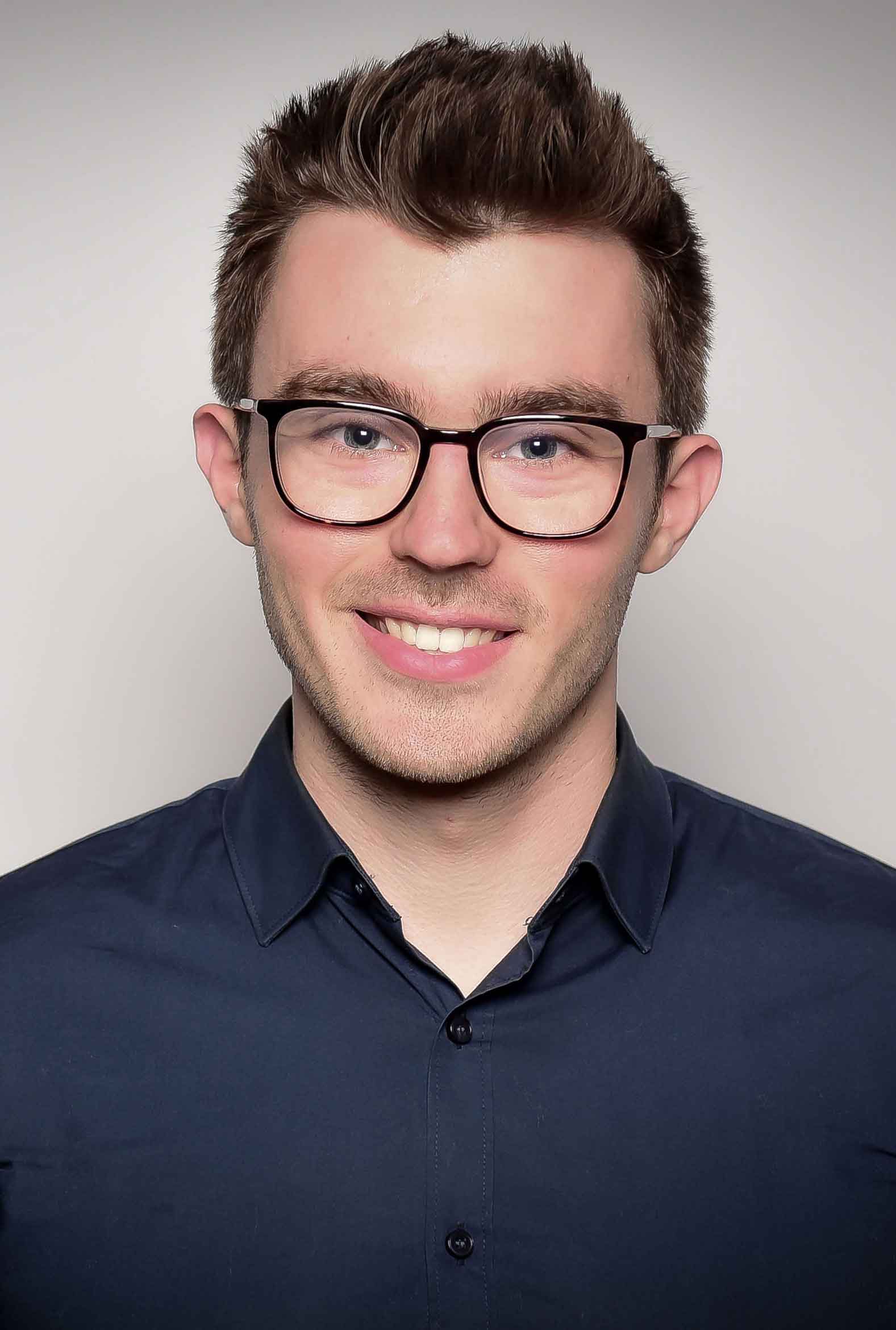}}]{Sven Kirchner} joined the Chair of Robotics, Artificial Intelligence, and Real-time Systems at the Technical University of Munich (TUM) in Germany as a Ph.D. Student in 2023. He received his B.Sc. and M.Sc. in Automotive Engineering from RWTH Aachen University. During his studies, he focused on the control of automated vehicles and automotive systems engineering. His research interests include autonomous driving, safe architectures for software-defined vehicles, and artificial intelligence. \end{IEEEbiography}

\begin{IEEEbiography}[{\includegraphics[width=1in,height=1.25in,clip,keepaspectratio]{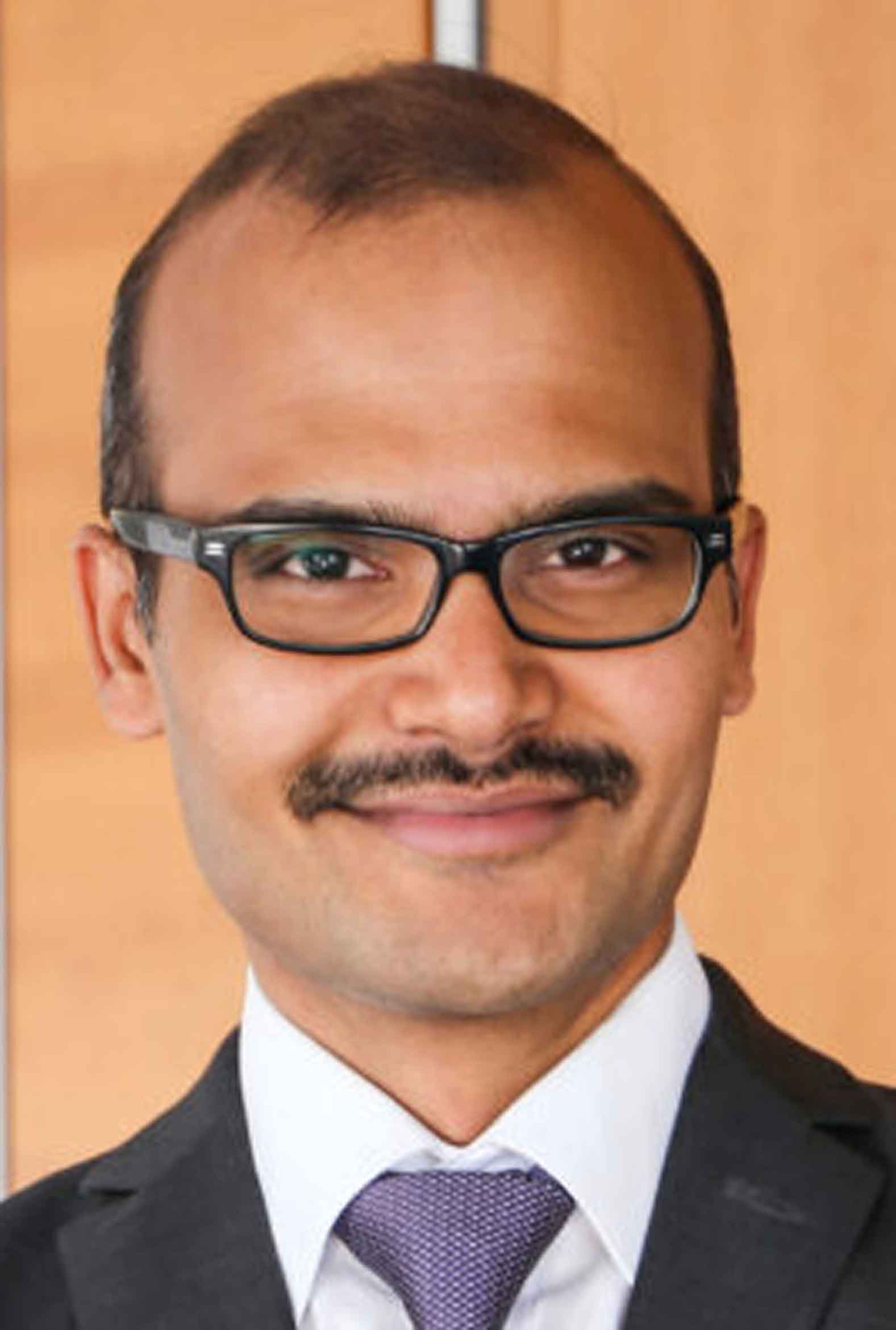}}]{Venkatnarayanan Lakshminarasimhan} is a Ph.D. candidate and Research Assistant at the Chair of Robotics, Artificial Intelligence, and Real-time Systems, TUM School of Computation, Information and Technology, Technical University of Munich. His current research interests include V2X communication in intelligent transportation systems. \end{IEEEbiography}

\begin{IEEEbiography}[{\includegraphics[width=1in,height=1.25in,clip,keepaspectratio]{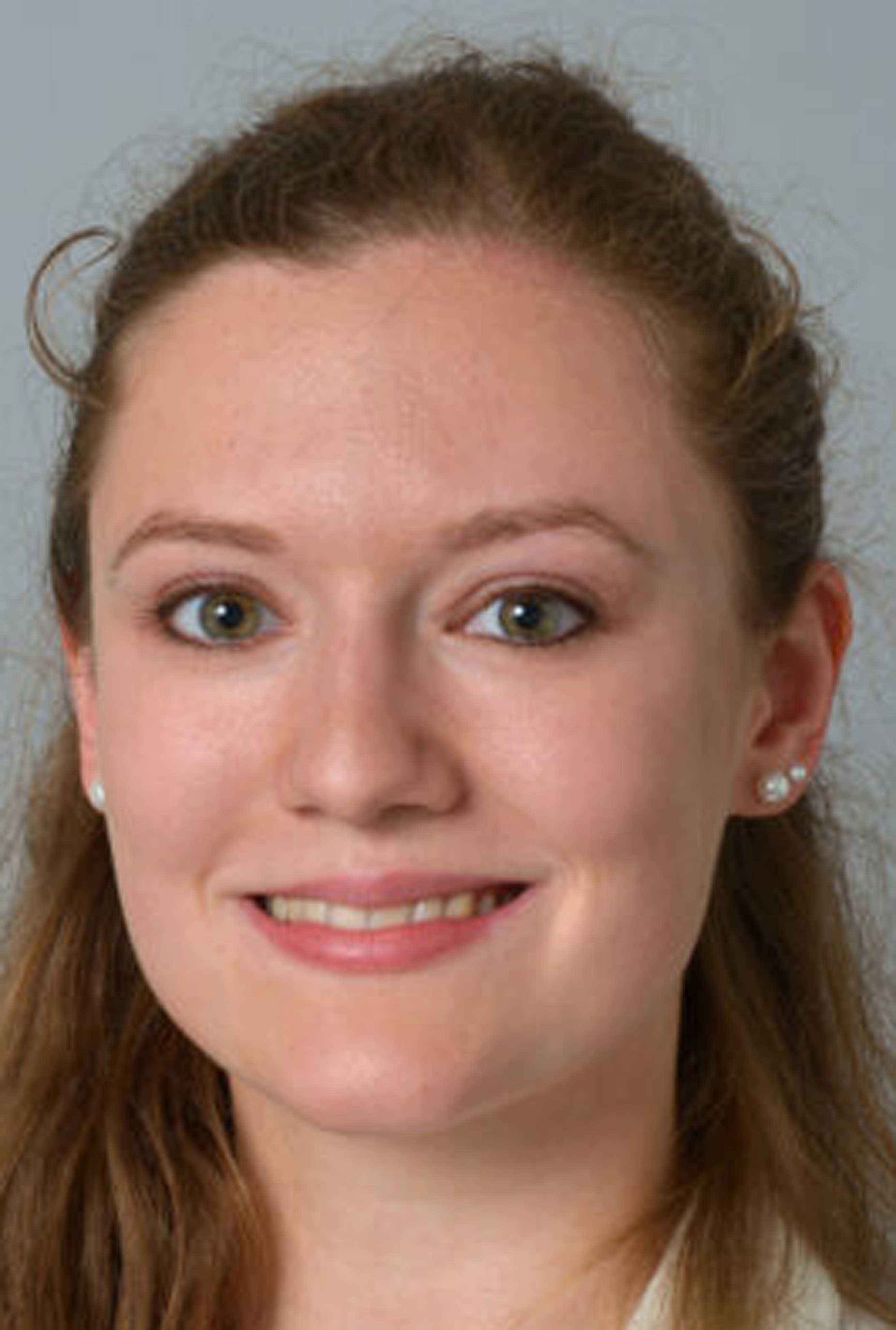}}]{Leah Strand} joined the Chair of Robotics, Artificial Intelligence, and Real-time Systems at the TUM as a research assistant and a Ph.D. candidate in March 2020. She received her Master’s degree in Mechatronics and Information Technology from the TUM after finishing her Bachelor’s degree in Engineering Science at the TUM. Her research interests are autonomous driving, multi-object tracking, sensor calibration, and fusion. \end{IEEEbiography}

\begin{IEEEbiography}[{\includegraphics[width=1in,height=1.25in,clip,keepaspectratio]{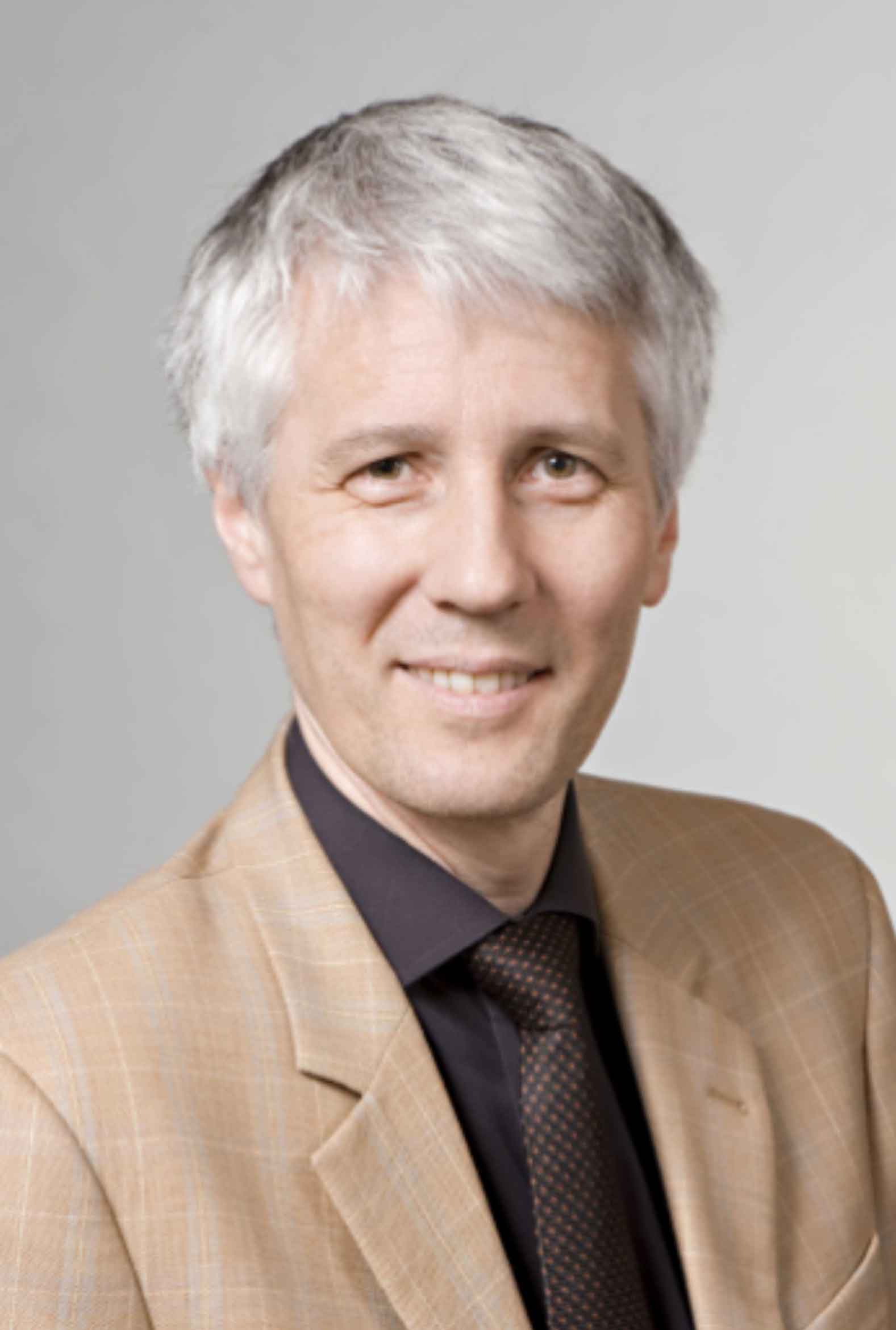}}]{Alois Knoll} (Senior Member) received the diploma (M.Sc.) degree in electrical/communications engineering from the University of Stuttgart, Stuttgart, Germany, in 1985, and the	Ph.D. (summa cum laude) degree in computer science from the Technical University of Berlin (TU Berlin), Berlin, Germany, in 1988. He served on the faculty of the Computer Science Department at TU Berlin until 1993. He joined the University of Bielefeld, Germany, as a Full Professor and served as the Director of the Technical Informatics Research Group until 2001. Since 2001, he has been a Professor at the Department of Informatics, Technical University of Munich (TUM), Munich, Germany. He was also on the Board of Directors of the Central Institute of Medical Technology, TUM (IMETUM). From 2004 to 2006, he was the Executive Director of the Institute of Computer Science, TUM. Between 2007 and 2009, he was a member of the EU’s highest advisory board on information technology, the Information Society Technology Advisory Group, and a member of its subgroup on Future and Emerging Technologies (FET). In this capacity, he was actively involved in developing the concept of the EU’s FET Flagship projects. His research interests include cognitive, medical, and sensor-based robotics, multi-agent systems, data fusion, adaptive systems, multimedia information retrieval, model-driven development of embedded systems with applications to automotive software and electric transportation, and simulation systems for robotics and traffic.\end{IEEEbiography}

\end{document}